\documentclass{article} %
\usepackage{colm2024_conference}
\colmfinalcopy %

\usepackage{amsmath,amsfonts,bm}
\usepackage{wrapfig}

\def\eqref#1{equation~\ref{#1}}

\def\1{\bm{1}}

\DeclareMathAlphabet{\mathsfit}{\encodingdefault}{\sfdefault}{m}{sl}
\SetMathAlphabet{\mathsfit}{bold}{\encodingdefault}{\sfdefault}{bx}{n}

\DeclareMathOperator*{\argmax}{arg\,max}
\DeclareMathOperator*{\argmin}{arg\,min}

\usepackage{here}
\usepackage{comment}
\usepackage{color}
\usepackage{xcolor}

\usepackage{bm}
\usepackage{graphicx}
\usepackage{pdfpages}
\usepackage{algorithm}
\usepackage{algorithmic}
\usepackage{latexsym}
\usepackage{amssymb}
\usepackage{amsmath,amssymb}
\usepackage{mathtools}
\usepackage{url}

\usepackage{hyperref}

\hypersetup{%
  colorlinks=true,%
  citecolor=blue,
  linkcolor=red,%
  linkbordercolor=red,%
}

\usepackage{subcaption}
\usepackage[english]{babel}
\usepackage{amsthm}
\theoremstyle{definition}

\renewcommand{\algorithmicrequire}{\textbf{Params:}}
\renewcommand{\algorithmicensure}{\textbf{init:}}

\newtheorem{prop}{Proposition}[section]

\newtheorem{rem}{Remark}[section]

\newtheorem{assum}{Assumption}[section]

\usepackage{lscape}
\usepackage{wrapfig}

\title{Takeuchi's Information Criteria as a Generalization Measure \\for Deep Neural Networks Close to the NTK Regime}

\author{%
  Hiroki Naganuma$^{1,2}$\thanks{Corresponding author: \texttt{naganuma.hiroki@mila.quebec}. This work was partially conducted while H.~Naganuma was participating in the research residency program at CyberAgent AI Lab. This manuscript is a revised version of a submission to ICLR 2022 that was not accepted.}
  \quad
  Taiji Suzuki$^{3,4}$
  \quad
  Rio Yokota$^{5}$
  \\
  \textbf{Masahiro Nomura$^{6}$\thanks{Work done while at CyberAgent AI Lab.}}
  \quad
  \textbf{Kohta Ishikawa$^{7}$}
  \quad
  \textbf{Ikuro Sato$^{6,7}$}
  \\[1em]
  $^{1}$Department of Computer Science and Operations Research, Universit\'{e} de Montr\'{e}al, Canada
  \\  
  $^{2}$Mila -- Quebec AI Institute, Canada
  \\
  $^{3}$Department of Mathematical Informatics, The University of Tokyo, Japan
  \\
  $^{4}$Center for Advanced Intelligence Project, RIKEN, Japan
  \\
  $^{5}$Supercomputing Research Center, Institute of Science Tokyo, Japan
  \\
  $^{6}$Department of Computer Science, Institute of Science Tokyo, Japan
  \\
  $^{7}$DENSO IT Laboratory, Japan
}

\begin{document}
\maketitle

\begin{abstract}
Generalization measures have been studied extensively in the machine learning community to better characterize generalization gaps. However, establishing a reliable generalization measure for statistically singular models such as deep neural networks (DNNs) is difficult due to their complex nature.
This study focuses on Takeuchi's information criterion (TIC) to investigate the conditions under which this classical measure can effectively explain the generalization gaps of DNNs. Importantly, the developed theory indicates the applicability of TIC near the neural tangent kernel (NTK) regime.
In a series of experiments, we trained more than 5,000 DNN models with 12 architectures, including large models (e.g., VGG-16), on four datasets, and estimated the corresponding TIC values to examine the relationship between the generalization gap and the TIC estimates.
We applied several TIC approximation methods with feasible computational costs and assessed the accuracy trade-off. Our experimental results indicate that the estimated TIC values correlate well with the generalization gap under conditions close to the NTK regime. However, we show both theoretically and empirically that outside the NTK regime such correlation disappears. Finally, we demonstrate that TIC provides better trial pruning ability than existing methods for hyperparameter optimization.

\end{abstract}

\section{Introduction}
\label{sec_intro}
Deep neural networks (DNNs) exhibit highly desirable generalization capabilities in many applications, but the mechanism of generalization is not fully understood~\citep{Neyshabur2014, Zhang2016, Recht2019}.
Establishing a reliable generalization measure is important for obtaining a good model from limited data resources, including in the context of hyperparameter search.
Many attempts~\citep{Arora2018, Wei2019, Neyshabur2018} have been made to better understand the generalization phenomenon in deep learning models from a theoretical perspective.
There have also been intensive empirical studies~\citep{Keskar2016, Liang2017, Bartlett2017} in search of conditions that are likely to yield high model performance.

Work by~\citet{Yiding2019} indicates that a measure that includes both the Hessian $\boldsymbol{H}(\boldsymbol{\theta})$\footnote{$\boldsymbol{H}(\boldsymbol{\theta})$ and $\boldsymbol{C}(\boldsymbol{\theta})$ are defined in equation \ref{eq/ch}.} and the covariance $\boldsymbol{C}(\boldsymbol{\theta})$, defined from the loss and the network parameters $\boldsymbol{\theta}$ near a local minimum, may show good correlation with generalization performance.
Another study by \citet{Novak2018} has indicated that the use of only a single measure, either $\boldsymbol{H}(\boldsymbol{\theta})$ or $\boldsymbol{C}(\boldsymbol{\theta})$, fails to capture generalization performance.

The generalization gap inherently stems from a discrepancy between the empirical and the true data distributions.
A minimizer of the empirical loss is affected by the noise due to a finite number of samples and by the shape of the loss landscape near the minimum.
The former can be characterized as noise ($\boldsymbol{C}(\boldsymbol{\theta})$) and the latter as curvature ($\boldsymbol{H}(\boldsymbol{\theta})$).

Taking these findings into account, we sought to model the generalization gap and found that a classical information criterion, Takeuchi's Information Criteria (TIC)~\citep{Takeuchi1976}, effectively expresses the generalization gap in the neural tangent kernel (NTK) regime.
TIC has the following form:
\begin{equation}
    \label{eq/tic}
    \underbrace{ \mathrm{TIC}(\bm\theta) }_{\text{Information Criteria}} = - 
    \underbrace{ \mathbb{E}_{\hat{p}}[l ( y, f(x, \bm\theta))]}_{\text{Mean Empirical Error}} + 
    \underbrace{\mathrm{Tr}\left(\boldsymbol{H}(\bm{ {\theta})}^{-1} \boldsymbol{C}(\bm{{\theta}})\right)}_{\text{Estimated Bias Term}},
\end{equation}
where $f$ is a smooth function over $\bm \theta \in \mathbb{R}^{d}$ with input $x$ and target $y$, and $l$ is the negative log-likelihood, also referred to as the loss function.
The first term on the right-hand side is the mean empirical loss, which takes the expectation over an empirical data distribution $(x_i,y_i) \sim \hat{p}$.
In what follows, we use $\bm{\hat{\theta}}$ to denote the value that minimizes the empirical loss, i.e., $\bm{\hat{\theta}} = \argmin_\theta \mathbb{E}_{\hat{p}}[l ( y, f(x, \bm\theta))]$, and $\bm{{\theta}^*}$ to denote the parameters that minimize the expected loss with respect to the true data distribution $(x,y) \sim p$, i.e., $\bm{{\theta}^*} = \argmin_\theta \mathbb{E}_{p}[l ( y, f(x, \bm\theta))]$.

Intuitively, $\mathrm{Tr}(\boldsymbol{H}^{-1}\boldsymbol{C})$ measures the effective number of parameters by quantifying the mismatch between the curvature of the loss landscape ($\boldsymbol{H}$) and the noise in the gradients ($\boldsymbol{C}$). When the model is well-specified, $\boldsymbol{H} = \boldsymbol{C}$ and the bias reduces to $d$, recovering the AIC. The derivation of this bias correction is given in detail in Appendix~\ref{appendix:a12}.

For a DNN of practical size, exact computation of the matrices $\boldsymbol{H}(\boldsymbol{\theta})$ and $\boldsymbol{C}(\boldsymbol{\theta})$ is nearly infeasible due to their large dimensionality.
To make the computation feasible, we adopted a strategy of exploiting shared components among the matrices $\boldsymbol{H}(\boldsymbol{\theta})$, $\boldsymbol{C}(\boldsymbol{\theta})$, and the Fisher information matrix $\boldsymbol{F}(\boldsymbol{\theta})$ to estimate TIC with fewer computations.
To further reduce the computational cost for the bias term, we examined approximation methods and lower bounds so that TIC estimation for DNNs of practical sizes becomes tractable.

While \citet{Thomas2019} first demonstrated the potential of TIC as a generalization measure, their study was limited to small two-layer networks with at most a few hundred parameters, and did not provide theoretical justification for applying TIC to DNNs, which are singular models.
Our work extends this line of research in three directions:
{
\setlength{\leftmargini}{10pt}
\begin{itemize}
    \setlength{\itemsep}{1pt}

    \item We provide a theoretical basis for the applicability of TIC to DNNs by showing that the regularity conditions required for TIC are satisfied in the NTK regime (Proposition~1), and we empirically verify that the correlation between TIC and the generalization gap holds under conditions close to this regime, while it disappears outside of it.

    \item We develop and evaluate computationally feasible approximation methods for TIC, enabling its estimation for practical-scale DNNs (up to VGG-16 with $\sim$134M parameters). Through experiments with more than 5,000 models across 12 architectures and 4 datasets, we systematically characterize the accuracy-cost trade-off of these approximations.

    \item We demonstrate a practical application of TIC as a criterion for trial pruning in hyperparameter optimization (HPO), showing that it can prevent promising candidates from being pruned prematurely compared to the standard validation-loss-based approach.
\end{itemize}
}

\section{Generalization Measures}

Generalization measures quantify the generalization ability of statistical models.
Typically, the generalization gap, which is defined as the difference between training loss and validation loss, is used to quantify generalization ability.

\subsection{Which Generalization Measure is Promising?}

To answer this question, and before demonstrating the effectiveness of TIC, we begin by reviewing the development of research in this area and the motivation for this work.
Two major quantification approaches are available for understanding generalization behavior: quantifying generalization bounds and developing complexity measures.

Quantifying generalization bounds is the approach pursued by theoretical groups to prove bounds on the generalization gap \citep{DR17}.
Although tight bounds can be proven, they are often based on assumptions that do not hold in practical DNN settings.
Moreover, no bounds have been shown to describe the performance of current DNNs to a satisfactory degree.

On the other hand, the approach of quantifying complexity measures, which does not necessarily certify bounds, follows the principle of Occam's razor in evaluating the complexity of the model.
Theoretically motivated complexity measures, including VC-dimension \citep{Vapnik1971}, the PAC-Bayesian framework \citep{McAllester1999}, and the norm of parameters \citep{Neyshabur2015b}, are often discussed as significant components of generalization bounds, and a monotonic relationship between complexity measures and generalizations has been mathematically established.
In contrast, empirically motivated generalization measures, such as sharpness \citep{Keskar2016}, are justified by experiments and observations.
In particular, for DNNs, \citet{Yiding2019} have conducted exhaustive experiments to evaluate the effectiveness of generalization measures for three groups:
norm-based measures, sharpness-based measures, and noise-based measures.

{
\setlength{\leftmargini}{10pt}%
\begin{itemize}
  \setlength{\itemsep}{1pt}

  \item \textbf{Norm-based measures: $|\boldsymbol{\theta}|$.}
  Most of the proposed norm-based measures are based on the Fisher-Rao metric \citep{Liang2017}, which does not capture generalization well.
  \citet{Yiding2019} reported that spectral complexity measures such as the product of spectral norms of the layers \citep{Bartlett2017} show limited correlation with the generalization gap in their extensive evaluation.
  Overall, norm-based metrics alone have not been sufficient to explain the generalization behavior of modern overparameterized DNNs.

  \item \textbf{Sharpness-based measures: $\boldsymbol{H}(\boldsymbol{\theta})$.}
  Sharpness-based metrics, such as sharp minima and flat minima \citep{Keskar2016} and the PAC-Bayesian framework \citep{McAllester1999},
  are not only associated with intuitive understanding but also empirically show a strong correlation with the generalization gap.
  However, some model architectures are known to show poor correlation \citep{dinh2017sharp}, and these measures are sensitive to reparameterization of the network.
  Sharpness-aware minimization (SAM) \citep{Foret2021} has popularized sharpness as a training objective, but the underlying metric remains $\boldsymbol{H}(\boldsymbol{\theta})$ alone and does not account for the gradient noise structure $\boldsymbol{C}(\boldsymbol{\theta})$.

  \item \textbf{Noise-based measures: $\boldsymbol{C}(\boldsymbol{\theta})$.}
  Experimental results show that gradient-based generalization measures have potential \citep{Yiding2019}.
  In particular, \citet{Yiding2019} observed that while the variance of the gradient captures sharpness, it is not necessarily an effective generalization measure, depending on the model architecture.

\end{itemize}
}

These results suggest that studying generalization measures that use both $\boldsymbol{H}(\boldsymbol{\theta})$ and $\boldsymbol{C}(\boldsymbol{\theta})$ is promising.
TIC is distinguished from pure sharpness measures in that it captures not only the curvature of the loss landscape but also its interaction with the gradient noise structure; the ratio $\boldsymbol{H}^{-1}\boldsymbol{C}$ naturally accounts for the fact that sharp directions with low gradient variance may not harm generalization, while flat directions with high variance can.
However, since the combination of $\boldsymbol{H}(\boldsymbol{\theta})$ and $\boldsymbol{C}(\boldsymbol{\theta})$ seen in TIC is not feasible to compute for practical DNN settings, it was not considered within the scope of \citet{Yiding2019}.

\subsection{Information Matrix: Elements of Generalization Measures}
\label{sec_2_2}

Previous research has highlighted the importance of information matrices such as $\boldsymbol{H}(\boldsymbol{\theta})$ and $\boldsymbol{C}(\boldsymbol{\theta})$ in generalization measures for DNNs.
\citet{Thomas2019, Kunstner2020} remarked that these matrices are often confused and misused (for example, in the field of optimization), leading to incorrect conclusions, even though they play an essential role in the study of DNNs, especially in optimization \citep{Amari2020, Martens2015}, understanding implicit regularization in SGD \citep{Yeming2019, Zhanxing2019}, and Bayesian inference \citep{Zhang2017}.
Before discussing these generalization measures, we first clarify how each of the information matrices is defined.

In this paper, the uncentered gradient covariance matrix is denoted as $\boldsymbol{C}(\boldsymbol{\theta})$.
We define $q_{\bm \theta}(x,y) = p(x) f(y|x,\bm\theta)$ as the model distribution, i.e., the joint distribution obtained by combining the marginal input distribution $p(x)$ with the model's conditional distribution $f(y|x,\bm\theta)$.
Furthermore, we employ the data distributions $\hat{p}$ and $p$ introduced in the previous section as the empirical and true data distributions, respectively.
The matrices $\boldsymbol{H}(\boldsymbol{\theta})$, $\boldsymbol{C}(\boldsymbol{\theta})$, and $\boldsymbol{F}(\boldsymbol{\theta})$ are then defined as follows:

\begin{align}\label{eq/ch}
\begin{split}
&{\boldsymbol{H}(\bm{{\theta}})} = \mathbb{E}_{p}\left[ \frac{\partial^{2} l( y, f(x, {\bm{\theta}}))}{\partial \boldsymbol{\theta} \partial \boldsymbol{\theta}^{T}} \right] \in \mathbb{R}^{d \times d},\\
&{\boldsymbol{C}(\bm{{\theta})}} = \mathbb{E}_{{p}}\left[ \frac{\partial l( y, f(x, {\bm{\theta}}))}{\partial \boldsymbol{\theta}} \frac{\partial l( y, f(x, {\bm{\theta}}))}{\partial \boldsymbol{\theta}^{T}} \right] \in \mathbb{R}^{d \times d}, \\
&{\boldsymbol{F}(\bm{{\theta})}} = \mathbb{E}_{q_{\bm \theta}}\left[ \frac{\partial l( y, f(x, {\bm{\theta}}))}{\partial \boldsymbol{\theta}} \frac{\partial l( y, f(x, {\bm{\theta}}))}{\partial \boldsymbol{\theta}^{T}} \right] \in \mathbb{R}^{d \times d}.
\end{split}
\end{align}

The conditions under which these matrices are equal will be discussed in detail in Section \ref{sec/4_1}.
The relation between $\boldsymbol{C}(\boldsymbol{\theta})$ and $\boldsymbol{F}(\boldsymbol{\theta})$ is often misunderstood, as they both involve the outer product of the gradients but differ in the distribution used when computing the expectation.

In a subsequent study, \citet{Novak2018} concluded that consideration of either $\boldsymbol{H}(\boldsymbol{\theta})$ or $\boldsymbol{C}(\boldsymbol{\theta})$ alone is insufficient to estimate the generalization of DNNs and that both are essential.
In particular, $\boldsymbol{H}(\boldsymbol{\theta})$ does not depend on the distribution of input data; however, as the generalization ability depends on the data distribution, it is natural to also consider $\boldsymbol{C}(\boldsymbol{\theta})$, which is related to noise in the gradient.
Furthermore, as supporting evidence for the claim of \citet{Novak2018}, \citet{Thomas2019} empirically showed the effectiveness of TIC, a generalization measure that considers both $\boldsymbol{H}(\boldsymbol{\theta})$ and $\boldsymbol{C}(\boldsymbol{\theta})$ as expressed in Equation \ref{eq/tic}.
However, \citet{Thomas2019} conducted their experiments with only very small-scale NNs, recognizing the challenge of calculating TIC values for DNNs of practical size.
In fact, even the ResNet-8 model used in the small-scale image classification benchmark CIFAR-10 is not feasible, as nearly 200 TB of memory is required to calculate the exact TIC value.

The computation of curvature matrices for DNNs has received considerable attention in the context of Bayesian deep learning.
The Laplace approximation, which uses a Gaussian centered at the MAP estimate with covariance given by the inverse Hessian, relies on the same matrices that appear in TIC.
\citet{Daxberger2021} proposed scalable Laplace approximations by using subnetwork and last-layer variants, enabling posterior-based model selection for practical DNNs.
Their work demonstrates that curvature-based quantities can be computed at scale, which is complementary to our approximation strategy for TIC.
While the Laplace approximation targets the posterior predictive distribution, TIC targets the frequentist bias correction; both share the computational challenge of handling $\boldsymbol{H}(\boldsymbol{\theta})$.

\begin{rem}
\label{rem:singular}
A statistical model $f(y|x,\bm\theta)$ is called \emph{regular} if the Fisher information matrix $\boldsymbol{F}(\bm\theta)$ is positive definite at the true parameter, the map from parameters to distributions is one-to-one, and standard asymptotic normality of the maximum likelihood estimator holds \citep{Watanabe2009}.
A model that violates any of these conditions is called \emph{singular}.
DNNs are singular models because they possess parameter symmetries (e.g., permutation of hidden units) and degenerate Fisher information matrices.
TIC is an information criterion derived under the regularity assumptions and its theoretical justification for singular models such as general DNNs is not established.
However, as we show in Section~\ref{sec_3}, the NTK regime effectively restores the regularity conditions, which provides a theoretical basis for applying TIC to DNNs in this regime.
\end{rem}

\subsection{Deriving TIC as the Generalization Gap in the NTK Regime}
\label{sec_3}
This section outlines the derivation of TIC in Equation \ref{eq/tic}, considering the generalization gap of DNNs within the framework of the NTK regime.
We employ the setting introduced in Section \ref{sec_intro}, where $f$ is a smooth function over $\bm \theta \in \mathbb{R}^{d}$, the parameter of the statistical model.
We further assume that the following conditions hold for $f$ and $\bm \theta \in \mathbb{R}^{d}$ in the NTK regime.

\begin{assum}\label{assum:1}
\text{ }
\begin{enumerate}
\item[{\em (A1)}]
Global convergence: the optimization landscape has a unique minimizer $\hat{\bm\theta}$ of the empirical loss. This does not require that $q_{\bm \theta} = p$ (allowing for a misspecified model). In the NTK regime, this condition is satisfied because the loss landscape is locally convex around the initialization when the network width is sufficiently large, and the solution is uniquely determined as shown in Equation~\ref{eq:unique} (Appendix~\ref{appendix:a12}).

\item[{\em (A2)}]
Asymptotic normality: the maximum likelihood estimator $\hat{\bm{\theta}}$ from the empirical data distribution $\hat{p}$ and the minimizer $\bm {\theta^{*}}$ of the expected loss under the true data distribution $p$ satisfy asymptotic normality. In the NTK regime, this follows from the Gaussian process behavior of the network output \citep{Lee2018}.
\end{enumerate}
\end{assum}

\begin{prop}[\textbf{Generalization Gap in NTK Regime is Equal to TIC}]
\text{ } 
  
        Under assumptions (A1) and (A2), the bias $b$ of the empirical loss as an estimator of the expected loss is given by
   
\begin{align}\label{eq/tic_ntk}
\begin{split}
        b &= \mathbb{E}_{p} \left[  \mathbb{E}_{\hat{p}}[l(\bm y,f(\bm x, \hat{\bm{\theta}}))] - \mathbb{E}_{p}[l(\bm y, f(\bm x, \hat{\bm{\theta}}))]   \right] \\
        & = \mathrm{Tr}\left({\boldsymbol{H}}_p(\bm{{\theta^{*}})}^{-1} {\boldsymbol{C}}_p(\bm{{\theta^{*}}})\right)
\end{split}.
\end{align}
        
        Here $\boldsymbol{H}_p(\boldsymbol{\theta}^{*})$ and $\boldsymbol{C}_p(\boldsymbol{\theta}^{*})$ are the Hessian and covariance, respectively, evaluated at $\boldsymbol{\theta}^{*}$ under the true data distribution $p$.
        Since the true data distribution $p$ and the parameter $\boldsymbol{\theta}^{*}$ that minimizes the expected loss are unknown, their consistent estimators based on the empirical data distribution $\hat{p}$ and the parameter $\hat{\boldsymbol{\theta}}$ are used in practice, which yields the TIC. A more detailed derivation is given in Appendix \ref{appendix:a12}.
    
\end{prop}

\begin{rem}
\label{rem_bound}
The bias term of the TIC is formulated as $\mathrm{Tr}\left(\boldsymbol{H}(\bm{ {\theta})}^{-1} \boldsymbol{C}(\bm{{\theta}})\right)$. However, there is no guarantee that $\boldsymbol{H}(\bm{ {\theta})}$ is positive definite in practice. To prevent this problem, addition of a small multiple of the identity matrix, called damping, is performed as  $\tilde{\boldsymbol{H}}(\bm{ {\theta})}^{-1} = {\left( \boldsymbol{H}(\bm{ {\theta})}+ \lambda I \right)}^{-1}$.
Alternatively, consider the case where the TIC is calculated by approximation with a matrix of only the diagonal components of the respective matrices, as $\mathrm{Tr}\left(\boldsymbol{H}(\bm{ {\theta})}^{-1} \boldsymbol{C}(\bm{{\theta}})\right) \approx \mathrm{Tr}\left(\boldsymbol{H}_\text{diag}(\bm{ {\theta})}^{-1} \boldsymbol{C}_\text{diag}(\bm{{\theta}})\right) $
In this case, the following lower bound is given for the diagonal approximated TIC:
\begin{align}\label{eq/lower_bound}
\begin{split}
        \mathrm{Tr}\left(\boldsymbol{H}_\text{diag}(\bm{ {\theta})}^{-1} \boldsymbol{C}_\text{diag}(\bm{{\theta}})\right) 
        >
        \frac{\mathrm{Tr}(\boldsymbol{C}_\text{diag}(\bm{{\theta}))}}{\mathrm{Tr}(\boldsymbol{H}_\text{diag}(\bm{{\theta}))}}
        =
        \frac{\mathrm{Tr}(\boldsymbol{C}(\bm{{\theta}))}}{\mathrm{Tr}(\boldsymbol{H}(\bm{{\theta}))}}
\end{split}
\end{align}
\end{rem}

\begin{rem}
\label{rem:ntk_indicator}
We note that not all DNNs are in the NTK regime.
In this work, we use the ratio $d/n$ as a practical proxy for proximity to the NTK regime; a large ratio suggests that the network is sufficiently overparameterized for the NTK approximation to be reasonable.
However, $d/n$ alone is not a sufficient condition.
The NTK regime additionally requires that the network width is large enough for the empirical NTK $\mathcal{K}_t$ to remain approximately constant during training \citep{Arthur2018}, which can be affected by learning rate scaling, initialization scheme, and architectural choices such as residual connections and batch normalization \citep{Lee2019, Yang2021}.
More refined diagnostics, such as measuring the change in the empirical NTK during training $\|\mathcal{K}_t - \mathcal{K}_0\|$ or the linearization gap $\|f_t - f_t^{\mathrm{lin}}\|$ \citep{Fort2020}, could provide a more precise characterization of NTK proximity, though we leave such analysis for future work.
In our experiments (Section~\ref{section:experiments}), we observe that models with larger $d/n$ and those using shortcut connections (which reduce the effective nonlinearity) tend to show stronger correlation between TIC and the generalization gap.
\end{rem}

\begin{rem}
\label{rem:waic}
Outside the NTK regime, DNNs undergo feature learning where the kernel $\mathcal{K}_t$ changes substantially during training, violating the local convexity and uniqueness assumptions that underpin TIC.
In this regime, DNNs are singular models (see Remark~\ref{rem:singular}): the Fisher information matrix becomes degenerate due to parameter symmetries, the maximum likelihood estimator is no longer asymptotically normal, and the bias correction $\mathrm{Tr}(\boldsymbol{H}^{-1}\boldsymbol{C})$ loses its theoretical justification.
The WAIC \citep{Watanabe2012}, derived from singular learning theory \citep{Watanabe2009}, is theoretically more appropriate in this setting, as it accounts for the algebraic singularities of the posterior through the real log canonical threshold (RLCT).
However, WAIC requires sampling from the posterior distribution, and its computational cost is prohibitive for large-scale DNNs.
Recent work on scalable Laplace approximations \citep{Daxberger2021} and stochastic gradient MCMC provides potential pathways for posterior-based model selection at scale, but these remain substantially more expensive than the frequentist TIC approach.
Similarly, when the loss function includes a regularization term, the GIC \citep{Konishi1996} is technically more appropriate than TIC, but GIC introduces additional computational complexity.
The practical advantage of TIC is that it can be computed from the trained model alone, without posterior sampling, making it applicable at scale with the approximation methods described in Section~\ref{sec/4_0}.
\end{rem}

\section{Approximation of TIC}
\label{sec/4_0}

\subsection{Hessian, Generalized Gauss-Newton Matrix (GGN) and FIM}
\label{sec/4_1}
In this section, we describe the conditions under which the Hessian, GGN, and FIM become equivalent.
This equivalence can be exploited to reduce the computational cost of computing TIC.
Calculating TIC requires the computation of $\boldsymbol{H}(\boldsymbol{\theta})$ and $\boldsymbol{C}(\boldsymbol{\theta})$; however, the computational cost of $\boldsymbol{H}(\boldsymbol{\theta})$ is relatively high.
For NNs that consist of linear, convolutional, and pooling layers, along with piecewise linear activations, the Hessian is equal to the GGN \citep{Schraudolph2002}.
This holds true for most CNNs used in practice.
The GGN is an extension of the Gauss-Newton matrix $\bm{\tilde{G}}(\bm{\theta}) = \mathbb{E}_{p} \left[ (\bm{J}_{\bm{\theta}})^{T} \bm{J}_{\bm{\theta}} \right] $.

\begin{equation}
    \bm{G}(\bm{\theta}) = \mathbb{E}_{p} \left[ (\bm{J}_{\bm{\theta}})^{T}  \bm{H}_{f} \bm{J}_{\bm{\theta}} \right]
\end{equation}

where $\bm{H}_{f} = \frac{\partial^2 l(y, z)}{\partial z \partial z^T}\big|_{z=f(x,\bm\theta)}$ is the Hessian of the loss with respect to the model output $f$, and $\bm{J}_{\bm{\theta}} = \frac{\partial f(x, \bm\theta)}{\partial \bm\theta}$ is the Jacobian of $f$ with respect to ${\bm{\theta}}$.
Furthermore, the GGN is equal to the FIM for any NN that uses the softmax cross-entropy.
Therefore, we can assume the following for most practical DNN problem settings.

\begin{assum}\label{assum:2}
\text{ } 
\begin{enumerate}
\item[{\em (B1)}]
Loss function: $l$ is the softmax cross-entropy function
\item[{\em (B2)}]
Activation function: inside $f$, the second derivative of all activation functions is always zero, such as in case of ReLU or the identity function.
\end{enumerate}
\end{assum}

\begin{prop}[\textbf{$\boldsymbol{H}(\boldsymbol{\theta})$ is equal to $\boldsymbol{F}(\boldsymbol{\theta})$ through the GGN}]
\text{ } 

   Under assumptions (B1) and (B2), $\boldsymbol{H}(\boldsymbol{\theta})$ and $\boldsymbol{F}(\boldsymbol{\theta})$ are exactly equal through the GGN. They are also guaranteed to be positive semi-definite.
   
\begin{equation}
    \bm{H}(\bm{\theta}) = \bm{G}({\bm{\theta}}) = \bm{F}(\bm{\theta})
    \label{eq/h_g_f}
\end{equation}
A more detailed proof is given in \citet{martens2020new}.

\end{prop}

\subsection{Approximation of Matrices and Trace Estimation}

As noted in Section \ref{sec_2_2}, information matrices are needed for many applications, including TIC.
However, for a model in which the number of parameters $d$ is large, as in the case of DNNs, it is necessary to compute a matrix of size $d \times d$.
For this reason, approximation methods ranging from approximating the information matrix itself \citep{Roux2008} to approximating the matrix-vector product directly have been developed for optimization \citep{Pearlmutter1994} and other applications.
We propose the following approximation methods to calculate TIC and experimentally verify the trade-off between accuracy and computation time.

\setlength{\leftmargini}{10pt}%
\begin{itemize}
  \setlength{\itemsep}{1pt} 
    \item \textbf{Replacing $\boldsymbol{H}(\boldsymbol{\theta})$ with $\boldsymbol{F}(\boldsymbol{\theta})$ and fast estimation of $\boldsymbol{F}(\boldsymbol{\theta})$ via Monte Carlo sampling.}
    As shown in Equation \ref{eq/h_g_f}, $\boldsymbol{F}(\boldsymbol{\theta})$ can be used in place of $\boldsymbol{H}(\boldsymbol{\theta})$ under assumptions (B1) and (B2). We use this property to speed up the calculation by simultaneously computing $\boldsymbol{C}(\boldsymbol{\theta})$ and $\boldsymbol{F}(\boldsymbol{\theta})$, which share a common term.
    Furthermore, since the number of classes for the classification task is 10 in MNIST and 100 in CIFAR-100, the computational cost of $\boldsymbol{F}(\boldsymbol{\theta})$ is large; consequently, we approximate $\boldsymbol{F}(\boldsymbol{\theta})$ using $\boldsymbol{F}_\text{mmc}(\boldsymbol{\theta})$, which is a Monte Carlo approximation. \citet{martens2015optimizing} used $m = 1$ in practice. We follow this setting, using $\boldsymbol{F}_\text{1mc}(\boldsymbol{\theta})$ for the approximation of $\boldsymbol{F}(\boldsymbol{\theta})$.
    
    \item \textbf{Block-diagonalization and diagonalization.}
    In the NTK regime, the correlation between layers is small, making block-diagonalization a reasonable approximation.
    This is supported by the empirical findings of \citet{Karakida2020}, who showed that the off-diagonal blocks of the Fisher information matrix (corresponding to cross-layer correlations) are negligible relative to the diagonal blocks in overparameterized networks.
    The computational complexity can be reduced from $O(d^3)$ to $O(d_l^3)$\footnote{$d_l$ is the number of parameters in the layer with the largest number of parameters in the network.} by block-diagonal approximation.
    Diagonalization is a simpler approximation that ignores the correlation between DNN units. It has been reported to be sufficient for some applications \citep{Sidak2020}.
    It can also be calculated as element-wise operations on vectors rather than matrices, significantly reducing computational complexity and memory consumption. In particular, the diagonal approximation reduces the inverse computation of $\boldsymbol{H}(\boldsymbol{\theta})$ from $O(d^3)$ to $O(d)$.
    
    \item \textbf{Lower bound of the diagonal approximation.}
    As shown in Equation \ref{eq/lower_bound}, by using the lower bound of the diagonal approximation, it is possible to compute the TIC bias term by computing the trace of each matrix separately, without computing the diagonal components. This also avoids the need to ensure that $\boldsymbol{F}(\boldsymbol{\theta})$ is positive definite.
    
    \item \textbf{Hutchinson's method for fast estimation of $\mathrm{Tr}(\bm{H}(\bm{\theta}))$.}
    Rather than approximating the matrix itself, we also introduce a method to accelerate the computation of its trace.
    For optimization in deep learning, it suffices to calculate the product of the Hessian and an arbitrary vector, rather than the Hessian itself (Hessian-vector product; Hvp).
    \citet{Pearlmutter1994} proposed a fast algorithm to compute Hvp in NNs during backpropagation.
    Hvp can be applied to non-optimization applications, such as approximating $\mathrm{Tr}(\bm{H}(\bm{\theta}))$ \citep{Avron2011}.
    Hutchinson's method~\citep{hutchinson1989stochastic} approximates the trace by computing the expectation of the quadratic form of the Hessian with Rademacher random vectors (each element takes $1$ or $-1$ with probability $1/2$).
    
\end{itemize}

\begin{figure}[h]
\centering
    \begin{minipage}{0.24\linewidth}
    	\centering
    	\includegraphics[width=\linewidth]{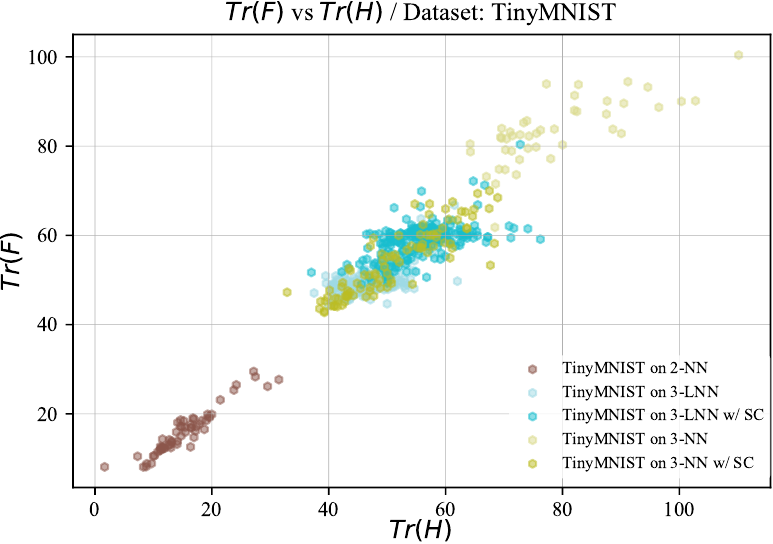}
    	\subcaption{\scriptsize{:  Tr(H) vs Tr(F)}}
    	\label{fig:tinymnist_fast_tic_f}
    \end{minipage}
    \begin{minipage}{0.24\linewidth}
    	\centering
    	\includegraphics[width=\linewidth]{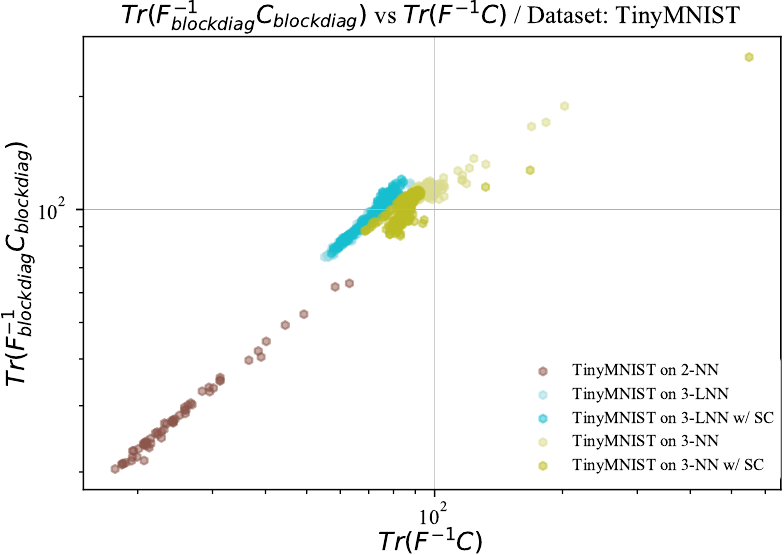}      
    	\subcaption{\scriptsize{: Exact vs block-diagonal}}
    	\label{fig:tinymnist_fast_tic_f}
    \end{minipage}
    \begin{minipage}{0.24\linewidth}
    	\centering
    	\includegraphics[width=\linewidth]{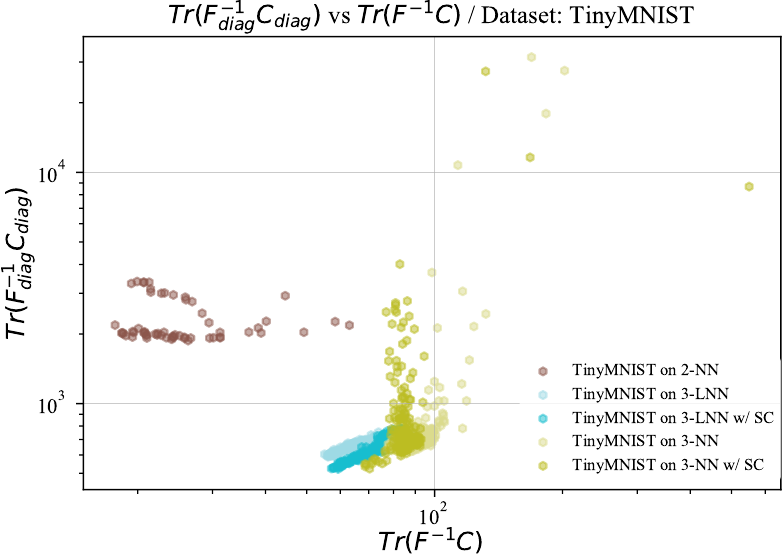}
    	\subcaption{\scriptsize{: Exact vs diagonal}}
    	\label{fig:mnist_fast_tic_f}
    \end{minipage}
    \begin{minipage}{0.24\linewidth}
    	\centering
    	\includegraphics[width=\linewidth]{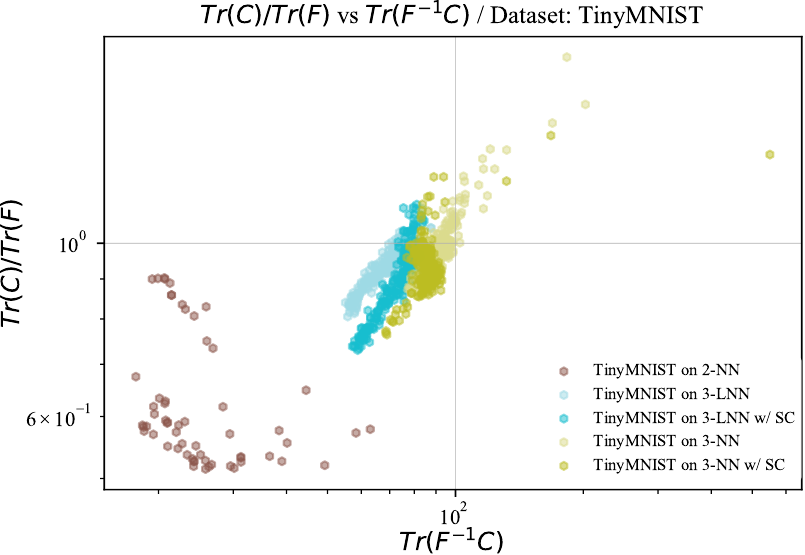}
    	\subcaption{\scriptsize{: Exact vs lower bound}}
    	\label{fig:cifar_fast_tic_f}
    \end{minipage}
    
\caption{\textbf{Approximation comparison experiments in a small-scale setting.} Panel (a) shows the equality of $Tr(H)$ and $Tr(F)$. Panels (b), (c), and (d) compare different approximation methods for the TIC estimation  to the exact case. Full results are shown in Appendix \ref{appendix:approx-exp}.}
\label{fig:main_approx}
\end{figure}

\section{Experiments}
\label{section:experiments}

\subsection{Overview}
\label{section:experiments-overview}
The goal of this paper is to clarify the correlation between TIC estimates and the generalization gap.
To make our study of TIC as comprehensive as possible, we trained models on 4 different datasets with 12 different DNN architectures.
Using these combinations, we searched for hyperparameters for each of the 15 problem settings and evaluated the parameters of the trained models.
By comparing these results, we are able to observe how the effectiveness of TIC changes with the model and problem settings.
In our experiments, the bias term of the TIC is estimated by using validation data; the generalization gap is the absolute value of the difference in loss between the training and test data, using all of the data in each dataset, not just a part of the data.
The problem settings for the experiment are divided into two main categories. Table \ref{table:experimental_category} shows the two categories, along with the corresponding dataset and model sizes.

{
\begin{table}[htb]
\caption{\textbf{Two categories of experimental settings.} Problem settings with $\sharp$ and $\star$ indicate the use of a linear neural network and that the problem setting is considered to be nearly in the NTK regime, respectively. For hyperparameter search, we performed Bayesian optimization for all experimental settings. Appendix \ref{appendix:hyperparameters} provides further detailed configurations of hyperparameters and other settings for the experiment. The remaining experimental settings are explained in Appendix \ref{appendix:setting}.}
\label{table:experimental_category}
\begin{tabular}{llll}
\hline
Category                                                                                                                               & TIC Estimates                                                                                        & Problem Setting: Dataset \& Model                                                                                                                                                                                                                                            & Ratio: $d/n$                                                                                   \\ \hline
\begin{tabular}[c]{@{}l@{}}\textbf{Small-scale}\\ Data\textless1 MB\\ Model\textless  50 KB\end{tabular}                                   & \begin{tabular}[c]{@{}l@{}}Exact and Approx. \\ (Block Diag, Diag, \\ and Lower Bound)\end{tabular} & \begin{tabular}[c]{@{}l@{}}TinyMNIST on 2-NN w/o SC\footnotemark \\ TinyMNIST on 3-NN w/o and w/ SC\\ $\sharp$ TinyMNIST on 3-LNN w/o and w/ SC\end{tabular}                                                                                                                                           & \begin{tabular}[c]{@{}l@{}}0.09\\ 0.02\\ 0.02\end{tabular}                                        \\ \hline
\begin{tabular}[c]{@{}l@{}} \textbf{Practical-scale}\\ Data\textgreater 20 MB\\ Model\textgreater 0.5 MB \end{tabular} & \begin{tabular}[c]{@{}l@{}}Approx. \\ (Diag and \\ Lower Bound)\end{tabular}                       & \begin{tabular}[c]{@{}l@{}}$\star$ MNIST on 6-NN w/o and w/ SC\\ $\sharp\star$ MNIST on 6-LNN w/o and w/ SC\\ $\star$ MNIST on Simple CNN\\ $\sharp\star$ CIFAR10 on 6-LNN w/o and w/ SC\\ $\star$ CIFAR10 on ResNet8 w/o BN \footnotemark \\ $\star$ CIFAR10 on VGG16 w/o BN \\ $\star$ CIFAR100 on ResNet8 w/o BN \end{tabular} & \begin{tabular}[c]{@{}l@{}}2.50 \\ 2.50 \\ 268.92\\ 8.72\\ 122.65\\ 3357.53\\ 122.65\end{tabular} \\ \hline

\end{tabular}
\end{table}

}

In particular, ResNet-8, which is commonly used as a benchmark for training CIFAR-10, requires over 200 TB of memory to compute an exact $\boldsymbol{H}(\boldsymbol{\theta})$. This means that even the state-of-the-art NVIDIA A100 GPU with 80 GB of device memory is insufficient.
Hence, in our \textbf{small-scale} experiment, we use a small dataset called TinyMNIST to limit the size of the DNN model in comparing our approximation method and exact calculation. TinyMNIST is a resized version of the MNIST image, which reduces the dimension of the input layer of the DNN. 
As \textbf{practical-scale} experiments, we evaluated the real-world datasets and DNN models. We used diagonal approximations and their lower bound approximations to estimate TIC.

\begin{rem}
\label{rem:experimental_conditions}
Our experimental configurations do not use batch normalization or data augmentation (see Appendix~\ref{appendix:hyperparameters} for details). These choices are intentional: both batch normalization and data augmentation introduce dependencies that can move the training dynamics away from the NTK regime. By omitting them, we isolate the effect of overparameterization ($d/n$) on NTK proximity and make the regime boundary more interpretable. We acknowledge that modern training recipes typically include these techniques, and investigating how TIC behaves under such conditions is an important direction for future work.
\end{rem}

{
\footnotetext{2-NN and 3-LNN denote the 2-Layer Nonlinear Neural Network and 3-Layer Linear Neural Network, respectively. SC denotes Skip-Connection.} 
\footnotetext{BN denotes Batch Normalization.} 
}

\subsection{Small-Scale Experiments: Comparing Approximation and Exact Results}

\begin{figure}[t]
\centering
    \begin{minipage}{0.32\linewidth}
    	\centering
    	\includegraphics[width=\linewidth]{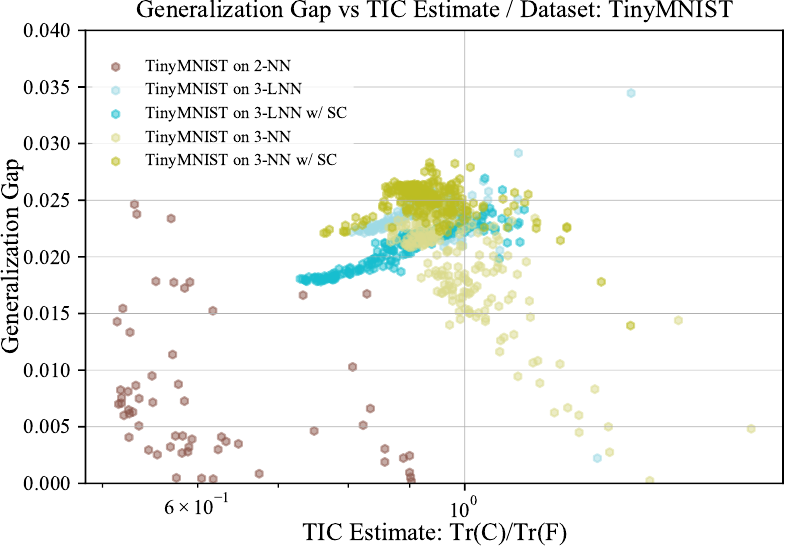}       
    	\subcaption{\scriptsize{: TinyMNIST}}
    	\label{fig:tinymnist_fast_tic_f}
    \end{minipage}
    \begin{minipage}{0.32\linewidth}
    	\centering
    	\includegraphics[width=\linewidth]{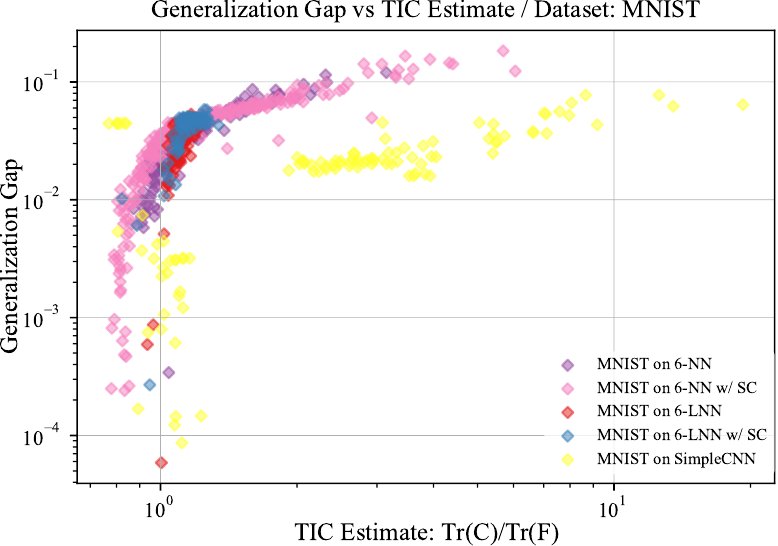}       
    	\subcaption{\scriptsize{: MNIST}}
    	\label{fig:mnist_fast_tic_f}
    \end{minipage}
    \begin{minipage}{0.32\linewidth}
    	\centering
    	\includegraphics[width=\linewidth]{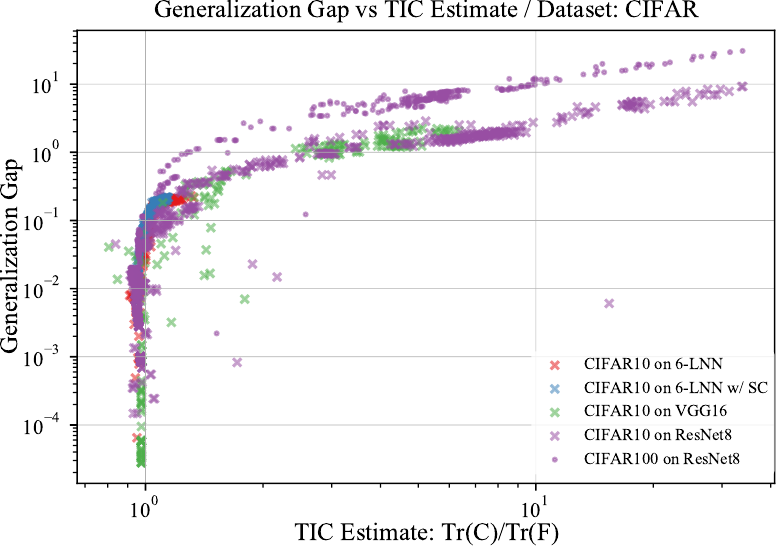}    
    	\subcaption{\scriptsize{: CIFAR10 and CIFAR100}}
    	\label{fig:cifar_fast_tic_f}
    \end{minipage}

\caption{\textbf{Correlation between the generalization gap and the TIC estimates. } Panel (a) shows a problem setting outside the NTK regime, where the correlation between TIC and the generalization gap is weak; Panels (b) and (c) show a problem setting close to the NTK regime, where the correlation is stronger. Full results are given in Appendix \ref{appendix:additional-experiment}.}
\label{fig:main_correlation}
\end{figure}

\begin{figure}[h]
\begin{center}
\includegraphics[width=\linewidth]{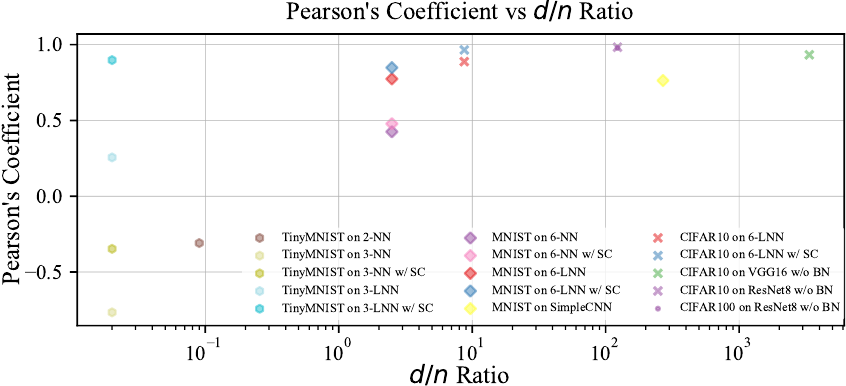}
\end{center}
\caption{\textbf{Relationship between Pearson's correlation (generalization gap and TIC estimates) and $d/n$}. Each point represents one workload (architecture and dataset combination); the Pearson correlation coefficient is computed across all trained models within that workload. The correlation between TIC estimates and the generalization gap is high in regions with large $d/n$, which are considered to be close to the NTK regime. Full results for other metrics, including Spearman's correlation and Kendall's $\tau$, are shown in Appendix \ref{appendix:additional-experiment}.}
\label{fig:pearson_correlation}
\end{figure}
In our \textbf{small-scale} experiments, we trained Tiny MNIST on 5 experimental settings: 3-LNNs and NNs, each w/ and w/o SC, and a wide model, with 2-NNs without SC.
We subsequently evaluated the approximation of $\mathrm{Tr}\left(\boldsymbol{H}(\bm{ {\theta})}^{-1} \boldsymbol{C}(\bm{{\theta}})\right)$, the bias term of TIC, for $\boldsymbol{H}(\boldsymbol{\theta})$ and $\boldsymbol{C}(\boldsymbol{\theta})$, using block-diagonal approximation, diagonal approximation, and its lower bound.
Additionally, as noted in Equation \ref{eq/h_g_f}, to speed up the computation, we also estimate TIC using $\boldsymbol{F}(\boldsymbol{\theta})$ as an alternative to $\boldsymbol{H}(\boldsymbol{\theta})$, since $\boldsymbol{F}(\boldsymbol{\theta})$ and $\boldsymbol{C}(\boldsymbol{\theta})$ share common computational elements.

\begin{rem}
It should be mentioned that the above five settings are different from the situation of the NTK, since $d \ll n$.
However, we observed that the estimation of the TIC was effective for LNNs.
\end{rem}

We will first discuss the results of our experiments with respect to the quality of the approximations.
In general, from the exact computation to the block-diagonal approximation, i.e., the approximation that ignores the correlation between layers, we can confirm that the value and the rank correlation are maintained.
As for the LNN, the rank correlation is maintained for the block-diagonal approximation, the diagonal approximation, and its lower bound, though the value fluctuates.
On the other hand, in the case of NN w/ SC, we confirmed that the rank correlation is maintained between the exact calculation and the block-diagonal approximation, and between the diagonal approximation and its lower bound.
These results show that LNNs and NNs with more layers and SC tend to yield higher approximation quality.

With regard to the correlation between the TIC estimates and the generalization gap,
we observed that LNN is in the effective regime of the TIC and has a high correlation with the generalization gap for all of the approximations.
For the NNs, similarly high correlations were observed for the models w/ SC.
For the 3-NN w/o SC approach, an inverse correlation was observed even in the exact case.
In the case of 2-NN, the approximate correlation also collapsed, resulting in no correlation with the generalization gap.

From these results, we conclude that the performance of the TIC with respect to the correlation with the generalization gap is higher for the NN models with more layers and SC, and that the correlation does not change significantly  before and after the approximation.

\subsection{Practical Scale Experiments: Correlation to the Generalization Gap and TIC Lower Bound, TIC with Diagonal Approximation}

\begin{figure}[t]
\centering
    \begin{minipage}{0.45\linewidth}
    	\centering
    	\includegraphics[width=\linewidth]{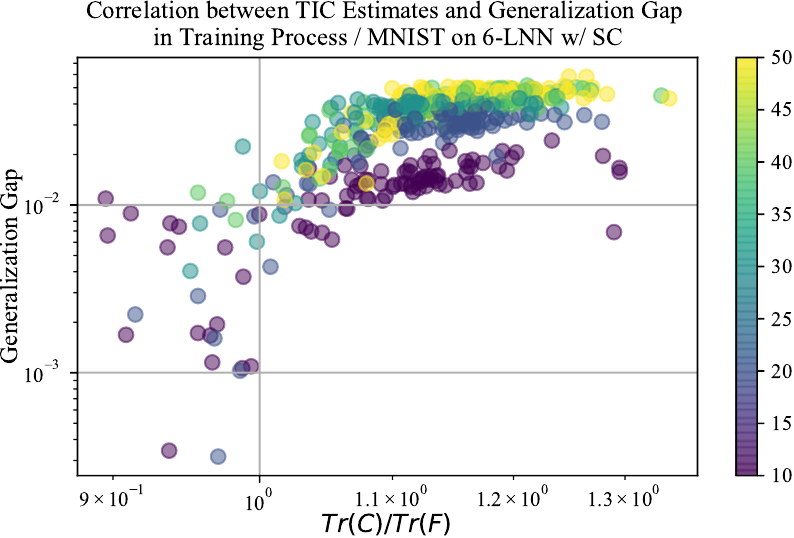}    
    	\subcaption{\scriptsize{: MNIST on 6-LNN}}
    	\label{fig:tinymnist_fast_tic_f}
    \end{minipage}
    \begin{minipage}{0.45\linewidth}
    	\centering
    	\includegraphics[width=\linewidth]{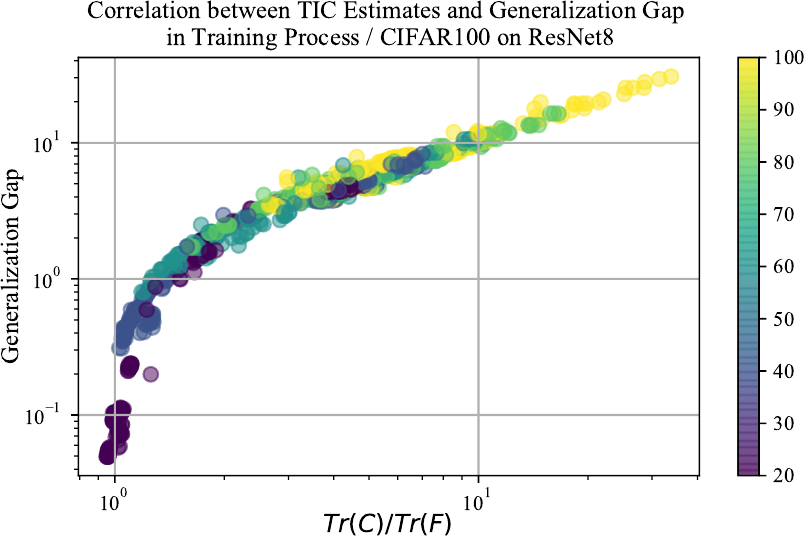} 
    	\subcaption{\scriptsize{: CIFAR100 on ResNet-8}}
    	\label{fig:mnist_fast_tic_f}
    \end{minipage}

\caption{\textbf{Correlation between the generalization gap and the TIC estimates in the training process}. The color bar represents the number of trained model epochs. Full results are shown in Appendix \ref{appendix_medium_scale}.}
\label{fig:main_correlation_in_training}
\end{figure}

In our \textbf{practical-scale} experiments, we focused on problems in which $d \gg n$, which is a necessary (though not sufficient) condition for proximity to the NTK regime (see Remark~\ref{rem:ntk_indicator}).
Here, we used the MNIST, CIFAR10, and CIFAR100 datasets for our evaluations.

We will first give the results for the MNIST case.
The LNN settings showed a strong correlation with the generalization gap in the lower bound approximation, as was the case in the small-scale experiment.
In the case of the NN model, a strong correlation with the generalization gap was observed, unlike in the small-scale setting. 
Furthermore, in the case of NN and LNN w/ SC, there was less variance and a stronger correlation with the generalization gap.
In the Simple CNN case, the correlation with the generalization gap was weaker than in the previous cases, but there was still a correlation.
Moreover, there was no correlation with the generalization gap when using $\mathrm{Tr}\left(\boldsymbol{H}(\bm{ {\theta})} \right)$, $\mathrm{Tr}\left(\boldsymbol{F}(\bm{ {\theta})} \right)$, or $\mathrm{Tr}\left(\boldsymbol{C}(\bm{ {\theta})} \right)$ individually. Detailed experimental results are shown in Figure \ref{fig:medium2} in Appendix \ref{appendix_medium_scale}.

In the cases of CIFAR10 and CIFAR100, both measures using the lower bound and diagonal approximation showed a high correlation with the generalization gap.
For LNNs, the correlation was more linear in the w/ SC case.
For VGG16 and ResNet8, the correlation was not as strong as for LNN; however, we confirmed the effectiveness of TIC in the NTK regime.
Additionally, no correlation was found between the generalization gap and the trace itself.
The trace values had different patterns depending on the network, and it was found that this single factor alone is insufficient for estimating the generalization gap.

\begin{rem}
\label{rem:training_process}
The TIC estimates captured the trend of the generalization gap throughout the training process, as shown in Figure \ref{fig:main_correlation_in_training}.
This property is important for practical applications such as early stopping and HPO trial pruning (Section~\ref{sec:hpo}), where intermediate evaluations of model quality are needed before training is complete.
Unlike the validation loss, which can be unstable due to the randomness of the holdout split, TIC provides a principled correction for the optimistic bias of the training loss and is asymptotically equivalent to LOOCV (Appendix~\ref{appendix:loocv}).
\end{rem}

\subsection{Runtime Measurement Experiments}

Our runtime measurement experiments were run on NVIDIA Tesla V100 16 GB GPUs, with an average of 10 trials each.
Significant speedup was achieved by approximating the matrix structure, replacing $\boldsymbol{H}(\boldsymbol{\theta})$ with $\boldsymbol{F}(\boldsymbol{\theta})$, and using Monte Carlo estimation of $\boldsymbol{F}(\boldsymbol{\theta})$, as described in Section \ref{sec/4_0}.
Even in the small-scale problem setting, the diagonal approximation with $\boldsymbol{F}(\boldsymbol{\theta})$ and $\boldsymbol{C}(\boldsymbol{\theta})$ was 50 times faster than the exact computation, while maintaining the rank correlation with the version using $\boldsymbol{H}(\boldsymbol{\theta})$ and $\boldsymbol{C}(\boldsymbol{\theta})$.
Since the number of parameters in the small-scale setting is at most 720, and VGG-16 has 186,530 times as many parameters, the effect of reducing the computational order from $O(d^3)$ to $O(d)$ is even more significant in the large-scale setting.
The full details are shown in Appendix \ref{sec:runtime_measurements}.
The combined strategy of using $\boldsymbol{F}(\boldsymbol{\theta})$ and $\boldsymbol{C}(\boldsymbol{\theta})$ together instead of $\boldsymbol{H}(\boldsymbol{\theta})$ and $\boldsymbol{C}(\boldsymbol{\theta})$, along with the matrix approximation methods, reduces the computation time dramatically.

\section{Application to Hyperparameter Optimization}
\label{sec:hpo}

To this point, we have shown that TIC is a reasonable estimator of the generalization gap, that it is effective during the training process, and that its value can be computed efficiently.
Motivated by these findings, we employed TIC values during training to accelerate hyperparameter optimization (HPO).
HPO is essential to achieving good performance in a wide range of machine learning algorithms \citep{feurer2019hyperparameter}.
In particular, the performance of DNNs depends significantly on the selection of hyperparameters such as learning rates, weight decay, and momentum \citep{lucic2018gans,henderson2018deep,dacrema2019we}.

The Successive Halving algorithm (SHA) \citep{jamieson2016non} shows promising performance in HPO by utilizing the iterative structure of DNNs.
SHA and its extension Hyperband \citep{Li2018hyperband} are representative multi-fidelity HPO methods that prune unpromising hyperparameters at an early stage by utilizing not only the final loss but also losses in the training process.
The validation loss obtained by the holdout method is typically used as the intermediate evaluation metric for SHA.
However, the validation loss is often numerically unstable, as shown in Figure~\ref{exp/hpo_tic}.
The key question in multi-fidelity HPO is the choice of intermediate evaluation metric: it must be both predictive of final performance and cheap to compute.
Commonly used proxies include the holdout validation loss, learning curve extrapolation \citep{Domhan2015}, and low-cost performance predictors.
TIC offers an alternative that is grounded in statistical theory and accounts for the bias inherent in the training loss.

To achieve stable optimization in SHA, we applied the TIC values for the intermediate loss.
One advantage of using TIC is that it accounts for the variance through the bias term in Equation \ref{eq/tic}, which is not captured by the validation loss obtained with the holdout method.
In particular, TIC is known to be asymptotically equivalent to leave-one-out cross-validation (LOOCV)~\citep{Stone1977} and is superior to the holdout method in terms of the order of estimation error. Details are given in Appendix \ref{appendix:loocv}.
To investigate the effectiveness of using the TIC values in SHA, we conducted an experiment.
Figure~\ref{exp/hpo_tic} shows the results.
As indicated, the TIC values with the proposed approximation method are able to select the 1st top trial, whereas the traditional method (SHA + the validation loss obtained with the holdout method) selects the 3rd top trial.
We note that this HPO experiment was conducted on a single problem setting (CIFAR-10 on ResNet-8); extending this evaluation to a wider range of architectures and datasets is an important direction for future work.

{
\begin{figure}[t]
    \begin{minipage}{0.29\hsize}
      \begin{center}
      \includegraphics[width=40mm]{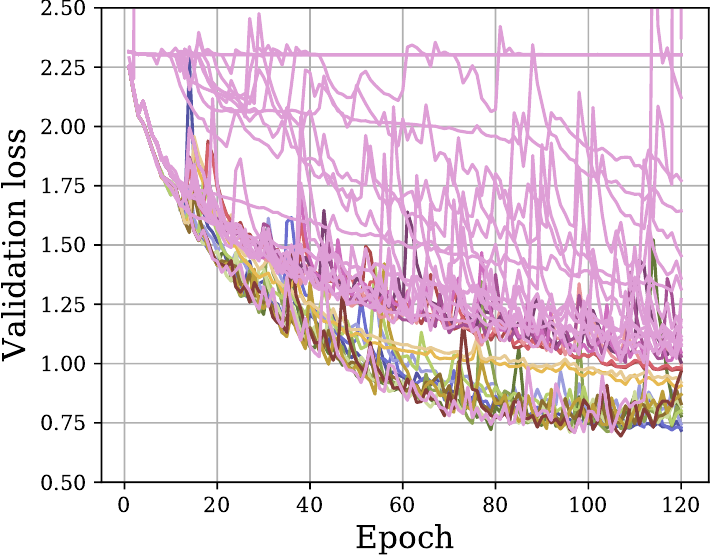}
       \subcaption{\scriptsize{: All trials without pruning}}
      \end{center}
     \end{minipage}
     \begin{minipage}{0.29\hsize}
      \begin{center}
      \includegraphics[width=40mm]{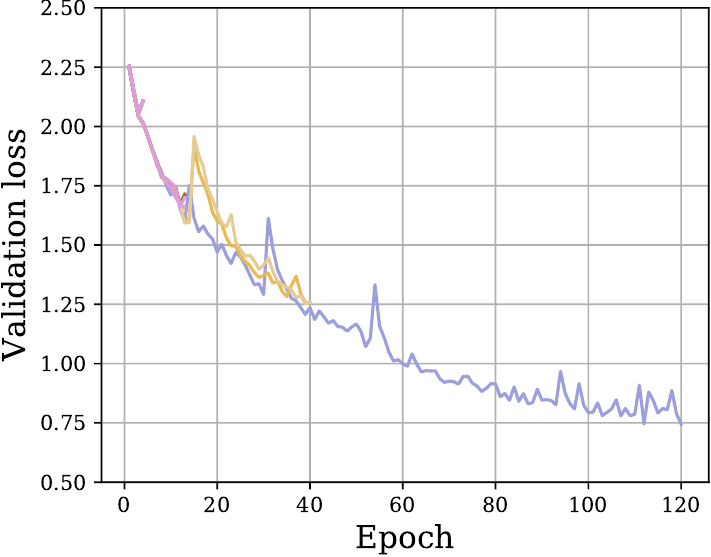}
       \subcaption{\scriptsize{: SHA pruning using validation loss}}
      \end{center}
     \end{minipage}
     \begin{minipage}{0.36\hsize}
      \begin{center}
    \includegraphics[width=52mm]{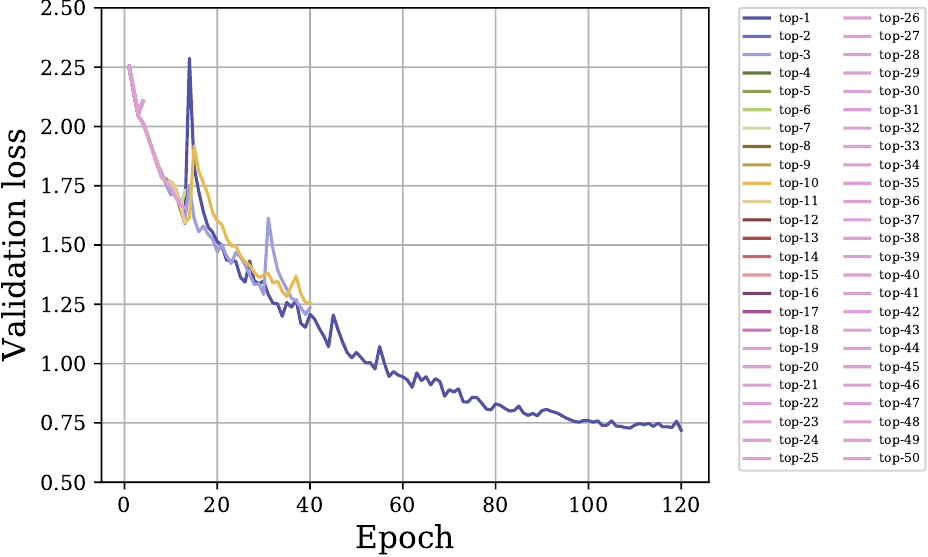}
       \subcaption{\scriptsize{: SHA pruning using TIC}}
      \end{center}
     \end{minipage}
     
     \caption{\textbf{A comparative experiment using TIC as an evaluation value for pruning with SHA in HPO for training of CIFAR10 on ResNet-8}: 
    Panel (a) shows the case in which all hyperparameter candidates are trained to the end without pruning. 
    Panel  (b) shows the case in which pruning is performed based on validation loss as a baseline. 
     Panel (c) shows the pruning method using TIC. The legends on the right-hand side show the trials with different hyperparameters;
     the final generalization performance (validation loss) to be reached is shown in descending order. 
     The 1st place trial is shown in dark purple; the 3rd place trial is in light purple.}
     \label{exp/hpo_tic}
  \end{figure}
 }

\section{Recent Related Work}
\label{sec:recent_related_work}

Since the original submission of this work (2022), several lines of research have advanced topics closely related to TIC, the NTK regime, and curvature-based model selection. We briefly survey these developments and discuss how they relate to our contributions.

\paragraph{Sharpness, curvature, and generalization revisited.}
\citet{Andriushchenko2023} conducted a systematic re-evaluation of the relationship between sharpness and generalization, showing that the correlation is more nuanced and less robust than earlier studies suggested.
In particular, they demonstrated that commonly used sharpness measures can be misleading due to scale sensitivity and reparameterization issues.
\citet{Kaur2023} provided further evidence showing that the maximum Hessian eigenvalue $\lambda_{\max}$ does not reliably predict generalization across training interventions: While higher learning rates and SAM reduce $\lambda_{\max}$, the generalization benefits vanish at larger batch sizes, and batch normalization improves generalization without consistently reducing $\lambda_{\max}$.
These findings further motivate the use of TIC, which incorporates both curvature ($\boldsymbol{H}$) and gradient noise ($\boldsymbol{C}$) through the trace $\mathrm{Tr}(\boldsymbol{H}^{-1}\boldsymbol{C})$, rather than relying on any single spectral property of the Hessian.
Relatedly, \citet{Wortsman2022} showed that averaging the weights of multiple fine-tuned models (``Model Soups'') improves accuracy without increasing inference cost, exploiting the fact that models fine-tuned from a shared initialization often lie in a connected flat basin of the loss landscape.
Although their method does not directly involve information criteria, the underlying mechanism --that the geometry of the loss landscape governs generalization---resonates with TIC's reliance on curvature $\boldsymbol{H}$ and gradient covariance $\boldsymbol{C}$ to quantify the training--test gap.

\paragraph{Scaling laws and the overparameterized regime.}
\citet{Hoffmann2022} established compute-optimal scaling laws for large language models, showing that training-data-efficient models require balancing the number of parameters $d$ and the number of training samples $n$.
Their findings highlight that practical large-scale models operate in highly overparameterized regimes with large $d/n$, a setting where, as discussed in Remark~\ref{rem:ntk_indicator}, the NTK approximation is more likely to hold and TIC is expected to be applicable.
More broadly, the scaling-law perspective situates the NTK regime within a wider $d$--$n$ tradeoff landscape, suggesting that its practical relevance may vary across different model scales.

\paragraph{Bayesian model selection and curvature computation at scale.}
\citet{Immer2021} developed a scalable marginal likelihood estimation for model selection in deep learning, demonstrating that marginal likelihood can serve as an effective criterion to compare architectures and hyperparameters.
\citet{Immer2023} further connected marginal likelihood optimization with NTK, showing that stochastic marginal likelihood gradients can be efficiently computed using the NTK structure, providing additional evidence that the NTK regime is a tractable setting for curvature-based model selection.
On the computational side, \citet{Eschenhagen2023} extended Kronecker-factored approximate curvature (KFAC) to modern architectures, including transformers, making the computation of the curvature matrix feasible for a broader class of models.
\citet{Daxberger2021} systematically evaluated post-hoc Laplace approximations across multiple inference tasks, showing that they are competitive with more expensive Bayesian approaches for uncertainty quantification and model selection while incurring minimal computational overhead.
These advances are directly relevant to TIC, as the same curvature matrices ($\boldsymbol{H}$, $\boldsymbol{F}$ and their approximations) appear both in the Laplace approximation and in the TIC estimation.

\paragraph{Functional variance and generalization gap estimation.}
\citet{Okuno2023} showed that the functional variance --the key quantity underlying the widely-applicable information criterion (WAIC) \citep{Watanabe2012} --characterizes the generalization gap even in overparameterized models where classical asymptotic theory does not apply.
They proposed the Langevin functional variance (Langevin FV), which approximates the functional variance using only first-order gradients via stochastic gradient Langevin dynamics, avoiding the expensive second-order computations required by TIC.
Their work provides a complementary computational approach to TIC: while TIC directly estimates $\mathrm{Tr}(\boldsymbol{H}^{-1}\boldsymbol{C})$, Langevin FV estimates the same underlying quantity through MCMC sampling, offering a scalable alternative for settings where the Hessian computation is infeasible.
On the numerical side, \citet{Ward2023} addressed the well-known instability of the trace term $\mathrm{Tr}(\hat{\boldsymbol{I}}^{-1}\hat{\boldsymbol{J}})$ in TIC by analyzing and extending ICE (Information Criterion by Estimation), which uses the same trace term as a regularizer during optimization rather than solely for model selection.
Ward demonstrated numerically stable approximations of the trace term and validated them on real datasets, showing that TIC and ICE can achieve practical model fitting at a reasonable computational cost when numerical instability is properly controlled.

\paragraph{Generalization bounds and their limitations.}
Following the comprehensive benchmark of generalization measures by \citet{Yiding2019}, \citet{Gastpar2024} proved a fundamental impossibility result: no generalization bound that depends only on the training data can be uniformly tight across all learning algorithms and distributions in the overparameterized setting.
This theoretical limitation contextualizes why TIC, like other training-data-dependent measures, may not predict generalization in certain regimes.
On a more positive note, \citet{Lotfi2022} showed that PAC-Bayes compression bounds based on quantizing neural network parameters in a linear subspace can produce non-vacuous generalization bounds that are tight enough to explain generalization across a variety of tasks including transfer learning.
Their compression-based approach is conceptually complementary to TIC: while TIC estimates the generalization gap via the curvature--noise interaction $\mathrm{Tr}(\boldsymbol{H}^{-1}\boldsymbol{C})$, compression bounds quantify generalization through the minimum description length of the learned parameters.
Most recently, \citet{Nakai2026} extended the benchmark of \citet{Yiding2019} along several axes: evaluating over 40 generalization measures---including information criteria and calibration-based metrics---across 10{,}000 hyperparameter configurations under both IID and out-of-distribution (OOD) settings.
Their study revealed that no single measure is universally predictive, and that the relative ranking of measure families can shift or even reverse depending on the distribution shift and architecture.
Notably, they found that information-criteria-based and calibration-based measures, which often exhibit negligible predictive power under IID evaluation, can become highly predictive under distribution shift, highlighting the importance of evaluating generalization measures beyond the standard IID setting.

\paragraph{Singular learning theory and complexity measures.}
\citet{Lau2023} introduced the local learning coefficient (LLC), a complexity measure for deep neural networks grounded in singular learning theory \citep{Watanabe2009}.
The LLC generalizes the real log canonical threshold (RLCT) to arbitrary minima in parameter space, providing a singularity-aware characterization of model complexity that accounts for the degenerate geometry of neural network loss landscapes.
Although TIC measures the effective degrees of freedom through $\mathrm{Tr}(\boldsymbol{H}^{-1}\boldsymbol{C})$ under regularity conditions that hold near the NTK regime, LLC characterizes complexity through the scaling exponent of the log marginal likelihood ($\lambda \log n$) near singularities, making it applicable to singular models where the Fisher information matrix is degenerate.
Investigating the relationship between these two families of generalization measures--information-criteria-based (TIC, WAIC) and singularity-based (RLCT, LLC)--is a promising direction for future work.

\paragraph{Multi-fidelity hyperparameter optimization.}
\citet{Kadra2023} showed that predictable power-law relationships (scaling laws) can guide efficient resource allocation in multi-fidelity HPO, connecting the scaling behavior of validation metrics to the early stopping decisions made by algorithms such as Successive Halving and Hyperband.
This perspective complements our use of TIC as an intermediate evaluation metric for trial pruning (Section~\ref{sec:hpo}): while scaling laws exploit the predictability of the learning curve, TIC provides a theoretically grounded alternative that accounts for the bias inherent in training loss.

\section{Conclusion and Discussion}
In this study, we provided a theoretical basis for the applicability of TIC to DNNs through the NTK regime and conducted comprehensive experiments with more than 5,000 models to verify this.
Our results show that the TIC approximation methods effectively capture the generalization gap in practical DNN settings that are close to the NTK regime, while the correlation diminishes outside this regime.
Furthermore, we showed that TIC can track the generalization gap during the training process, even before the model is fully trained, and demonstrated its utility as an evaluation criterion for trial pruning in HPO.

Several directions remain for future work.
First, extending the theory to the feature learning regime (outside the NTK regime) is an important open problem.
In this regime, the empirical NTK changes substantially during training \citep{Fort2020}, and the regularity conditions underlying TIC no longer hold.
In singular learning theory \citep{Watanabe2009}, the real log canonical threshold (RLCT) characterizes the asymptotic behavior of the free energy and provides a natural generalization measure for singular models.
Investigating the relationship between TIC and RLCT, as well as developing computationally tractable approximations of WAIC \citep{Watanabe2012} for large-scale DNNs, would help bridge the gap between information criteria and singular learning theory.
Recent advances in scalable Laplace approximations \citep{Daxberger2021} offer a promising computational pathway for posterior-based model selection.
Second, a systematic comparison of TIC with other generalization measures, including the sharpness-based measures revisited by \citet{Foret2021} and the comprehensive benchmark of \citet{Yiding2019}, would clarify the relative strengths of TIC under various training conditions.
Third, our experimental conditions (no batch normalization, no data augmentation) were chosen to control for NTK proximity.
Extending the evaluation to modern training recipes with these techniques, as well as to contemporary architectures such as Vision Transformers, would clarify the practical scope of applicability.
Fourth, our HPO experiments were conducted on a single problem setting (CIFAR-10 on ResNet-8).
Evaluating TIC-based trial pruning across a wider range of architectures, datasets, and multi-fidelity HPO algorithms (such as Hyperband \citep{Li2018hyperband}) would strengthen the practical conclusions.
Finally, a more refined characterization of NTK proximity, using diagnostics such as the linearization gap $\|f_t - f_t^{\mathrm{lin}}\|$ or the change in the empirical NTK during training, would provide a more principled criterion for when TIC can be expected to be effective.

\section*{Acknowledgments}
We thank the anonymous reviewers for their constructive feedback on the original submission, which helped improve the paper.
The computation resource of this project is supported by ABCI\footnote{\url{https://abci.ai/}}.

\bibliography{iclr2022_conference}

@book{Watanabe2009,
  title={Algebraic geometry and statistical learning theory},
  author={Watanabe, Sumio},
  year={2009},
  publisher={Cambridge University Press}
}

@inproceedings{Fort2020,
  title={Deep learning versus kernel learning: an empirical study of loss landscape geometry and the time evolution of the Neural Tangent Kernel},
  author={Fort, Stanislav and Dziugaite, Gintare Karolina and Paul, Mansheej and Kharaghani, Sepideh and Roy, Daniel M. and Ganguli, Surya},
  booktitle={Advances in Neural Information Processing Systems},
  volume={33},
  year={2020}
}

@inproceedings{Yang2021,
  title={Tensor Programs {IV}: Feature Learning in Infinite-Width Neural Networks},
  author={Yang, Greg and Hu, Edward J.},
  booktitle={International Conference on Machine Learning},
  year={2021}
}

@inproceedings{Foret2021,
  title={Sharpness-Aware Minimization for Efficiently Improving Generalization},
  author={Foret, Pierre and Kleiner, Ariel and Mobahi, Hossein and Neyshabur, Behnam},
  booktitle={International Conference on Learning Representations},
  year={2021}
}

@inproceedings{Daxberger2021,
  title={Laplace Redux -- Effortless {B}ayesian Deep Learning},
  author={Daxberger, Erik and Kristiadi, Agustinus and Immer, Alexander and Eschenhagen, Runa and Bauer, Matthias and Hennig, Philipp},
  booktitle={Advances in Neural Information Processing Systems},
  volume={34},
  year={2021}
}

@inproceedings{Li2018hyperband,
  title={Hyperband: A Novel Bandit-Based Approach to Hyperparameter Optimization},
  author={Li, Lisha and Jamieson, Kevin and DeSalvo, Giulia and Rostamizadeh, Afshin and Talwalkar, Ameet},
  booktitle={Journal of Machine Learning Research},
  volume={18},
  number={185},
  pages={1--52},
  year={2018}
}

@inproceedings{Domhan2015,
  title={Speeding Up Automatic Hyperparameter Optimization of Deep Neural Networks by Extrapolation of Learning Curves},
  author={Domhan, Tobias and Springenberg, Jost Tobias and Hutter, Frank},
  booktitle={International Joint Conference on Artificial Intelligence},
  year={2015}
}

@inproceedings{Hoffmann2022,
  title={Training Compute-Optimal Large Language Models},
  author={Jordan Hoffmann and Sebastian Borgeaud and Arthur Mensch and Elena Buchatskaya and Trevor Cai and Eliza Rutherford and Diego de Las Casas and Lisa Anne Hendricks and Johannes Welbl and Aidan Clark and Tom Hennigan and Eric Noland and Katie Millican and George van den Driessche and Bogdan Damoc and Aurelia Guy and Simon Osindero and Karen Simonyan and Erich Elsen and Jack W. Rae and Oriol Vinyals and Laurent Sifre},
  booktitle={Advances in Neural Information Processing Systems},
  volume={35},
  year={2022}
}

@inproceedings{Wortsman2022,
  title = 	 {Model soups: averaging weights of multiple fine-tuned models improves accuracy without increasing inference time},
  author =       {Wortsman, Mitchell and Ilharco, Gabriel and Gadre, Samir Ya and Roelofs, Rebecca and Gontijo-Lopes, Raphael and Morcos, Ari S and Namkoong, Hongseok and Farhadi, Ali and Carmon, Yair and Kornblith, Simon and Schmidt, Ludwig},
  booktitle = 	 {Proceedings of the 39th International Conference on Machine Learning},
  pages = 	 {23965--23998},
  year = 	 {2022},
  editor = 	 {Chaudhuri, Kamalika and Jegelka, Stefanie and Song, Le and Szepesvari, Csaba and Niu, Gang and Sabato, Sivan},
  volume = 	 {162},
  series = 	 {Proceedings of Machine Learning Research},
  month = 	 {17--23 Jul},
  publisher =    {PMLR},
  url = 	 {https://proceedings.mlr.press/v162/wortsman22a.html}
}

@inproceedings{Andriushchenko2023,
  title={A Modern Look at the Relationship between Sharpness and Generalization}, 
      author={Maksym Andriushchenko and Francesco Croce and Maximilian Müller and Matthias Hein and Nicolas Flammarion},
  booktitle={International Conference on Machine Learning},
  year={2023}
}

@inproceedings{Kadra2023,
  title={Scaling Laws for Hyperparameter Optimization}, 
      author={Arlind Kadra and Maciej Janowski and Martin Wistuba and Josif Grabocka},
  booktitle={Advances in Neural Information Processing Systems},
  volume={36},
  year={2023}
}

@inproceedings{Immer2023,
  title={Stochastic Marginal Likelihood Gradients using Neural Tangent Kernels}, 
      author={Alexander Immer and Tycho F. A. van der Ouderaa and Mark van der Wilk and Gunnar Rätsch and Bernhard Schölkopf},
  booktitle={International Conference on Machine Learning},
  year={2023}
}

@inproceedings{Immer2021,
  title={Scalable Marginal Likelihood Estimation for Model Selection in Deep Learning}, 
      author={Alexander Immer and Matthias Bauer and Vincent Fortuin and Gunnar Rätsch and Mohammad Emtiyaz Khan},
  booktitle={International Conference on Machine Learning},
  year={2021}
}

@inproceedings{Eschenhagen2023,
  title={Kronecker-Factored Approximate Curvature for Modern Neural Network Architectures},
author={Runa Eschenhagen and Alexander Immer and Richard E Turner and Frank Schneider and Philipp Hennig},
booktitle={Thirty-seventh Conference on Neural Information Processing Systems},
year={2023},
url={https://openreview.net/forum?id=Ex3oJEKS53}
}

@inproceedings{Lau2023,
title={The Local Learning Coefficient: A Singularity-Aware Complexity Measure},
author={Edmund Lau and Zach Furman and George Wang and Daniel Murfet and Susan Wei},
booktitle={The 28th International Conference on Artificial Intelligence and Statistics},
year={2025},
url={https://openreview.net/forum?id=1av51ZlsuL}
}

@inproceedings{Gastpar2024,
title={Fantastic Generalization Measures are Nowhere to be Found},
author={Michael Gastpar and Ido Nachum and Jonathan Shafer and Thomas Weinberger},
booktitle={The Twelfth International Conference on Learning Representations},
year={2024},
url={https://openreview.net/forum?id=NkmJotfL42}
}

@article{Okuno2023,
      title={A generalization gap estimation for overparameterized models via the Langevin functional variance}, 
      author={Akifumi Okuno and Keisuke Yano},
      year={2023},
      eprint={2112.03660},
      archivePrefix={arXiv},
      primaryClass={stat.ML},
      url={https://arxiv.org/abs/2112.03660}, 
}

@inproceedings{Lotfi2022,
title={{PAC}-Bayes Compression Bounds So Tight That They Can Explain Generalization},
author={Sanae Lotfi and Marc Anton Finzi and Sanyam Kapoor and Andres Potapczynski and Micah Goldblum and Andrew Gordon Wilson},
booktitle={Advances in Neural Information Processing Systems},
editor={Alice H. Oh and Alekh Agarwal and Danielle Belgrave and Kyunghyun Cho},
year={2022},
url={https://openreview.net/forum?id=o8nYuR8ekFm}
}

@inproceedings{Kaur2023,
  title = 	 {On the Maximum Hessian Eigenvalue and Generalization},
  author =       {Kaur, Simran and Cohen, Jeremy and Lipton, Zachary Chase},
  booktitle = 	 {Proceedings on "I Can't Believe It's Not Better!  - Understanding Deep Learning Through Empirical Falsification" at NeurIPS 2022 Workshops},
  pages = 	 {51--65},
  year = 	 {2023},
  editor = 	 {Antorán, Javier and Blaas, Arno and Feng, Fan and Ghalebikesabi, Sahra and Mason, Ian and Pradier, Melanie F. and Rohde, David and Ruiz, Francisco J. R. and Schein, Aaron},
  volume = 	 {187},
  series = 	 {Proceedings of Machine Learning Research},
  month = 	 {03 Dec},
  publisher =    {PMLR},
  url = 	 {https://proceedings.mlr.press/v187/kaur23a.html},
}

@article{Nakai2026,
  title={Revisiting Generalization Measures Beyond {IID}: An Empirical Study under Distributional Shift},
  author={Nakai, Sora and Fadhloun, Youssef and Mathlouthi, Kacem and Yoshida, Kotaro and Talluri, Ganesh and Mitliagkas, Ioannis and Naganuma, Hiroki},
  journal={arXiv preprint arXiv:2602.01718},
  year={2026}
}

@article{Ward2023,
  title={Improving the Performance and Stability of {TIC} and {ICE}},
  author={Ward, Tyler},
  journal={Entropy},
  volume={25},
  number={3},
  pages={512},
  year={2023}
}

@article{Neyshabur2014,
  title={In search of the real inductive bias: On the role of implicit regularization in deep learning},
  author={Neyshabur, Behnam and Tomioka, Ryota and Srebro, Nathan},
  journal={arXiv preprint arXiv:1412.6614},
  year={2014}
}

@article{Zhang2016,
  title={Understanding deep learning requires rethinking generalization},
  author={Zhang, Chiyuan and Bengio, Samy and Hardt, Moritz and Recht, Benjamin and Vinyals, Oriol},
  journal={arXiv preprint arXiv:1611.03530},
  year={2016}
}

@inproceedings{Recht2019,
  title={Do imagenet classifiers generalize to imagenet?},
  author={Recht, Benjamin and Roelofs, Rebecca and Schmidt, Ludwig and Shankar, Vaishaal},
  booktitle={International Conference on Machine Learning},
  pages={5389--5400},
  year={2019},
  organization={PMLR}
}

@inproceedings{Arora2018,
  title={Stronger generalization bounds for deep nets via a compression approach},
  author={Arora, Sanjeev and Ge, Rong and Neyshabur, Behnam and Zhang, Yi},
  booktitle={International Conference on Machine Learning},
  pages={254--263},
  year={2018},
  organization={PMLR}
}

@article{Wei2019,
  title={Data-dependent sample complexity of deep neural networks via lipschitz augmentation},
  author={Wei, Colin and Ma, Tengyu},
  journal={arXiv preprint arXiv:1905.03684},
  year={2019}
}

@inproceedings{Neyshabur2018,
  title={A PAC-Bayesian Approach to Spectrally-Normalized Margin Bounds for Neural Networks},
  author={Neyshabur, Behnam and Bhojanapalli, Srinadh and Srebro, Nathan},
  booktitle={International Conference on Learning Representations},
  year={2018}
}

@article{Keskar2016,
  title={On large-batch training for deep learning: Generalization gap and sharp minima},
  author={Keskar, Nitish Shirish and Mudigere, Dheevatsa and Nocedal, Jorge and Smelyanskiy, Mikhail and Tang, Ping Tak Peter},
  journal={arXiv preprint arXiv:1609.04836},
  year={2016}
}

@article{hutchinson1989stochastic,
  title={A stochastic estimator of the trace of the influence matrix for Laplacian smoothing splines},
  author={Hutchinson, Michael F},
  journal={Communications in Statistics-Simulation and Computation},
  volume={18},
  number={3},
  pages={1059--1076},
  year={1989},
  publisher={Taylor \& Francis}
}

@inproceedings{martens2015optimizing,
  title={Optimizing neural networks with kronecker-factored approximate curvature},
  author={Martens, James and Grosse, Roger},
  booktitle={International conference on machine learning},
  pages={2408--2417},
  year={2015},
  organization={PMLR}
}

@inproceedings{Yiding2019,
  title={Fantastic Generalization Measures and Where to Find Them},
  author={Jiang, Yiding and Neyshabur, Behnam and Mobahi, Hossein and Krishnan, Dilip and Bengio, Samy},
  booktitle={International Conference on Learning Representations},
  year={2019}
}

@inproceedings{Novak2018,
  title={Sensitivity and Generalization in Neural Networks: an Empirical Study},
  author={Novak, Roman and Bahri, Yasaman and Abolafia, Daniel A and Pennington, Jeffrey and Sohl-Dickstein, Jascha},
  booktitle={International Conference on Learning Representations},
  year={2018}
}

@inproceedings{Thomas2019,
  title={On the interplay between noise and curvature and its effect on optimization and generalization},
  author={Thomas, Valentin and Pedregosa, Fabian and Merri{\"e}nboer, Bart and Manzagol, Pierre-Antoine and Bengio, Yoshua and Le Roux, Nicolas},
  booktitle={International Conference on Artificial Intelligence and Statistics},
  pages={3503--3513},
  year={2020},
  organization={PMLR}
}

@article{Takeuchi1976,
  title={Distribution of information statistic and validity criterion of models,"},
  author={Takeuchi, Kei},
  journal={Mathematical Science},
  number={153},
  pages={12--18},
  year={1976}
}

@inproceedings{Arthur2018,
  title={Neural tangent kernel: convergence and generalization in neural networks},
  author={Jacot, Arthur and Gabriel, Franck and Hongler, Cl{\'e}ment},
  booktitle={Proceedings of the 32nd International Conference on Neural Information Processing Systems},
  pages={8580--8589},
  year={2018}
}

@inproceedings{McAllester1999,
  title={PAC-Bayesian model averaging},
  author={McAllester, David A},
  booktitle={Proceedings of the twelfth annual conference on Computational learning theory},
  pages={164--170},
  year={1999}
}

@incollection{Vapnik1971,
  title={On the uniform convergence of relative frequencies of events to their probabilities},
  author={Vapnik, Vladimir N and Chervonenkis, A Ya},
  booktitle={Measures of complexity},
  pages={11--30},
  year={2015},
  publisher={Springer}
}

@article{martens2020new,
  title={New Insights and Perspectives on the Natural Gradient Method},
  author={Martens, James},
  journal={Journal of Machine Learning Research},
  volume={21},
  pages={1--76},
  year={2020}
}

@inproceedings{Neyshabur2015b,
  title={Norm-based capacity control in neural networks},
  author={Neyshabur, Behnam and Tomioka, Ryota and Srebro, Nathan},
  booktitle={Conference on Learning Theory},
  pages={1376--1401},
  year={2015},
  organization={PMLR}
}

@article{Watanabe2012,
  title={A widely applicable Bayesian information criterion},
  author={Watanabe, Sumio},
  journal={Journal of Machine Learning Research},
  volume={14},
  number={Mar},
  pages={867--897},
  year={2013}
}

@article{Konishi1996,
  title={Generalised information criteria in model selection},
  author={Konishi, Sadanori and Kitagawa, Genshiro},
  journal={Biometrika},
  volume={83},
  number={4},
  pages={875--890},
  year={1996},
  publisher={Oxford University Press}
}

@inproceedings{Liang2017,
  title={Fisher-rao metric, geometry, and complexity of neural networks},
  author={Liang, Tengyuan and Poggio, Tomaso and Rakhlin, Alexander and Stokes, James},
  booktitle={The 22nd International Conference on Artificial Intelligence and Statistics},
  pages={888--896},
  year={2019},
  organization={PMLR}
}

@inproceedings{Bartlett2017,
  title={Spectrally-normalized margin bounds for neural networks},
  author={Bartlett, Peter L and Foster, Dylan J and Telgarsky, Matus},
  booktitle={Proceedings of the 31st International Conference on Neural Information Processing Systems},
  pages={6241--6250},
  year={2017}
}

@article{Moody1992,
  title={The Effective Number of Parameters: An Analysis of Generalization and Regularization in Nonlinear Learning Systems'. In JE Moody, SJ Hanson and RP Lippmann (eds.), Advances in Neural Information Processing Systems 4. San Mateo, CA: Morgan Kauffmann Publishers},
  author={Moody, J},
  journal={Neural Information Processing Systems 4},
  year={1992}
}

@article{Sidak2020,
  title={WoodFisher: Efficient second-order approximations for model compression},
  author={Singh, Sidak Pal and Alistarh, Dan},
  journal={arXiv preprint arXiv:2004.14340},
  year={2020}
}

@article{Stone1977,
  title={An asymptotic equivalence of choice of model by cross-validation and Akaike's criterion},
  author={Stone, Mervyn},
  journal={Journal of the Royal Statistical Society: Series B (Methodological)},
  volume={39},
  number={1},
  pages={44--47},
  year={1977},
  publisher={Wiley Online Library}
}

@article{White1982,
  title={Maximum likelihood estimation of misspecified models},
  author={White, Halbert},
  journal={Econometrica: Journal of the econometric society},
  pages={1--25},
  year={1982},
  publisher={JSTOR}
}

@inproceedings{Lee2018,
  title={Deep Neural Networks as Gaussian Processes},
  author={Lee, Jaehoon and Bahri, Yasaman and Novak, Roman and Schoenholz, Samuel S and Pennington, Jeffrey and Sohl-Dickstein, Jascha},
  booktitle={International Conference on Learning Representations},
  year={2018}
}

@inproceedings{Novak2020,
  title={Bayesian Deep Convolutional Networks with Many Channels are Gaussian Processes},
  author={Novak, Roman and Xiao, Lechao and Bahri, Yasaman and Lee, Jaehoon and Yang, Greg and Hron, Jiri and Abolafia, Daniel A and Pennington, Jeffrey and Sohl-dickstein, Jascha},
  booktitle={International Conference on Learning Representations},
  year={2018}
}

@article{Lee2019,
  title={Wide neural networks of any depth evolve as linear models under gradient descent},
  author={Lee, Jaehoon and Xiao, Lechao and Schoenholz, Samuel S and Bahri, Yasaman and Novak, Roman and Sohl-Dickstein, Jascha and Pennington, Jeffrey},
  journal={arXiv preprint arXiv:1902.06720},
  year={2019}
}

@article{Arora2019,
  title={On exact computation with an infinitely wide neural net},
  author={Arora, Sanjeev and Du, Simon S and Hu, Wei and Li, Zhiyuan and Salakhutdinov, Ruslan and Wang, Ruosong},
  journal={arXiv preprint arXiv:1904.11955},
  year={2019}
}

@article{Yang2019,
  title={Scaling limits of wide neural networks with weight sharing: Gaussian process behavior, gradient independence, and neural tangent kernel derivation},
  author={Yang, Greg},
  journal={arXiv preprint arXiv:1902.04760},
  year={2019}
}

@inproceedings{Amari2020,
  title={When does preconditioning help or hurt generalization?},
  author={Amari, Shun-ichi and Ba, Jimmy and Grosse, Roger Baker and Li, Xuechen and Nitanda, Atsushi and Suzuki, Taiji and Wu, Denny and Xu, Ji},
  booktitle={International Conference on Learning Representations},
  year={2020}
}

@inproceedings{Martens2015,
  title={Optimizing neural networks with kronecker-factored approximate curvature},
  author={Martens, James and Grosse, Roger},
  booktitle={International conference on machine learning},
  pages={2408--2417},
  year={2015},
  organization={PMLR}
}

@article{Yeming2019,
  title={Interplay between optimization and generalization of stochastic gradient descent with covariance noise},
  author={Wen, Yeming and Luk, Kevin and Gazeau, Maxime and Zhang, Guodong and Chan, Harris and Ba, Jimmy},
  journal={arXiv preprint arXiv:1902.08234},
  year={2019}
}

@misc{Zhanxing2019,
      title={The Anisotropic Noise in Stochastic Gradient Descent: Its Behavior of Escaping from Sharp Minima and Regularization Effects}, 
      author={Zhanxing Zhu and Jingfeng Wu and Bing Yu and Lei Wu and Jinwen Ma},
      year={2019},
      eprint={1803.00195},
      archivePrefix={arXiv},
      primaryClass={stat.ML}
}

@inproceedings{Zhang2017,
  title={Noisy natural gradient as variational inference},
  author={Zhang, Guodong and Sun, Shengyang and Duvenaud, David and Grosse, Roger},
  booktitle={International Conference on Machine Learning},
  pages={5852--5861},
  year={2018},
  organization={PMLR}
}

@inproceedings{Roux2008,
  title={Topmoumoute Online Natural Gradient Algorithm.},
  author={Le Roux, Nicolas and Manzagol, Pierre-Antoine and Bengio, Yoshua},
  booktitle={NIPS},
  pages={849--856},
  year={2007},
  organization={Citeseer}
}

@inproceedings{DR17,
        title = {Computing Nonvacuous Generalization Bounds for Deep (Stochastic) Neural Networks with Many More Parameters than Training Data},
       author = {Gintare Karolina Dziugaite and Daniel M. Roy},
         year = {2017},
    booktitle = {Proceedings of the 33rd Annual Conference on Uncertainty in Artificial Intelligence (UAI)},
archivePrefix = {arXiv},
       eprint = {1703.11008},
}

@article{Karakida2020,
  title={Understanding Approximate Fisher Information for Fast Convergence of Natural Gradient Descent in Wide Neural Networks},
  author={Karakida, Ryo and Osawa, Kazuki},
  journal={arXiv preprint arXiv:2010.00879},
  year={2020}
}

@article{Kunstner2020,
  title={Limitations of the empirical fisher approximation for natural gradient descent},
  author={Kunstner, Frederik and Balles, Lukas and Hennig, Philipp},
  journal={arXiv preprint arXiv:1905.12558},
  year={2019}
}

@article{Schraudolph2002,
  title={Fast curvature matrix-vector products for second-order gradient descent},
  author={Schraudolph, Nicol N},
  journal={Neural computation},
  volume={14},
  number={7},
  pages={1723--1738},
  year={2002},
  publisher={MIT Press}
}

@article{Pearlmutter1994,
  title={Fast exact multiplication by the Hessian},
  author={Pearlmutter, Barak A},
  journal={Neural computation},
  volume={6},
  number={1},
  pages={147--160},
  year={1994},
  publisher={MIT Press}
}

@article{Avron2011,
  title={Randomized algorithms for estimating the trace of an implicit symmetric positive semi-definite matrix},
  author={Avron, Haim and Toledo, Sivan},
  journal={Journal of the ACM (JACM)},
  volume={58},
  number={2},
  pages={1--34},
  year={2011},
  publisher={ACM New York, NY, USA}
}

@incollection{feurer2019hyperparameter,
  title={{Hyperparameter Optimization}},
  author={Feurer, Matthias and Hutter, Frank},
  booktitle={Automated Machine Learning},
  pages={3--33},
  year={2019},
}

@inproceedings{lucic2018gans,
  title={{Are Gans Created Equal? A Large-Scale Study}},
  author={Lucic, Mario and Kurach, Karol and Michalski, Marcin and Gelly, Sylvain and Bousquet, Olivier},
  booktitle={Advances in neural information processing systems},
  pages={700--709},
  year={2018}
}

@inproceedings{henderson2018deep,
  title={{Deep Reinforcement Learning that Matters}},
  author={Henderson, Peter and Islam, Riashat and Bachman, Philip and Pineau, Joelle and Precup, Doina and Meger, David},
  booktitle={Thirty-Second AAAI Conference on Artificial Intelligence},
  year={2018}
}

@inproceedings{dacrema2019we,
  title={Are We Really Making Much Progress? A Worrying Analysis of Recent Neural Recommendation Approaches},
  author={Dacrema, Maurizio Ferrari and Cremonesi, Paolo and Jannach, Dietmar},
  booktitle={Proceedings of the 13th ACM Conference on Recommender Systems},
  pages={101--109},
  year={2019}
}

@inproceedings{jamieson2016non,
  title={Non-stochastic best arm identification and hyperparameter optimization},
  author={Jamieson, Kevin and Talwalkar, Ameet},
  booktitle={Artificial Intelligence and Statistics},
  pages={240--248},
  year={2016}
}

@inproceedings{dinh2017sharp,
  title={Sharp minima can generalize for deep nets},
  author={Dinh, Laurent and Pascanu, Razvan and Bengio, Samy and Bengio, Yoshua},
  booktitle={International Conference on Machine Learning},
  pages={1019--1028},
  year={2017},
  organization={PMLR}
}

@article{Choi19,
  title={On empirical comparisons of optimizers for deep learning},
  author={Choi, Dami and Shallue, Christopher J and Nado, Zachary and Lee, Jaehoon and Maddison, Chris J and Dahl, George E},
  journal={arXiv preprint arXiv:1910.05446},
  year={2019}
}

@article{Shallue19,
  author  = {Christopher J. Shallue and Jaehoon Lee and Joseph Antognini and Jascha Sohl-Dickstein and Roy Frostig and George E. Dahl},
  title   = {Measuring the Effects of Data Parallelism on Neural Network Training},
  journal = {Journal of Machine Learning Research},
  year    = {2019},
  volume  = {20},
  number  = {112},
  pages   = {1-49}
}

@inproceedings{Simonyan15,
  title={Very Deep Convolutional Networks for Large-Scale Image Recognition},
  author={Karen Simonyan and Andrew Zisserman},
  booktitle={International Conference on Learning representations},
  year={2015}
}
\bibliographystyle{colm2024_conference}

\newpage
\appendix
\section*{Appendix}

\section{Proofs}
\subsection{Derivation of the TIC in NTK Regime}
\label{sec_appendix_ntk}

\subsubsection{NTK: Neural Tangent Kernel}

In general, DNNs have a large number of parameters $p$ compared to the number of data points $n$, causing them to memorize data and adversely affecting their generalization ability.
Moreover, all data are subjected to nonlinear transformations, which results in the problem of minimizing a nonconvex objective function.
Due to these difficulties in analyzing the training dynamics of DNNs, the reasons for generalization of practical DNNs and the guarantee of global convergence remain open questions.
However, a theoretical framework called the neural tangent kernel (NTK) \citep{Arthur2018} has been developed to analyze the training dynamics of gradient descent in DNNs with sufficiently large width.

Assuming that the loss function to be minimized in NN training is $L = \frac{1}{2} \|f({x} ; \bm\theta)-{y}\|^{2}$, and
the parameters are updated by gradient descent with learning rate $\eta$, the parameter update $\Delta \bm\theta$ can be written as

\begin{equation}
    \label{eq/update}
    \Delta \bm\theta=-\eta \frac{\partial {L}(\bm\theta)}{\partial \bm\theta}=-\eta \nabla_{\bm\theta} {L}(\bm\theta)
\end{equation}

Taking the continuous-time limit (where $t$ denotes the training time), we obtain

\begin{eqnarray}
    \label{eq/continual}
    \frac{\partial \bm\theta}{\partial t}&=&-\eta \nabla_{\bm\theta} {L}(\bm\theta(t)) \\
    &=&-\eta J_{t}({x} ; \bm\theta)^{T} \nabla_{f_{t}({x})} {L}(f_{t}({x}))
\end{eqnarray}

 where $J_{t}({x} ; \bm\theta)$ is $\frac{\partial f({x} ; \bm\theta)}{\partial \bm\theta}=\nabla_{\bm\theta} f_{t}({x})$.

Next, considering the time evolution of the function output rather than the parameters, we obtain

\begin{eqnarray}
    \!\!\!\!\!\frac{\partial f_{t}({x; \bm\theta})}{\partial t} \!\!\!&=&\!\!\!\frac{\partial f_{t}({x; \bm\theta})}{\partial \bm\theta} \frac{\partial \bm\theta}{\partial t} \\
    &=&\!\!\!-\eta J_{t}({x} ; \bm\theta) J_{t}({x} ; \bm\theta)^{T} \nabla_{f_{t}({x})} {L}(f_{t}({x})) \\
    &=&\!\!\!-\eta \mathcal{K}_{t}({x}, {x}) \nabla_{f_{t}({x})} {L}(f_{t}({x}))
\end{eqnarray}

 where $\mathcal{K}_{t}({x}, {x})$ is $J_{t}({x} ; \bm\theta) J_{t}({x} ; \bm\theta)^{T}=\sum_{\text{layer}=1}^{\text{num of layer}} J_{t}\left({x} ; \bm\theta_{\text{layer}}\right) J_{t}\left({x} ; \bm\theta_{\text{layer}}\right)^{T}$

The issue is that $\mathcal{K}_{t}({x}, {x})$, the NTK at time $t$, depends on $\bm \theta$ and $x$.
However, it has been shown that if the width of the randomly initialized NN is sufficiently large,
$\mathcal{K}_{t}({x}, {x}) \approx \mathcal{K}_{0}({x}, {x})$ \citep{Arthur2018}.

Approximating the output of the neural network by its first-order Taylor expansion around the initial parameters, we have

\begin{equation}
    \label{eq:liniear}
    f_{t}^{\operatorname{lin}}\left(x ; \bm\theta_{t}\right) \approx f_{0}\left(x ; \bm\theta_{0}\right)+\nabla_{\bm\theta} f_{0}\left(x ; \bm\theta_{0}\right)^{T}\left(\bm\theta_{t}-\bm\theta_{0}\right)
\end{equation}

Under this linearization, the training dynamics can be expressed in closed form:

\begin{eqnarray}
    \label{eq:unique}
\bm\theta_{t} = \bm\theta_{0}-\nabla_{\theta} f_{0}({x})^{T} {\mathcal{K}}_{0}^{-1}\left(I-e^{-\eta {\mathcal{K}}_{0} t}\right)\left(f_{0}({x})-{y}\right) \\
f_{t}^{\operatorname{lin}}({x})=\left(I-e^{-\eta {\mathcal{K}}_{0} t}\right) {y}+e^{-\eta {\mathcal{K}}_{0} t} f_{0}({x})
\end{eqnarray}

NTK theory thus determines the dynamics of gradients in function space by introducing the NTK regime, which allows us to assume that the weights follow a Gaussian process even as training progresses. This is based on the result that a randomly initialized NN can be viewed as a Gaussian process when the hidden layer width becomes infinite \citep{Lee2018,Novak2020}.

The NTK allows us to prove the global convergence of gradient descent; furthermore, the equivalence between the trained model and the Gaussian process can be used to explain the generalization performance of DNNs.
The NTK framework has been extended to CNNs \citep{Arora2019} and RNNs \citep{Yang2019} in addition to MLPs, and exhaustive experiments have been conducted \citep{Lee2019}.

\subsubsection{Preliminaries for TIC in NTK Regime}
\label{appendix:a12}
TIC requires that the statistical model be a regular model. However, DNNs are generally singular models.
The requirements for a regular model are as follows:

{
\setlength{\leftmargini}{10pt}%
\begin{itemize}
    \setlength{\itemsep}{-3pt}      %
    \item The posterior distribution of the parameters can be approximated by a Gaussian distribution, and the number of samples is sufficiently large (as n increases, the prior distribution is ignored).
    \item There is only one optimal solution $\hat{\bm{\theta}}$ for $\argmax_{} \ell(\bm{\theta})$.
    \item ${\boldsymbol{H}_p(\bm{{\theta^{*}}})}$ is positive definite.
\end{itemize}
}

In machine learning, we often seek to minimize the negative log-likelihood, treating it as a loss function.
Let $f$ be the predictive distribution of the model parameterized by $\bm{{\theta}}\in \bm\Theta \subset \mathbb{R}^{p}$, and let $g$ be the true distribution.
We can compare models by measuring the KL divergence between $f$ and $g$ to see how well $f$ approximates $g$.

\begin{eqnarray}
  D_{KL}(g,f)\!\!\!\!&=&\!\!\!\!\mathbb{E}_p\left[\log \frac{g( y| x)}{f( y| x, \bm\theta)}\right] \\
  \!\!\!\!&=&\!\!\!\!\mathbb{E}_p [\log g( y| x)]-\mathbb{E}_p[\log f( y| x, \bm\theta)] \label{eq/kl}
\end{eqnarray}

Since the first term of Equation \ref{eq/kl} is independent of $\bm{{\theta}}$, the model is better if it maximizes the second term,
i.e., the mean log-likelihood $\mathcal{L}(\bm{\theta}) = \mathbb{E}_p[\log f( y| x, \bm\theta)]$.

The mean log-likelihood $\mathcal{L}(\bm{\theta})$ is an unknown quantity that cannot be computed directly, as it depends on the true data distribution $p$.
However, if a valid estimator of the mean log-likelihood can be obtained using the empirical distribution $\hat{p}$, it can serve as a criterion for evaluating the model.

In model selection, we consider a DNN model with output distribution $f( y| x, \bm\theta)$, where the parameters are estimated by maximum likelihood.
We obtain the fitted model $f( y| x, \bm{\hat{\theta}})$ by replacing the unknown parameters $\bm\theta$ with the maximum likelihood estimator $\bm{\hat{\theta}}$.

\begin{equation}
  \hat{\bm{\theta}}:=\argmax_{\bm{\theta} \in \Theta} \ell(\bm{\theta})
  \label{eq/map}
\end{equation}

Here $\ell(\bm{\theta}) \!\!=\!\! \mathbb{E}_{\hat{p}}[\log f( y| x, \bm\theta)] \!\!=\!\! \frac{1}{n} \sum_{i=1}^{n} \log f( y_i| x_i, \bm\theta)$ is the likelihood function over $\bm\theta \in \bm\Theta$.

Let $\boldsymbol{S_n}= \left\{ (x_{1},y_{1}) , (x_{2},y_{2}), \dots (x_{n},y_{n}) \right\} $ be the data observed according to the true data distribution $p$.
Let $\hat{p}$ be the empirical distribution based on this $\boldsymbol{S_n}$. 
By the law of large numbers,
$\mathbb{E}_{\hat{p}}[\log f( y| x, \hat{\bm{\theta}})]=\frac{1}{n} \sum_{i=1}^{n} \log f( y_i| x_i, \hat{\bm{\theta}})$
converges in probability to $\mathbb{E}_{p}[\log f( y| x, \hat{\bm{\theta}})]$ as $n \to \infty$.

\begin{equation}
  \frac{1}{n} \sum_{i=1}^{n} \log f( y_i| x_i, \hat{\bm{\theta}}) \xrightarrow[n \to +\infty]{} \mathbb{E}_{p}[\log f( y| x, \hat{\bm{\theta}})]
  \label{eq/law_of_number}
\end{equation}

Therefore, the estimator based on the empirical distribution in Equation \ref{eq/law_of_number} is a natural estimator of the mean log-likelihood.
While $\mathbb{E}_{\hat{p}}[\log f( y| x, \hat{\bm{\theta}})]$ is a natural estimator of $\mathbb{E}_{p}[\log f( y| x, \hat{\bm{\theta}})]$,
the parameter $\hat{\bm{\theta}}$ is itself estimated from the empirical data $(x_i,y_i) \sim \hat{p}$.
Since the same data are used both to estimate the parameters and to evaluate the mean log-likelihood of $f( y| x, \hat{\bm{\theta}})$, fair model selection is not possible without correction.

Thus, it is necessary to evaluate and correct for this bias to enable fair model selection.
The bias of estimating the mean log-likelihood $\mathbb{E}_{p}[\log f( y| x, \hat{\bm{\theta}})]$ with $\mathbb{E}_{\hat{p}}[\log f( y| x, \hat{\bm{\theta}})]$ is formulated as follows:

\begin{eqnarray}
  \!\!\!\!\!b\!\!\!\!\!&=&\!\!\!\!\!n \mathbb{E}_{p} \left[  \mathbb{E}_{\hat{p}}[\log f(\bm y|\bm x, \hat{\bm{\theta}})] - \mathbb{E}_{p}[\log f(\bm y|\bm x, \hat{\bm{\theta}})]   \right] \label{eq/bias1}\\
  \!\!\!\!\!&=&\!\!\!\!\! n \mathbb{E}_{p} \left[ \ell(\hat{\bm{\theta}}) - \mathcal{L}(\hat{\bm{\theta}})  \right] \label{eq/bias2}\\
  &=&\!\!\!\!\! n \mathbb{E}_{p}\left[ \ell(\hat{\bm{\theta}}) - \ell({\bm{\theta^{*}}})  \right] \quad \label{eq/biasA}\\
  &+&\!\!\!\!\! n \mathbb{E}_{p}\left[ \ell({\bm{\theta^{*}}}) - \mathcal{L}({\bm{\theta^{*}}})  \right] \quad \label{eq/biasB}\\
  &+&\!\!\!\!\! n \mathbb{E}_{p}\left[ \mathcal{L}({\bm{\theta^{*}}}) - \mathcal{L}(\hat{\bm{\theta}})  \right] \label{eq/biasC}
\end{eqnarray}

Here $\bm{\theta^{*}}$ is the maximizer of $\mathcal{L}({\bm{\theta}})$.
Thus, Equation \ref{eq/bias2} can be decomposed into Equations \ref{eq/biasA}, \ref{eq/biasB}, and \ref{eq/biasC}.
Moreover, Equation \ref{eq/biasB} converges to 0 because the expectation of $\ell(\bm{\theta}^{*})$ under $\hat{p}$ converges to $\mathcal{L}(\bm{\theta}^{*})$ by the law of large numbers. Equations \ref{eq/biasA} and \ref{eq/biasC} each converge to $\frac{1}{2}\mathrm{Tr}\left({\boldsymbol{H}}_p(\bm{{\theta^{*}})}^{-1} {\boldsymbol{D}}_p(\bm{{\theta^{*}}})\right)$ as $n\to\infty$.
This asymptotic validity holds under the regularity conditions of \citet{White1982}.

\subsubsection{Applying the NTK Regime for TIC Derivation}
\label{appendix:a13}

We can satisfy the above conditions by using the training dynamics of DNNs in the NTK regime.
Specifically, the NTK regime satisfies the first condition since it uses a locally linear approximation, as in Equation \ref{eq:liniear}, and treats the training of the NN as a Gaussian process.
In addition, as shown in Equation \ref{eq:unique}, the optimal solution at the $t$-th step is uniquely determined, and the optimization is convex.

The positive definiteness of $\mathcal{K}_{t}({x}, {x})$ is proved in Appendix A4 of \citet{Arthur2018}, under the assumption of non-polynomial Lipschitz nonlinearity.
From the definition of $\mathcal{K}_{t}({x}, {x})$, the Fisher information matrix (FIM) is positive definite in the NTK regime because $\mathcal{K}_{t}({x}, {x})$ and the FIM share the same eigenvalues through a duality relationship.

This condition is true when the DNN is considered to be in the NTK regime, i.e., when Assumption \ref{assum:1} is satisfied.

\begin{figure}[t]
  \centering
  \includegraphics[width=7cm, bb=0 0 807 705]{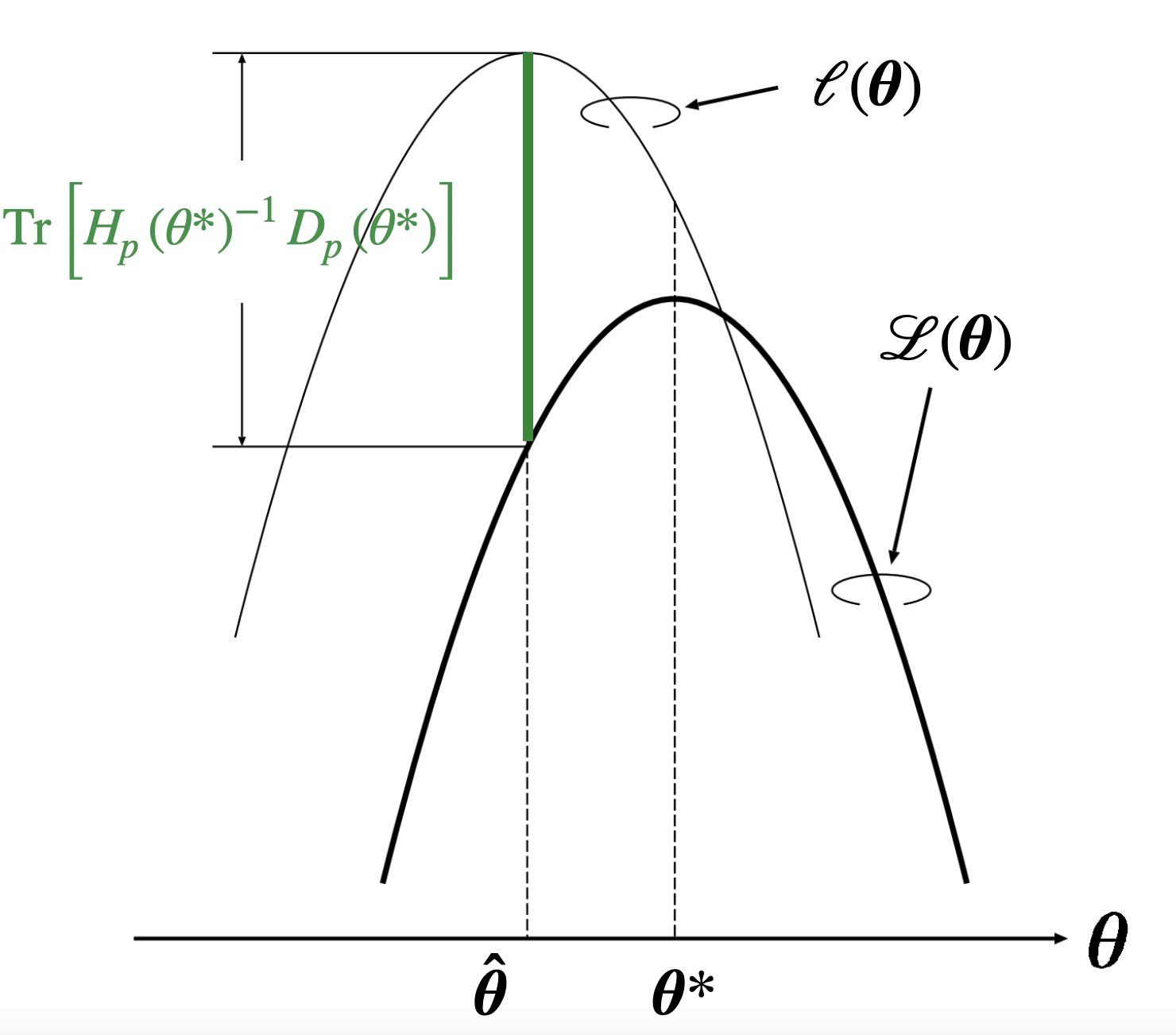}
  \caption{TIC takes into account the bias of the estimate.}
  \label{fig/tic_bias}
\end{figure}

Figure \ref{fig/tic_bias} shows a schematic diagram of the bias term $b$.
The matrices ${\boldsymbol{H}}_p(\bm{{\theta^{*}})}$ and ${\boldsymbol{D}}_p(\bm{{\theta^{*}}})$ are as follows:

\begin{equation}
  \begin{array}{rcl}
    {\boldsymbol{D}_p(\bm{{\theta^{*}}})} &=&\left. \mathbb{E}_{p}\left[ \frac{\partial \log f( y| x, {\bm{\theta}})}{\partial \boldsymbol{\theta}} \frac{\partial \log f( y| x, {\bm{\theta}})}{\partial \boldsymbol{\theta}^{T}}\right|_{\boldsymbol{\theta}={\boldsymbol{\theta^{*}}}} \right]\\
    {\boldsymbol{H}_p(\bm{{\theta^{*}}})}&=&\left. \mathbb{E}_{p}\left[ \frac{\partial^{2} \log f( y| x, {\bm{\theta}})}{\partial \boldsymbol{\theta} \partial \boldsymbol{\theta}^{T}}\right|_{\boldsymbol{\theta}={\boldsymbol{\theta^{*}}}} \right]
  \end{array}
  \label{eq/hf}
\end{equation}

If the true distribution $g$ is included in the assumed statistical model $f(\bm y|\bm x, \bm\theta)$, then ${\boldsymbol{D}_p(\bm{{\theta^{*}}})} = {\boldsymbol{H}_p(\bm{{\theta^{*}}})}$ is valid and $b = \mathrm{Tr}(\boldsymbol{I}) = d$, and thus the AIC can be derived.
In the DNN setting, this assumption does not hold, i.e., it is necessary to use the respective matrices for the misspecified situation.

Since bias $b$ depends on the true data distribution $p$, it needs to be estimated based on the observed data.
Assuming that the consistent estimators for ${\boldsymbol{D}_p(\bm{{\theta^{*}}})}$ and ${\boldsymbol{H}_p(\bm{{\theta^{*}}})}$ are ${\boldsymbol{C}(\bm{\hat{\theta}})}$ and ${\boldsymbol{H}(\bm{\hat{\theta}})}$, 
respectively, the estimates in equation \ref{eq/bias1} are as follows:

\begin{equation}
\begin{array}{rcl}
  {\boldsymbol{C}(\bm{\hat{\theta}})} &=&\left.\frac{1}{n} \sum_{i=1}^{n} \frac{\partial \log f( y_i| x_i, {\bm{\theta}})}{\partial \boldsymbol{\theta}} \frac{\partial \log f( y_i| x_i, {\bm{\theta}})}{\partial \boldsymbol{\theta}^{T}}\right|_{\boldsymbol{\theta}=\hat{\boldsymbol{\theta}}} \\
  {\boldsymbol{H}(\bm{\hat{\theta}})}&=&\left.\frac{1}{n} \sum_{i=1}^{n} \frac{\partial^{2} \log f( y_i| x_i, {\bm{\theta}})}{\partial \boldsymbol{\theta} \partial \boldsymbol{\theta}^{T}}\right|_{\boldsymbol{\theta}=\hat{\boldsymbol{\theta}}}
\end{array}
\label{eq/ch2}
\end{equation}

Using equation \ref{eq/ch},  $\hat{b}$ as an estimate of bias $b$ can be described as follows:

\begin{equation}
    \label{eq/tic_bias}
    \hat{b} = \mathrm{Tr}\left(\boldsymbol{H}(\bm{ \hat{\theta})}^{-1} \boldsymbol{C}(\bm{\hat{\theta}})\right)
\end{equation}

The term $b$ is called Moody's effective number of parameters \citep{Moody1992}.
The TIC is derived in equation \ref{eq/tic} by estimating the asymptotic bias of the mean log-likelihood with the log-likelihood of the statistical model.

\section{Asymptotic Equivalence of TIC to Cross-Validation}
\label{appendix:loocv}

Since deep learning usually requires a large amount of data and substantial training time, the holdout method is commonly used to divide the data into training data, validation data for model selection (especially hyperparameter optimization), and test data to verify model performance.
This method is relatively fast, but its evaluation varies depending on how the data are split, and it is not suitable when the dataset is small.
In $K$-fold cross-validation, the entire training set is divided into $K$ folds. One fold is used as validation data, while the remaining $K-1$ folds are used for training.
This process is repeated so that each fold serves as the validation set exactly once.
Leave-one-out cross-validation (LOOCV) is the special case where each fold consists of a single data point.
LOOCV is empirically known to yield reliable estimates and is often used when the dataset is small.
If the number of data points is $n$, the bias of the estimation error is $O(1/\sqrt{n})$ for the holdout method and $O(1/{n})$ for LOOCV~\citep{Stone1977}.
However, LOOCV requires $n$ times the computational cost.
For a dataset such as ImageNet-1K with 1.2 million training images, the current practice is to use the holdout method, which achieves $O(1/\sqrt{n})$ estimation error at low computational cost, rather than reducing the error to $O(1/{n})$ at prohibitive computational expense.

\section{TIC Experimental Details}
\label{appendix:setting}
\subsection{Implementation and Environment for Experiment}
We performed our experiments with the ABCI supercomputer. For the ABCI supercomputer,
each node is composed of NVIDIA Tesla V100$\times$4GPU and Intel Xeon Gold 6148 2.4 GHz, 20 Cores$\times$2CPU. As a software environment, we used Red Hat 4.8.5, gcc 7.4, Python 3.6.5, PyTorch 1.6.0, cuDNN 7.6.2, and CUDA 10.0.

\subsection{Hyperparameters and Detailed Configuration}
\label{appendix:hyperparameters}

Here, we report the hyperparameters' search space.
We searched learning rate $\eta$, learning rate decay rate $\rho$ and the timing to decay learning rate $\delta$, and regularization coefficient of weight decay $\lambda$. When $\delta = 0.7$, the learning rate decays when training passes 70\% of the total iterations.
Furthermore, a parameter to control momentum $\gamma$ was added to the hyperparameters. 

To set the range in which to search for each hyperparameter, we followed the configurations of \citet{Choi19} and \citet{Shallue19}. 
Table \ref{table:workloads} summarizes the workloads used in the experiment. We did not use batch normalization and the input image is simply normalized; no data augmentation was employed. The hyperparameter ranges are summarized in Tables \ref{table:hp-range-tinymnist}, \ref{table:hp-range-mnist}, \ref{table:hp-range-cifar10}, and \ref{table:hp-range-cifar100}. We conducted a Bayesian optimization to explore hyperparameters in the range described in the tables.

\bgroup

\setlength{\tabcolsep}{2pt}
\begin{table}[htb]
\centering
\caption{Experiments: Workloads}
\label{table:workloads}
\setlength{\tabcolsep}{0.5em} %
{\renewcommand{\arraystretch}{1.2}%
\begin{tabular}[t]{lccccc}
\hline
Model & Dataset & Batch Size & Step Budget & Epoch \\
\hline \hline
2-NN w/o SC \citep{Thomas2019} & TinyMNIST & 512 & 11343 & 120\\ \hline
3-LNN w/ SC & TinyMNIST    & 8192 & 300 & 60\\ \hline
3-LNN w/o SC & TinyMNIST    & 8192 & 300 & 60\\ \hline
3-NN w/ SC & TinyMNIST    & 8192 & 300 & 60\\ \hline
3-NN w/o SC & TinyMNIST    & 8192 & 300 & 60\\ \hline  \hline

6-LNN w/ SC & MNIST    & 8192 & 300 & 60\\ \hline
6-LNN w/o SC & MNIST    & 8192 & 300 & 60\\ \hline
6-NN w/ SC & MNIST    & 8192 & 300 & 60\\ \hline
6-NN w/o SC & MNIST    & 8192 & 300 & 60\\ \hline
Simple CNN Base \citep{Shallue19} & MNIST   & 256 & 9350 & 60\\ \hline
6-LNN w/ SC & CIFAR-10    & 256 & 10205 & 60\\ \hline
6-LNN w/o SC & CIFAR-10    & 256 & 10205 & 60\\ \hline
VGG-16 w/o BN \citep{Simonyan15} & CIFAR-10    & 128 & 78000 & 250\\ \hline
ResNet-8 w/o BN \citep{Shallue19} & CIFAR-10  & 256 & 15800 & 120\\ \hline
ResNet-8 w/o BN \citep{Shallue19} & CIFAR-100 & 256 & 15800 & 120\\ \hline 

\end{tabular}\\

\caption{Hyperparameter Search Range for TinyMNIST Dataset Experiments}
\label{table:hp-range-tinymnist}
\begin{tabular}[t]{lcccccccccc}

\hline
Model & $\eta$ & $\rho$ & $\delta$ & $\lambda$ & $\gamma$ &\\

\hline
2-LNN w/o SC &
\begin{tabular}{c}
     [1e-3, 1e-1]
\end{tabular}
& 
\begin{tabular}{c}
     [1e-2, 1] 
\end{tabular}
& 
\begin{tabular}{c}
     [1e-2, 1] 
\end{tabular}
&
\begin{tabular}{c}
     [0] 
\end{tabular}
&
\begin{tabular}{c}
     [0, 0.999]  
\end{tabular}\\ \hline

\hline
3-NN w/o and w/ SC &
\begin{tabular}{c}
     [1e-4, 1e-1]
\end{tabular}
& 
\begin{tabular}{c}
     [5e-1, 1] 
\end{tabular}
& 
\begin{tabular}{c}
     [5e-1, 1] 
\end{tabular}
&
\begin{tabular}{c}
     [0] 
\end{tabular}
&
\begin{tabular}{c}
     [0, 0.999]  
\end{tabular}\\ \hline

\hline
3-LNN w/o and w/ SC &
\begin{tabular}{c}
     [1e-4, 1e-1]
\end{tabular}
& 
\begin{tabular}{c}
     [5e-1, 1] 
\end{tabular}
& 
\begin{tabular}{c}
     [5e-1, 1] 
\end{tabular}
&
\begin{tabular}{c}
     [0] 
\end{tabular}
&
\begin{tabular}{c}
     [0, 0.999]  
\end{tabular}\\ \hline

\end{tabular}

\caption{Hyperparameter Search Range for MNIST Dataset Experiments}
\label{table:hp-range-mnist}
\begin{tabular}[t]{lcccccccccc}

\hline
Model & $\eta$ & $\rho$ & $\delta$ & $\lambda$ & $\gamma$ &\\

\hline
6-NN w/o and w/ SC &
\begin{tabular}{c}
     [1e-4, 1e-1]
\end{tabular}
& 
\begin{tabular}{c}
     [5e-1, 1] 
\end{tabular}
& 
\begin{tabular}{c}
     [5e-1, 1] 
\end{tabular}
&
\begin{tabular}{c}
     [0] 
\end{tabular}
&
\begin{tabular}{c}
     [0, 0.999]  
\end{tabular}\\ \hline

\hline
6-LNN w/o and w/ SC &
\begin{tabular}{c}
     [1e-4, 1e-1]
\end{tabular}
& 
\begin{tabular}{c}
     [5e-1, 1] 
\end{tabular}
& 
\begin{tabular}{c}
     [5e-1, 1] 
\end{tabular}
&
\begin{tabular}{c}
     [0] 
\end{tabular}
&
\begin{tabular}{c}
     [0, 0.999]  
\end{tabular}\\ \hline

\hline
Simple CNN &
\begin{tabular}{c}
     [1e-4, 1]
\end{tabular}
& 
\begin{tabular}{c}
     [5e-1, 1] 
\end{tabular}
& 
\begin{tabular}{c}
     [5e-1, 1] 
\end{tabular}
&
\begin{tabular}{c}
     [0] 
\end{tabular}
&
\begin{tabular}{c}
     [0, 0.999]  
\end{tabular}\\ \hline

\end{tabular}

\caption{Hyperparameter Search Range for CIFAR10 Dataset Experiments}
\label{table:hp-range-cifar10}
\begin{tabular}[t]{lcccccccccc}

\hline
Model & $\eta$ & $\rho$ & $\delta$ & $\lambda$ & $\gamma$ &\\

\hline
6-LNN w/o and w/ SC &
\begin{tabular}{c}
     [1e-4, 1e-1]
\end{tabular}
& 
\begin{tabular}{c}
     [0.5, 1] 
\end{tabular}
& 
\begin{tabular}{c}
     [0.5, 1] 
\end{tabular}
&
\begin{tabular}{c}
     [0]
\end{tabular}
&
\begin{tabular}{c}
     [0, 0.999]  
\end{tabular}\\ \hline

\hline
ResNet-8 w/o BN &
\begin{tabular}{c}
     [1e-6, 1e+1]
\end{tabular}
& 
\begin{tabular}{c}
     [0.5, 1] 
\end{tabular}
& 
\begin{tabular}{c}
     [0.5, 1] 
\end{tabular}
& [1e-5, 1e-4] &
\begin{tabular}{c}
     [1e-4, 0.999] 
\end{tabular}\\

\hline
VGG-16 w/o BN &
\begin{tabular}{c}
     [1e-3, 1e-0]
\end{tabular}
& 
\begin{tabular}{c}
     [0.5, 1] 
\end{tabular}
& 
\begin{tabular}{c}
     [0.5, 1] 
\end{tabular}
&
\begin{tabular}{c}
     [1e-4, 1e-1]
\end{tabular}
&
\begin{tabular}{c}
     [1e-4, 0.999]  
\end{tabular}\\ \hline
\end{tabular}

\caption{Hyperparameter Search Range for CIFAR100 Dataset Experiment}
\label{table:hp-range-cifar100}
\begin{tabular}[t]{lcccccccccc}

\hline
Model & $\eta$ & $\rho$ & $\delta$ & $\lambda$ & $\gamma$ &\\

\hline
ResNet-8 w/o BN &
\begin{tabular}{c}
     [1e-6, 1e+1]
\end{tabular}
& 
\begin{tabular}{c}
     [0.5, 1] 
\end{tabular}
& 
\begin{tabular}{c}
     [0.5, 1] 
\end{tabular}
& [1e-5, 1e-4] &
\begin{tabular}{c}
     [1e-4, 0.999] 
\end{tabular}\\ \hline

\end{tabular}

}\end{table}

\egroup
\clearpage

\subsection{Distribution of Train Loss and Generalization Gap}
\label{appendix:distribution_loss}

\begin{figure}[h]
    \centering
	\includegraphics[width=0.45\linewidth]{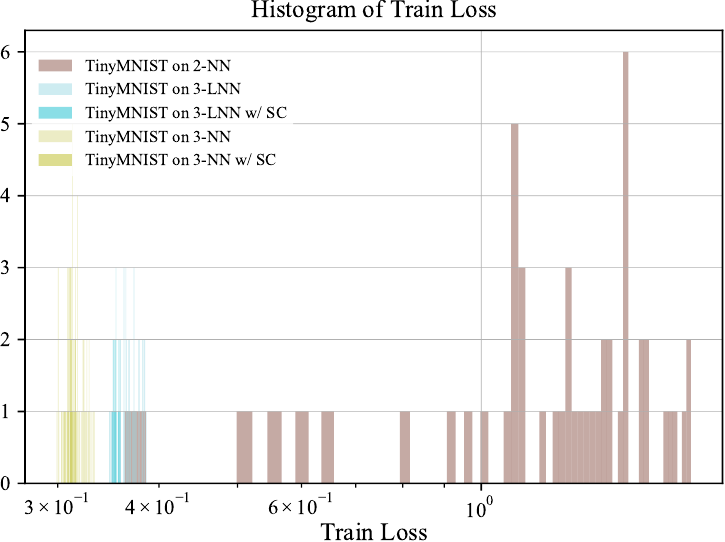}
	\includegraphics[width=0.45\linewidth]{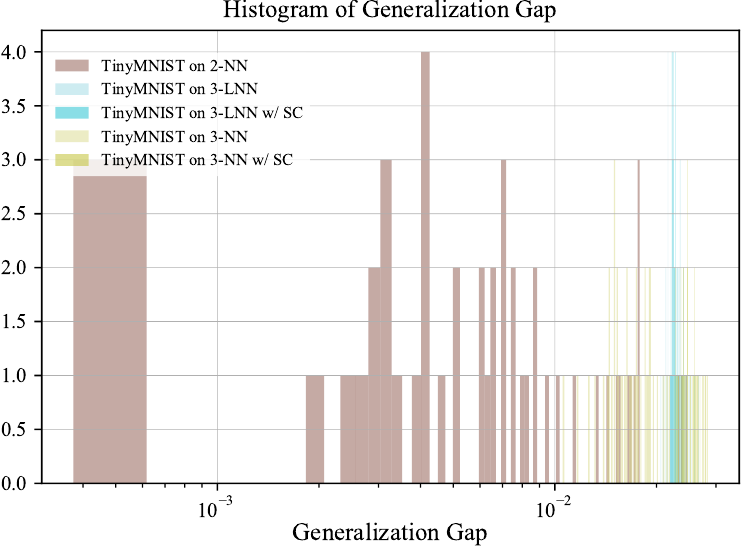}

	\includegraphics[width=0.45\linewidth]{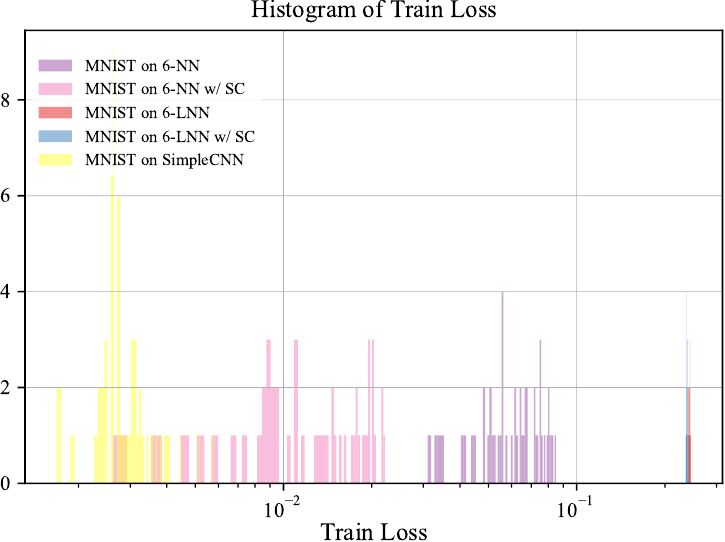}	
	\includegraphics[width=0.45\linewidth]{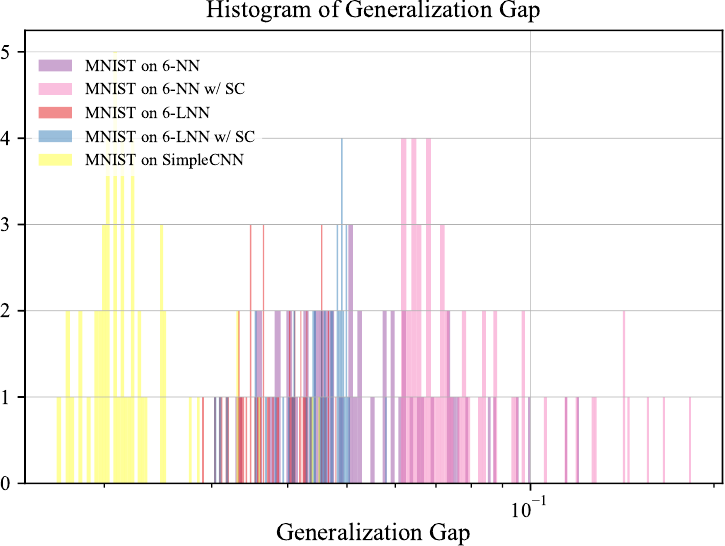}
	
	\includegraphics[width=0.45\linewidth]{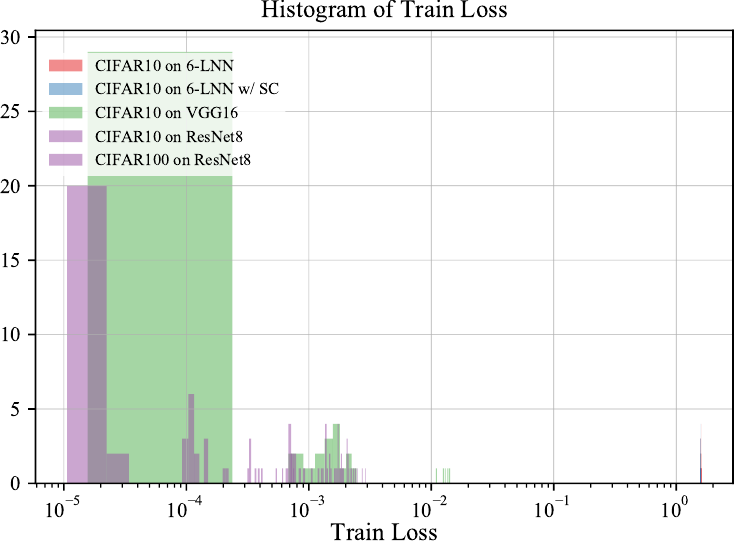}
	\includegraphics[width=0.45\linewidth]{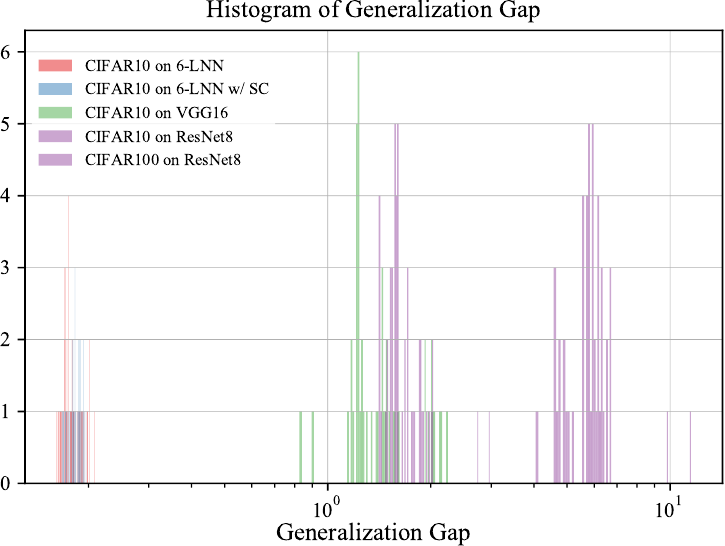}

	\caption{Distribution of training loss and generalization gap on the trained models.}
	\label{fig:dsitribution_loss}
\end{figure}

The histograms of the model losses and generalization gap used in the experiment are shown in Figure \ref{fig:dsitribution_loss}.

\clearpage

\section{Additional Experimental Results}
\label{appendix:additional-experiment}

\subsection{Full Results of Correlation between Generalization Gap and TIC Lower Bound Estimates}
\begin{figure}[h]
    \centering
	\includegraphics[width=0.99\linewidth]{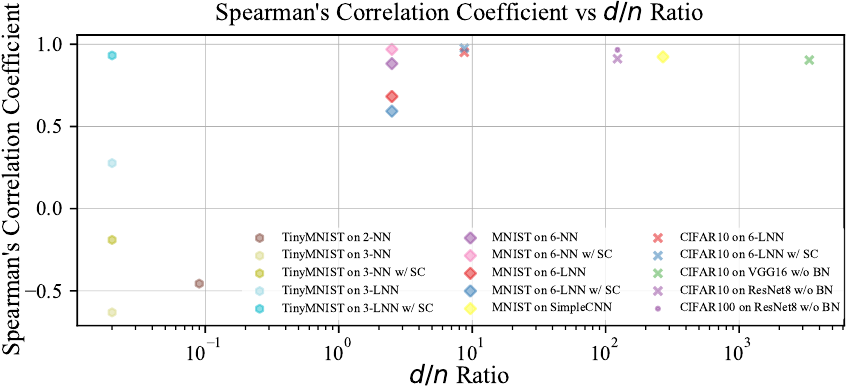}
	\includegraphics[width=0.99\linewidth]{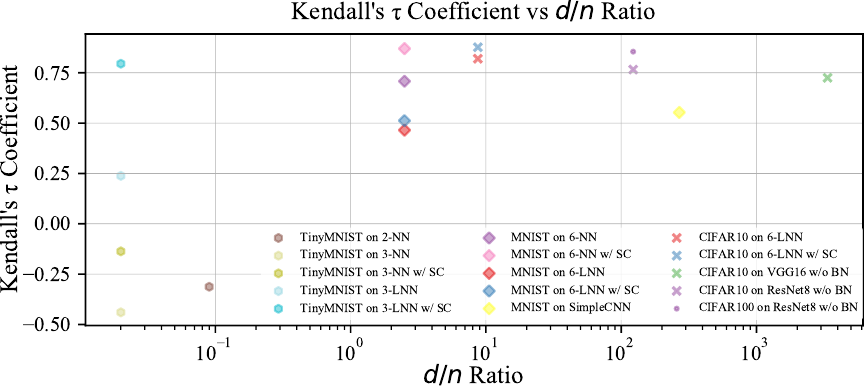}
	\includegraphics[width=0.99\linewidth]{figs/correlation/all_Pearson_s_Coefficient.pdf}
	\caption{Relationship between correlation coefficient and $d/n$. It should be noted that the correlation between the TIC estimates and the generalization gap is high in regions with large $d/n$, which are considered to be close to the NTK regime.}
	\label{fig:dp-vs-coeff}
\end{figure}

Table \ref{table:all_correlation} summarizes these results, evaluated for three different correlation coefficients. 
The results are for the three types of correlation coefficients calculated for the plots shown in figures \ref{fig:tinymnist_fast_tic_f}, \ref{fig:mnist_fast_tic_f}, and \ref{fig:cifar_fast_tic_f}.
The relationship between these correlation coefficients and the values of the ratios of the parameters to the number of data points is shown in Figure \ref{fig:dp-vs-coeff}.
The result of plotting these results, along with $d/n$, is shown in Figure \ref{fig:dp-vs-coeff}.

\begin{table}[h]
\caption{Correlation: TIC estimates $\mathrm{Tr}(\boldsymbol{C}(\bm{ {\theta})}) / \mathrm{Tr}(\boldsymbol{F}(\bm{ {\theta})})$ and generalization gap}
\centering
\label{table:all_correlation}
\setlength{\tabcolsep}{0.5em} %
{\renewcommand{\arraystretch}{1.2}%
\begin{tabular}{llccc}
\hline
Model             & Dataset     & Spearman's Correlation & Kendall's $\tau$ & Pearson's Correlation \\ \hline\hline
2-NN   & Tiny MNIST  & -0.456                 & -0.313                       & -0.309                \\
3-NN        & Tiny MNIST  & -0.631                 & -0.44                        & -0.766                \\
3-NN w/ SC  & Tiny MNIST  & -0.19                  & -0.137                       & -0.347                \\
3-LNN       & Tiny MNIST  & 0.277                  & 0.238                        & 0.256                 \\
3-LNN w/ SC & Tiny MNIST  & 0.932                  & 0.795                        & 0.898                 \\ \hline
6-NN        & MNIST       & 0.882                  & 0.708                        & 0.425                 \\
6-NN w/ SC  & MNIST       & 0.969                  & 0.87                         & 0.478                 \\
6-LNN       & MNIST       & 0.682                  & 0.465                        & 0.774                 \\
6-LNN w/ SC & MNIST       & 0.593                  & 0.512                        & 0.848                 \\
Simple CNN        & MNIST       & 0.923                  & 0.553                        & 0.763                 \\ \hline
6-LNN       & CIFAR10     & 0.951                  & 0.82                         & 0.888                 \\
6-LNN w/ SC & CIFAR10     & 0.976                  & 0.877                        & 0.965                 \\
VGG-16 w/o BN           & CIFAR10     & 0.904                  & 0.725                        & 0.933                 \\
ResNet-8 w/o BN         & CIFAR10     & 0.912                  & 0.766                        & 0.983                 \\
ResNet-8 w/o BN         & CIFAR100    & 0.966                  & 0.855                        & 0.978                 \\ \hline
\end{tabular}}
\end{table}

\clearpage

\subsection{Additional Results of Small-Scale Experiments}
\label{appendix_small_scale}

Here, we provide details of the small-scale experimental results that could not be included in the main paper.
In particular, we investigate the goodness of approximation of the information matrix in the small-scale case, since it can be computed exactly, although the execution time is longer.

\subsubsection{Empirical Relationship between H and F}

\begin{figure}[h]
    \centering
	\includegraphics[width=0.45\linewidth]{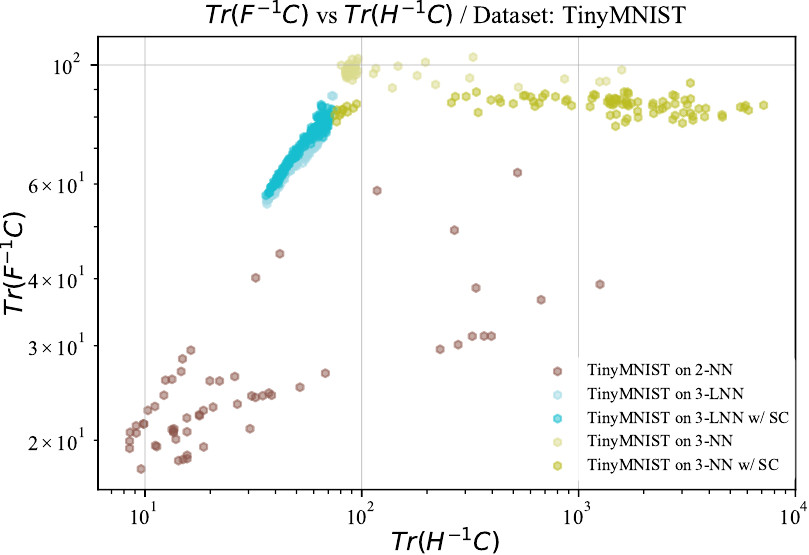}
    \includegraphics[width=0.45\linewidth]{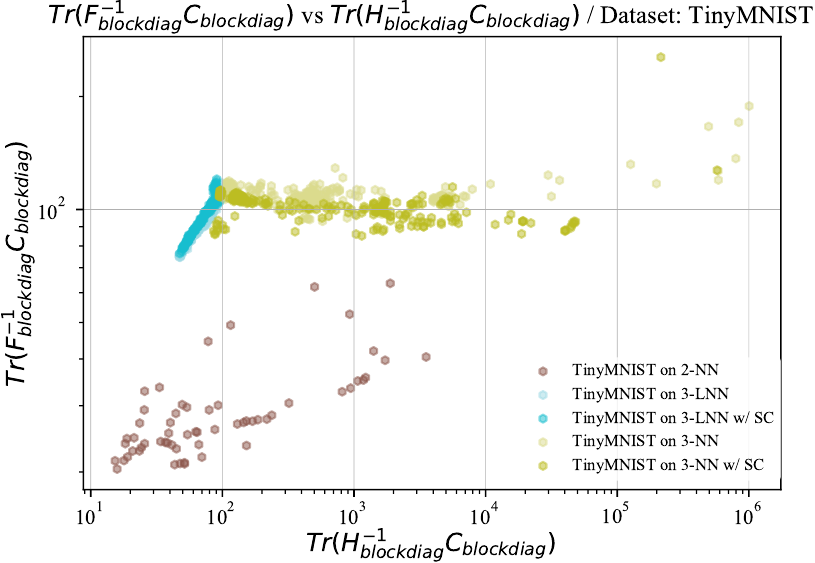}
    \includegraphics[width=0.45\linewidth]{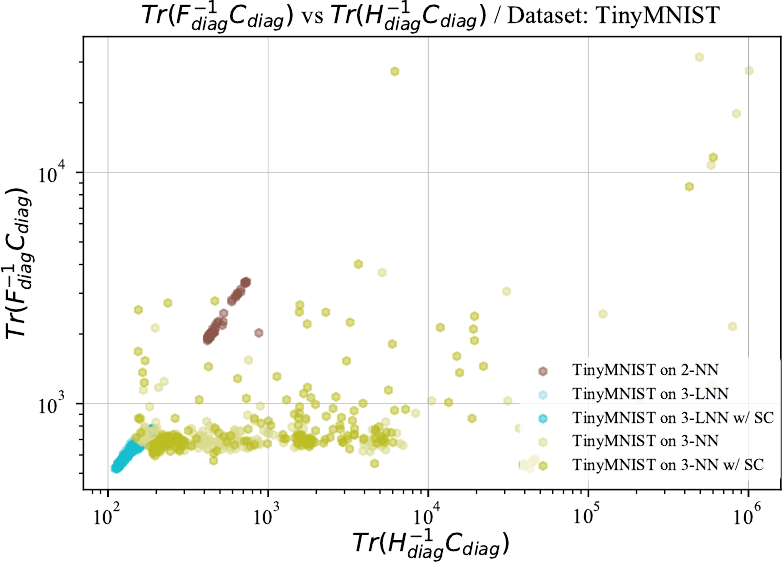}
    \includegraphics[width=0.45\linewidth]{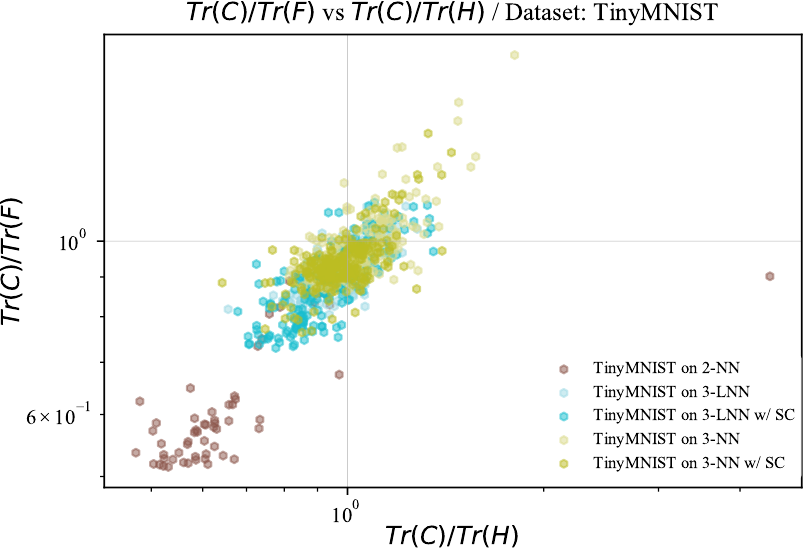}
    \includegraphics[width=0.45\linewidth]{figs/approx/h/5/v_trace_h_vs_v_trace_f.pdf}
	\caption{Small-scale experiments: comparison of the TIC estimate w/ $\boldsymbol{H}(\boldsymbol{\theta})$ and $\boldsymbol{F}(\boldsymbol{\theta})$. Dataset: TinyMNIST. }
	\label{fig:h_and_f}
\end{figure}

First, we examine the behavior of $\boldsymbol{H}(\boldsymbol{\theta})$ when it is approximated by $\boldsymbol{F}(\boldsymbol{\theta})$, or more precisely, $\boldsymbol{F}_{\text{1mc}}(\boldsymbol{\theta})$.
Figure \ref{fig:h_and_f} shows that $\boldsymbol{H}(\boldsymbol{\theta})$ is not in perfect agreement with $\boldsymbol{F}(\boldsymbol{\theta})$ due to the effect of the damping term.
However, for cases that do not require inverse calculations, such as trace calculations and lower bounds, these effects can be eliminated and a relatively good approximation can be achieved.
Furthermore, in the case of linear neural networks, $\boldsymbol{H}(\boldsymbol{\theta})$ and $\boldsymbol{F}(\boldsymbol{\theta})$ show a strong correlation.

\subsubsection{Effect of Matrix Shape Approximation on Estimation of TIC}
\label{appendix:approx-exp}

Next, we fixed the use of $\boldsymbol{F}(\boldsymbol{\theta})$ rather than $\boldsymbol{H}(\boldsymbol{\theta})$ and observed the change in the TIC estimates and the correlation with the generalization gap when the form of the matrix is changed by the approximation method.

\begin{figure}[h]
\begin{minipage}{\linewidth}
    \centering
	\includegraphics[width=0.32\linewidth]{figs/approx/f/5/shape/v_exact_tic_f_vs_v_block_diag_tic_f.pdf}
	\includegraphics[width=0.32\linewidth]{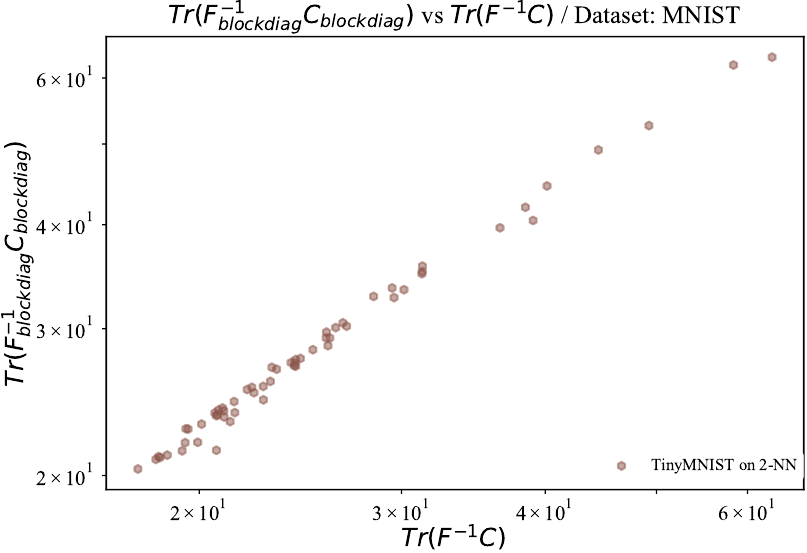}
	\includegraphics[width=0.32\linewidth]{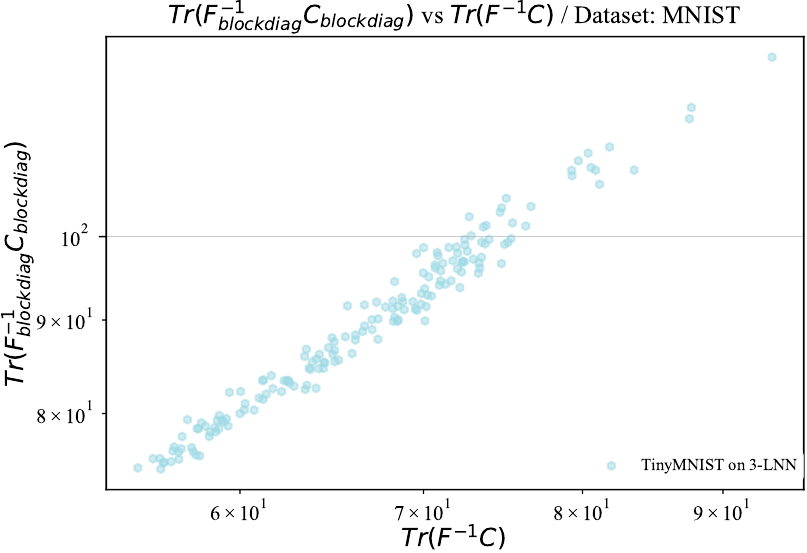}
	\includegraphics[width=0.32\linewidth]{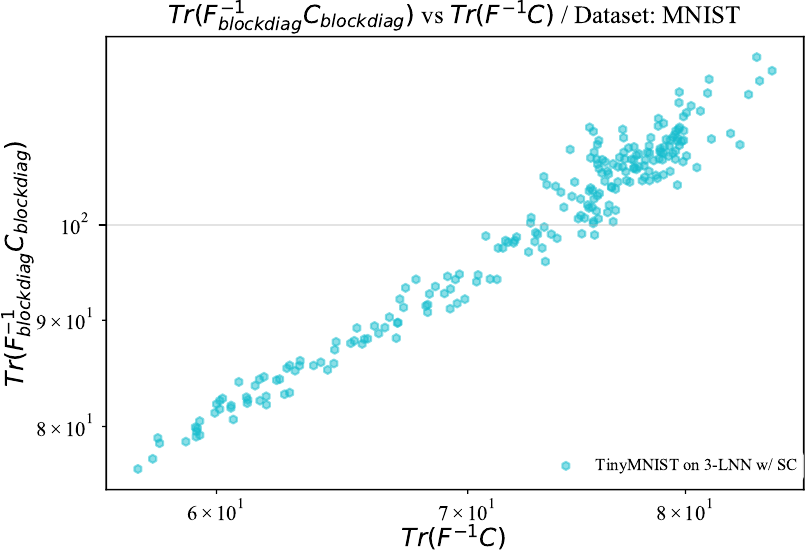}
	\includegraphics[width=0.32\linewidth]{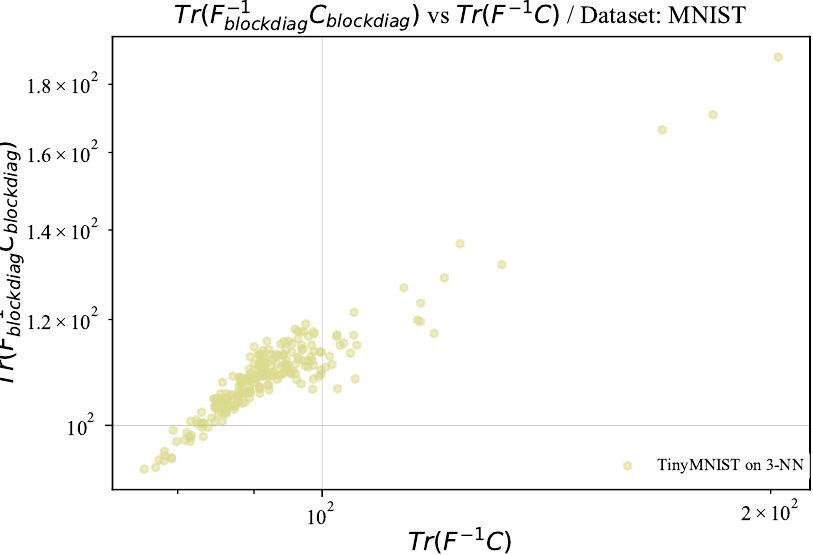}
	\includegraphics[width=0.32\linewidth]{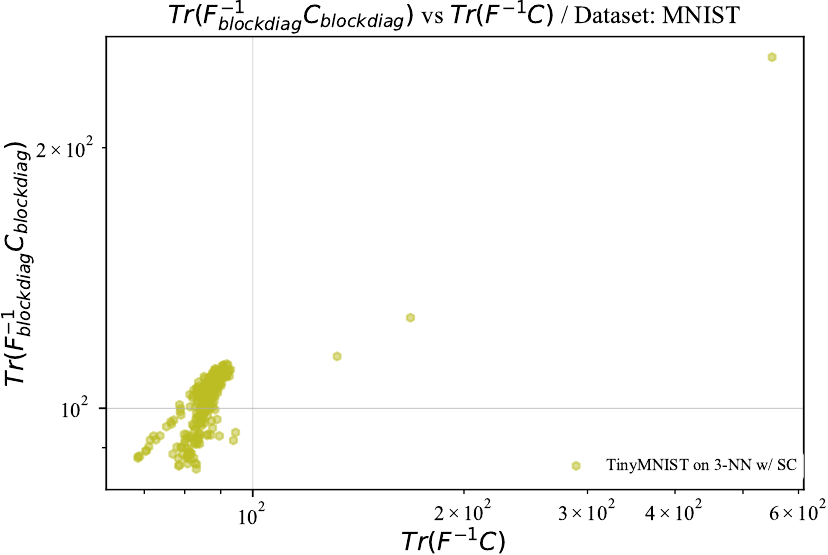}
	\subcaption{Exact vs block-diagonal.}
\end{minipage}
\begin{minipage}{\linewidth}
    \centering
	\includegraphics[width=0.32\linewidth]{figs/approx/f/5/shape/v_exact_tic_f_vs_v_diag_tic_f.pdf}
	\includegraphics[width=0.32\linewidth]{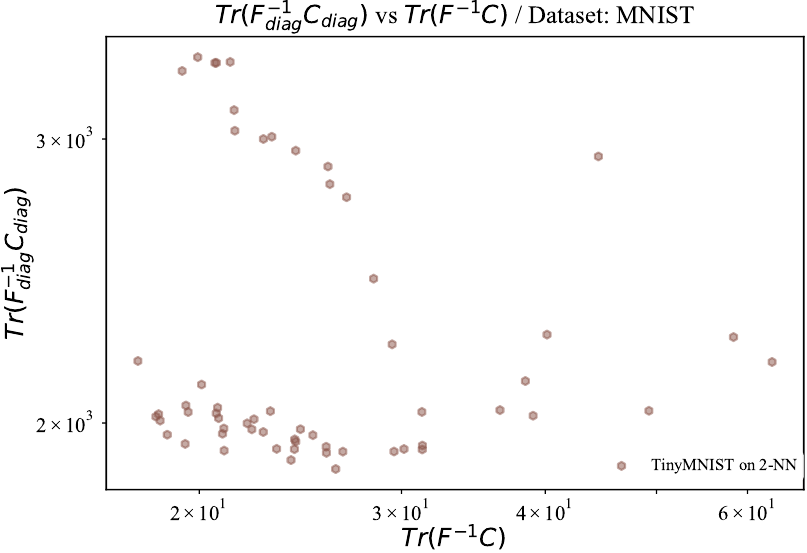}
	\includegraphics[width=0.32\linewidth]{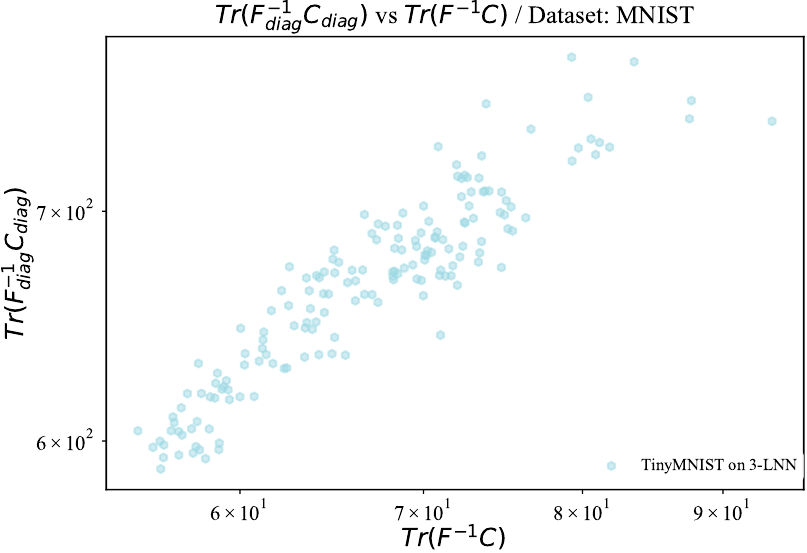}
	\includegraphics[width=0.32\linewidth]{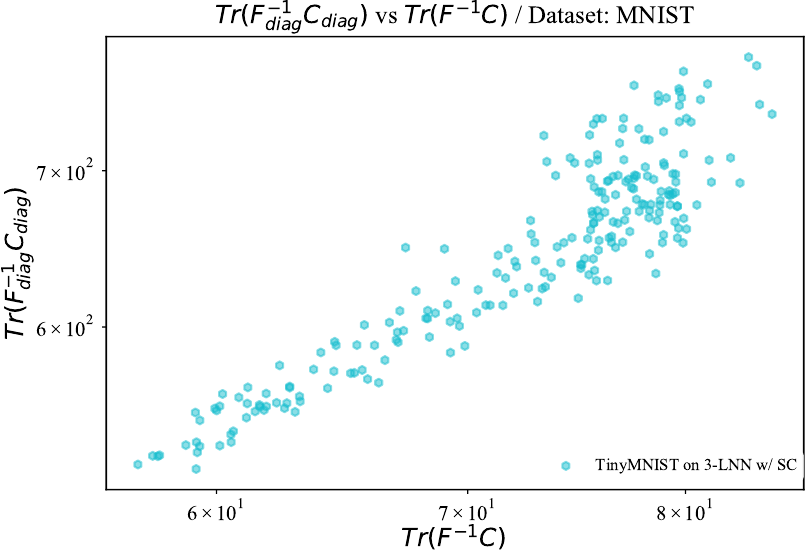}
	\includegraphics[width=0.32\linewidth]{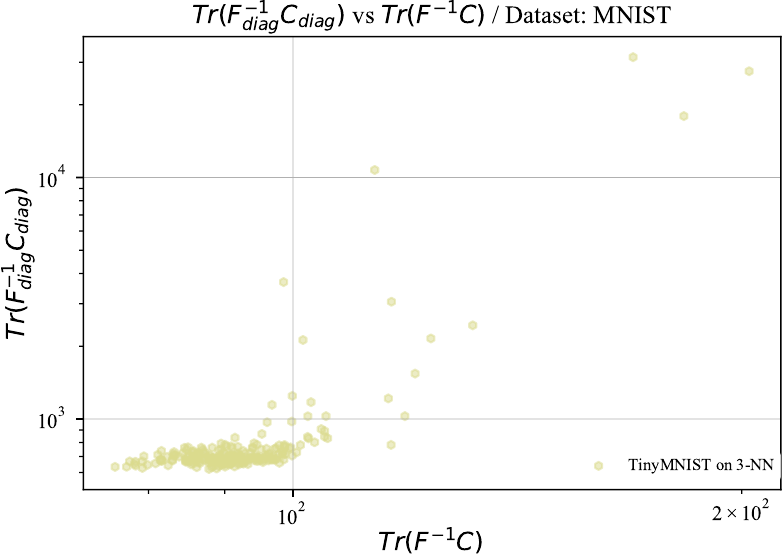}
	\includegraphics[width=0.32\linewidth]{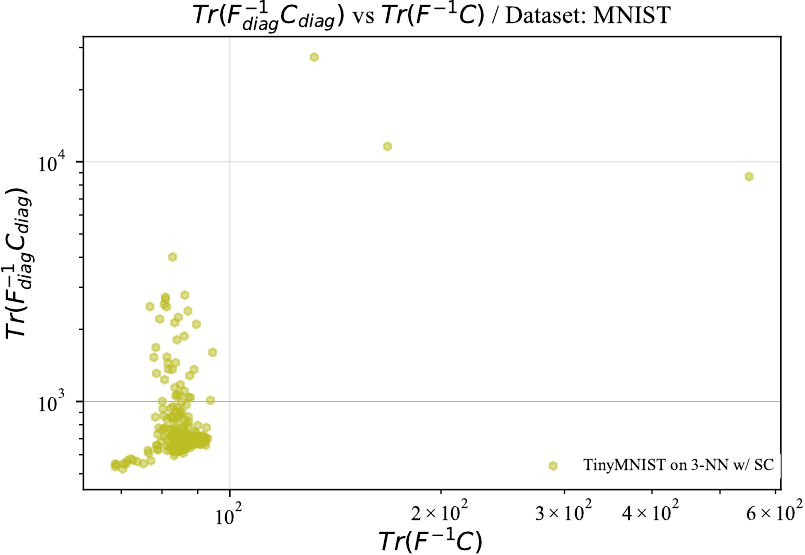}
	\subcaption{Exact vs diagonal.}
\end{minipage}
\begin{minipage}{\linewidth}
    \centering
	\includegraphics[width=0.32\linewidth]{figs/approx/f/5/shape/v_exact_tic_f_vs_v_fast_tic_f.pdf}
	\includegraphics[width=0.32\linewidth]{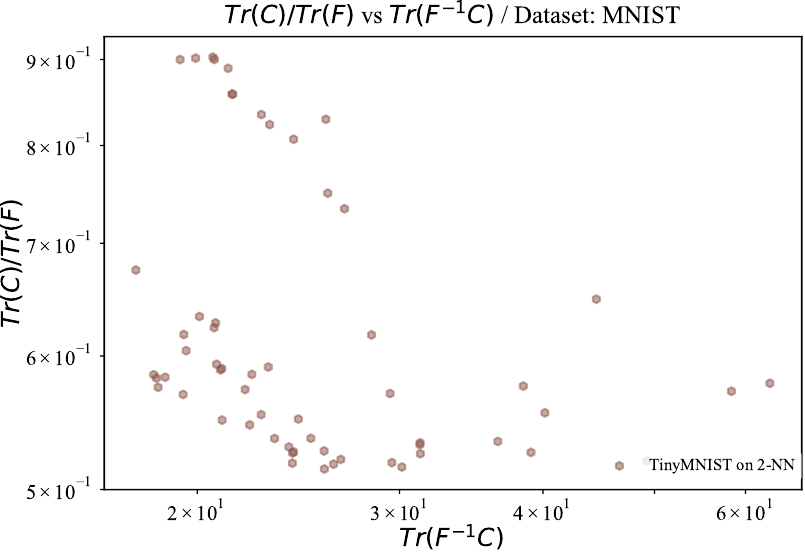}
	\includegraphics[width=0.32\linewidth]{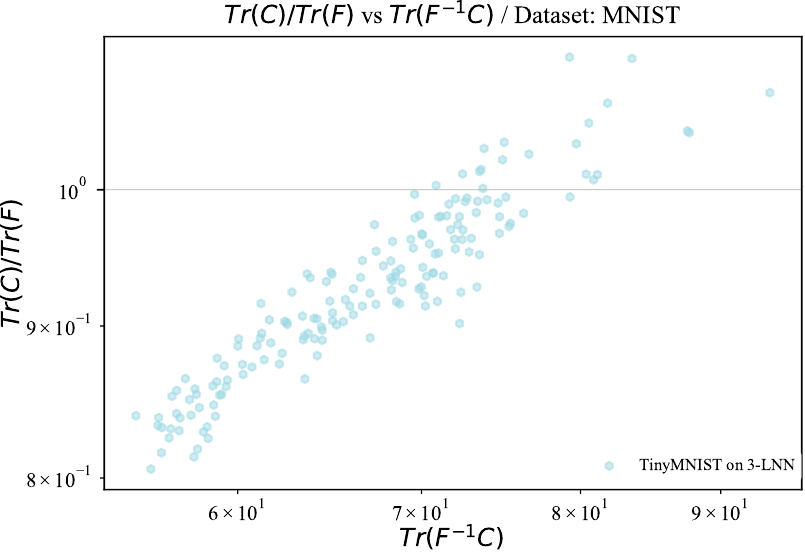}
	\includegraphics[width=0.32\linewidth]{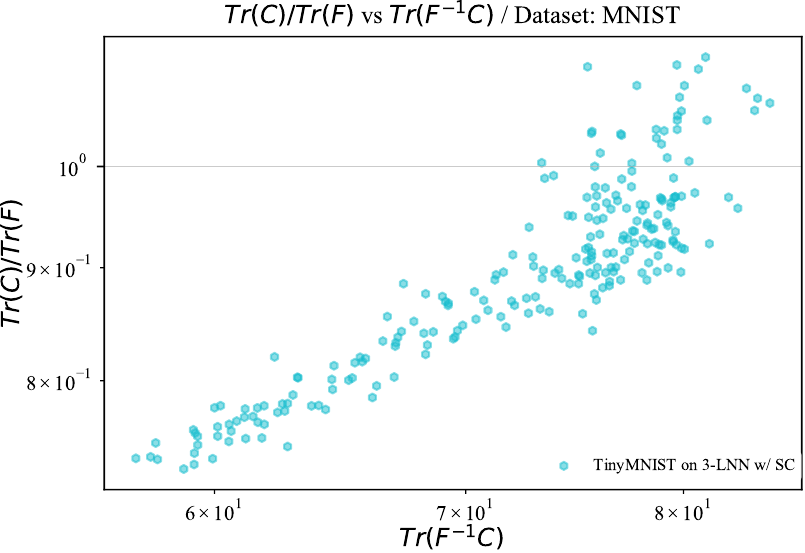}
	\includegraphics[width=0.32\linewidth]{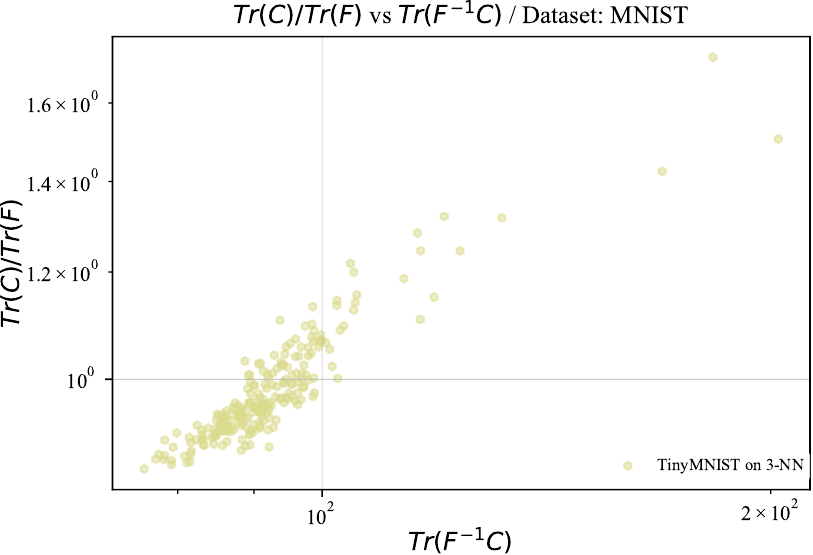}
	\includegraphics[width=0.32\linewidth]{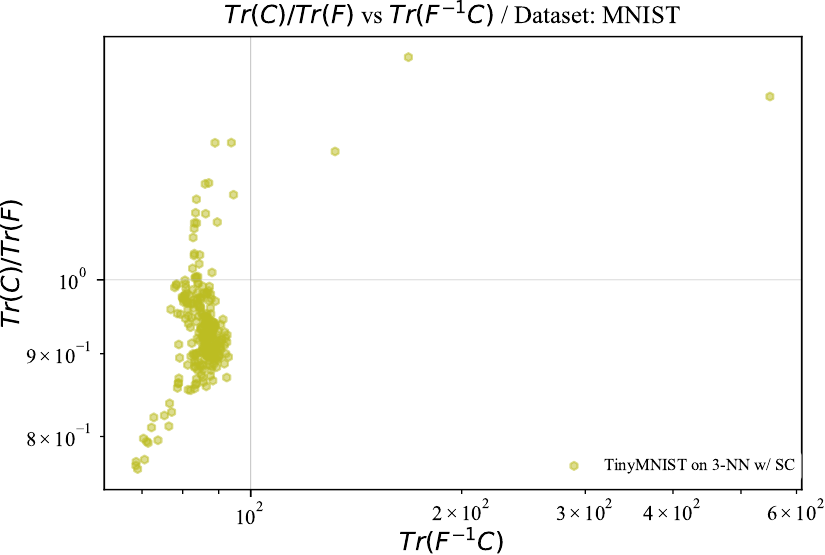}
	\subcaption{Exact vs lower bound.}
	\end{minipage}
	\caption{Small-scale experiments: comparison of the TIC estimate.}
	\label{fig:shape-comparison1}
\end{figure}

\begin{figure}[h]
\begin{minipage}{\linewidth}
    \centering
	\includegraphics[width=0.3\linewidth]{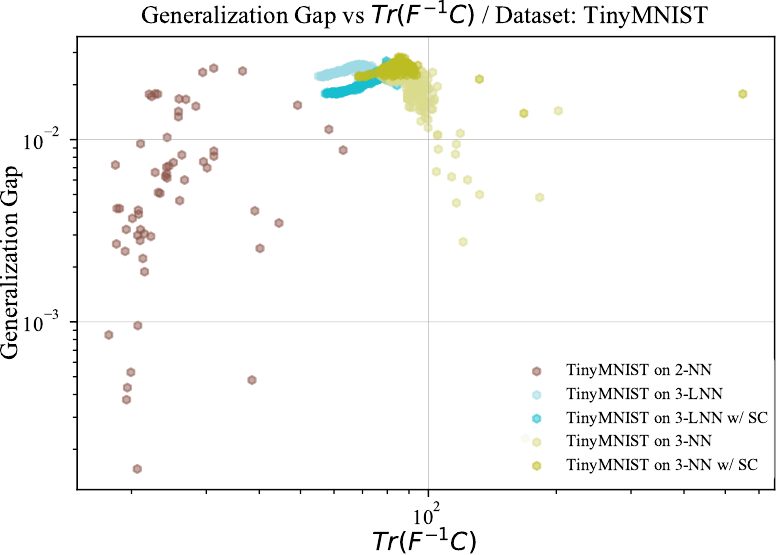}
	\includegraphics[width=0.3\linewidth]{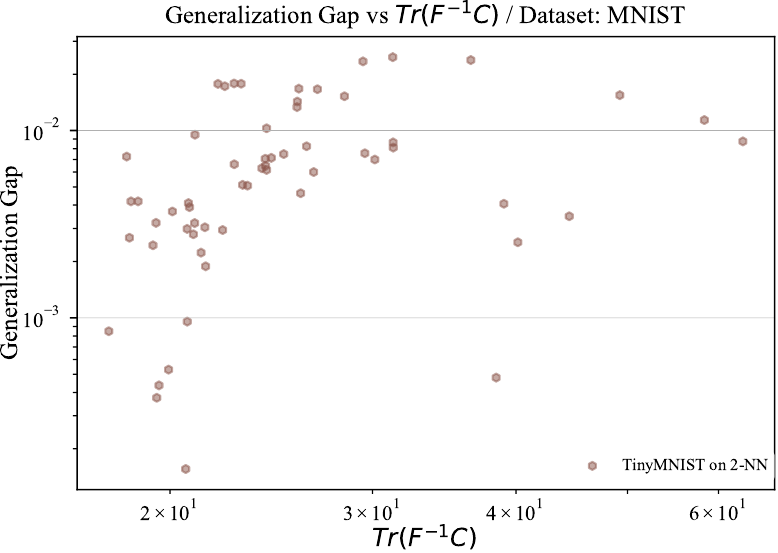}
	\includegraphics[width=0.3\linewidth]{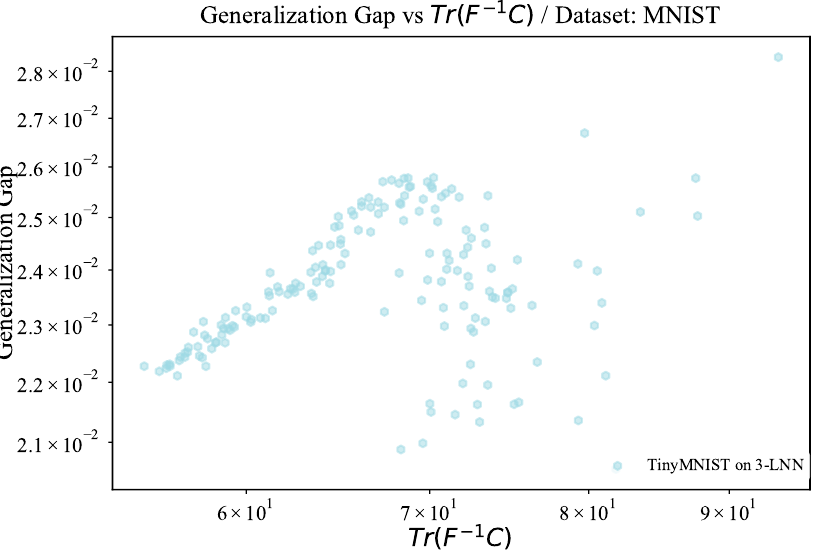}
	\includegraphics[width=0.3\linewidth]{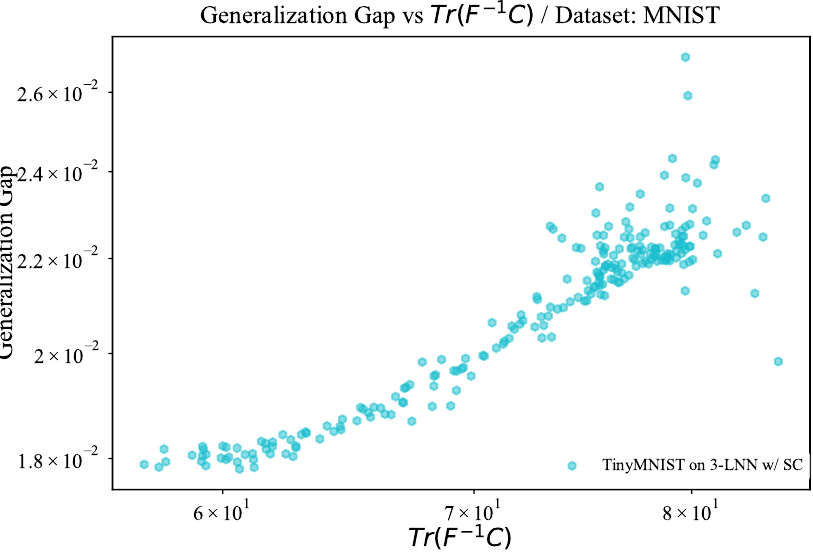}
	\includegraphics[width=0.3\linewidth]{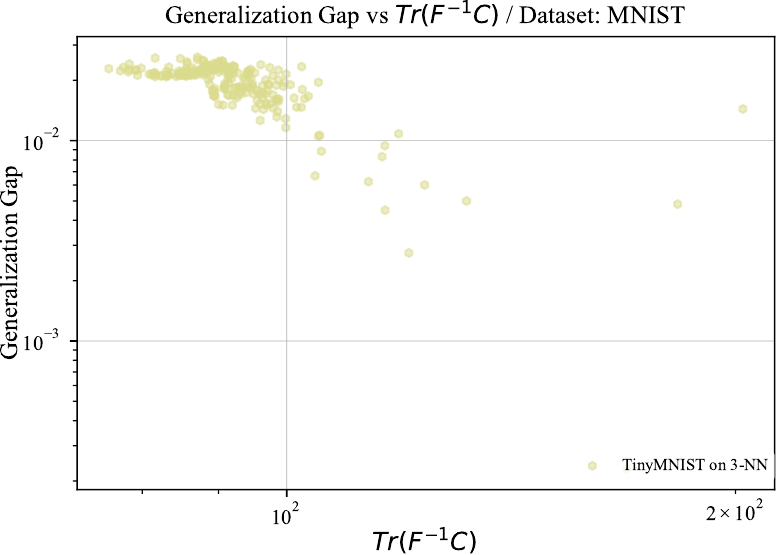}
	\includegraphics[width=0.3\linewidth]{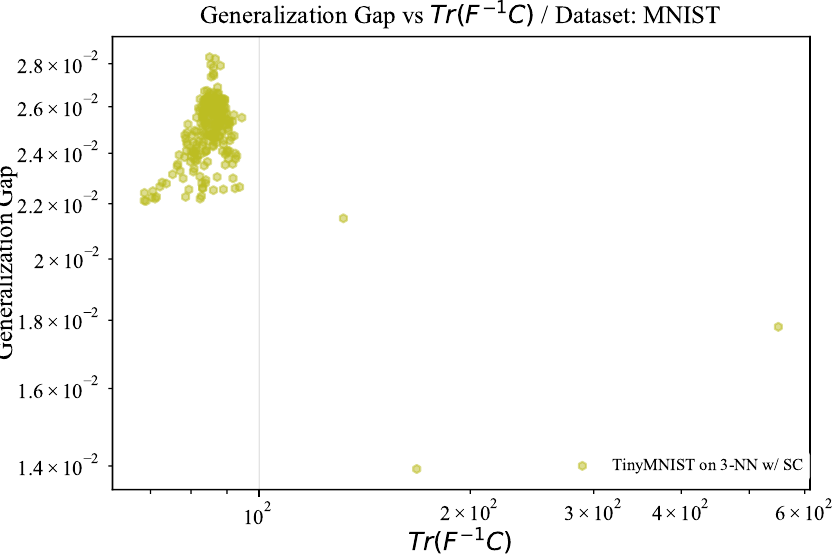}
	\subcaption{\scriptsize{: Generalization gap vs exact TIC.}}

\end{minipage}
\begin{minipage}{\linewidth}
    \centering
	\includegraphics[width=0.3\linewidth]{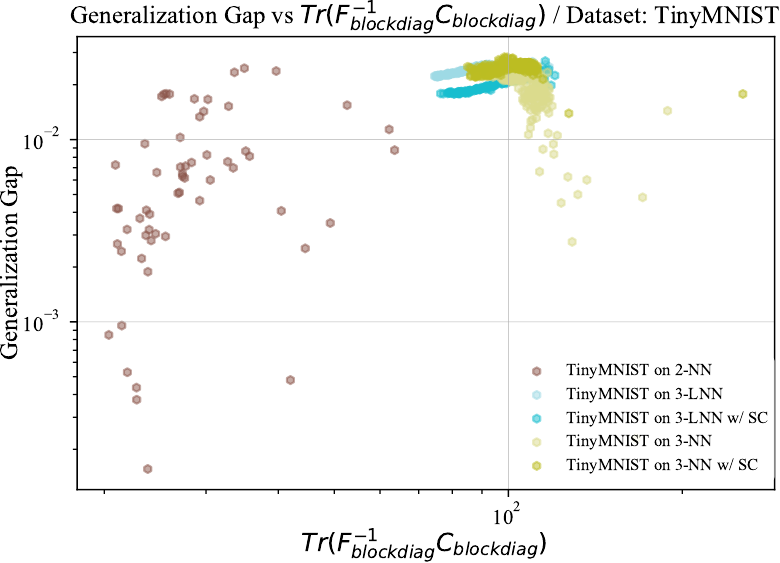}
	\includegraphics[width=0.3\linewidth]{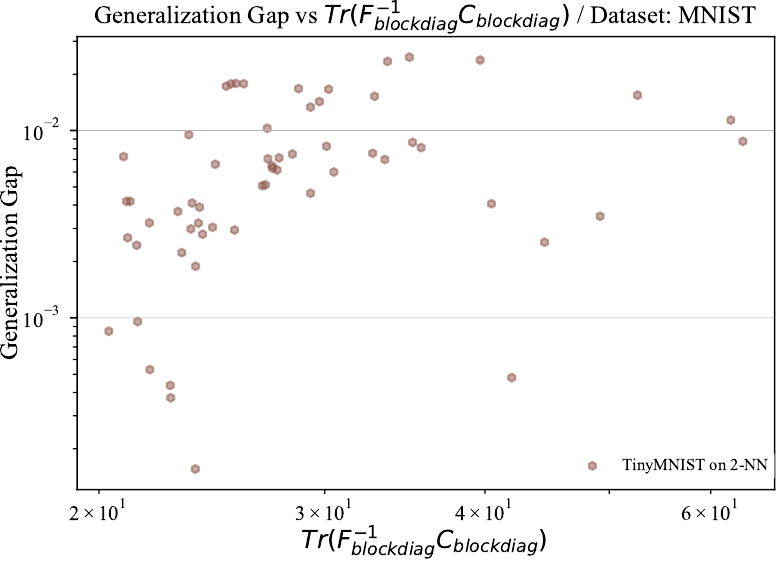}
	\includegraphics[width=0.3\linewidth]{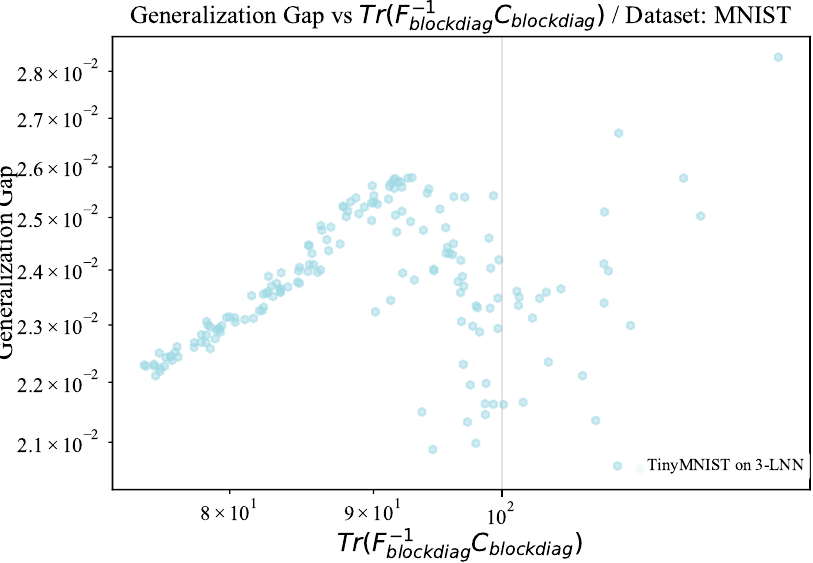}
	\includegraphics[width=0.3\linewidth]{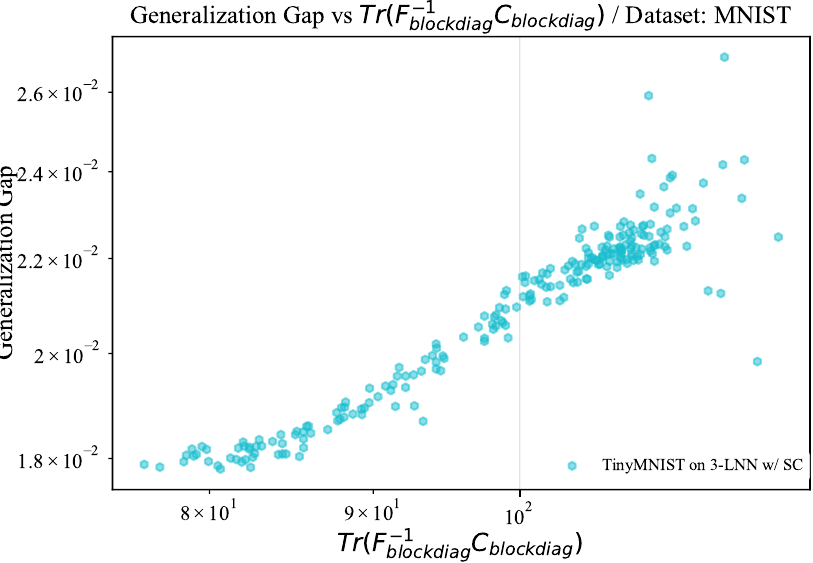}
	\includegraphics[width=0.3\linewidth]{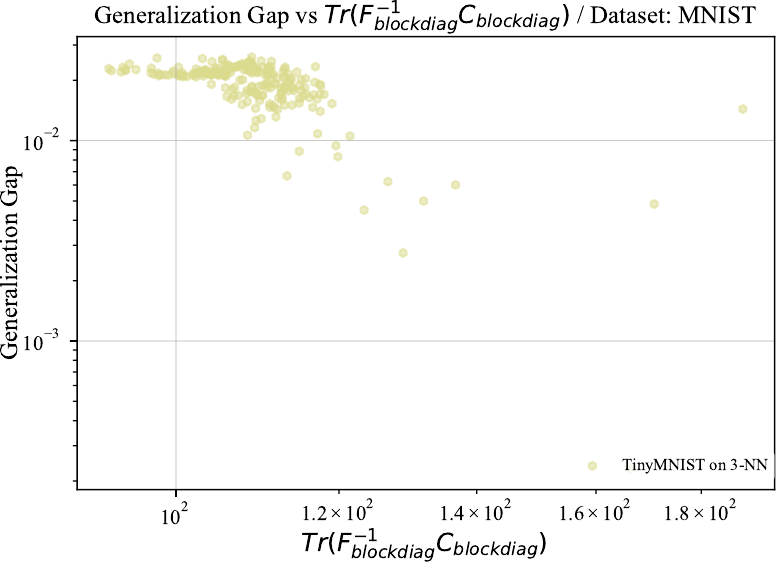}
	\includegraphics[width=0.3\linewidth]{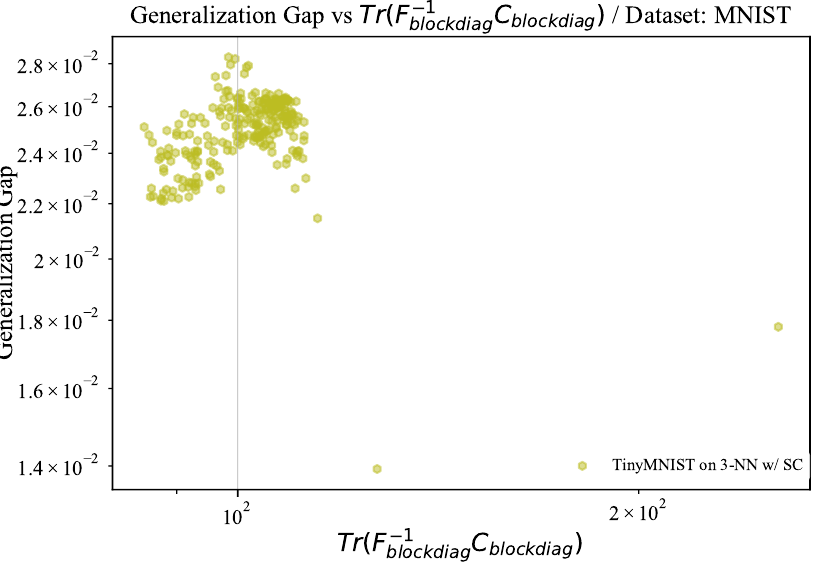}
	\subcaption{\scriptsize{: Generalization gap vs block-diagonal TIC.}}

\end{minipage}

	\caption{Small-scale experiments: comparison of the TIC estimate and generalization gap.}
	\label{fig:tic-gen-comparison2}
\end{figure}

Figures \ref{fig:shape-comparison1} and \ref{fig:tic-gen-comparison2} show the correlation between the TIC estimates and the generalized gaps in the case of approximation.
Figure \ref{fig:shape-comparison1} shows a comparison of the shape of matrix and its estimates, and Figure \ref{fig:tic-gen-comparison2} shows the correlation between generalization gap and TIC estimates.
These values are evaluated using three different correlation metric and summarized in Tables \ref{table:small_exact_coef}, \ref{table:small_block_coef}, and \ref{table:small_diag_coef}.

\newpage

\begin{figure}[h]
\begin{minipage}{\linewidth}
    \centering
	\includegraphics[width=0.3\linewidth]{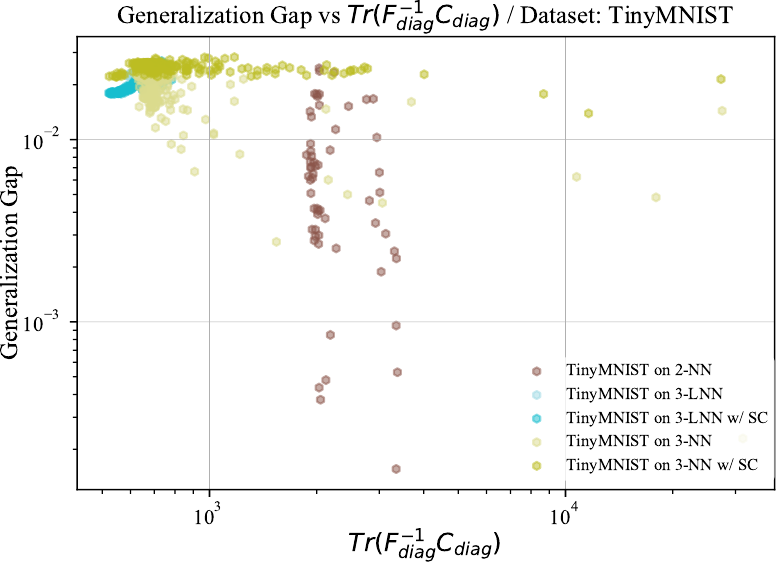}
	\includegraphics[width=0.3\linewidth]{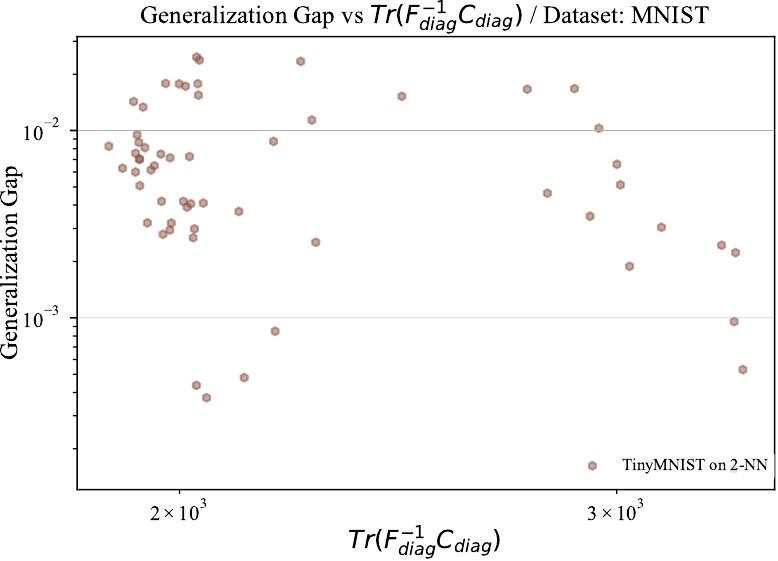}
	\includegraphics[width=0.3\linewidth]{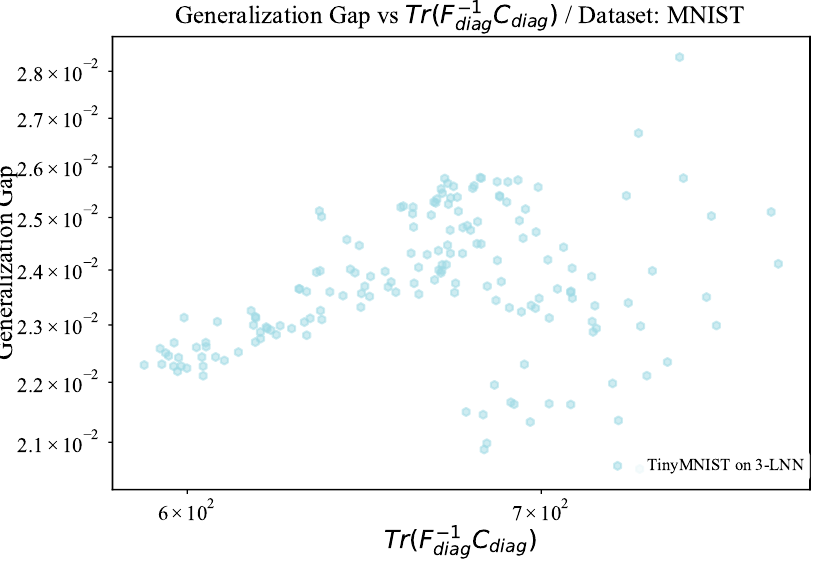}
	\includegraphics[width=0.3\linewidth]{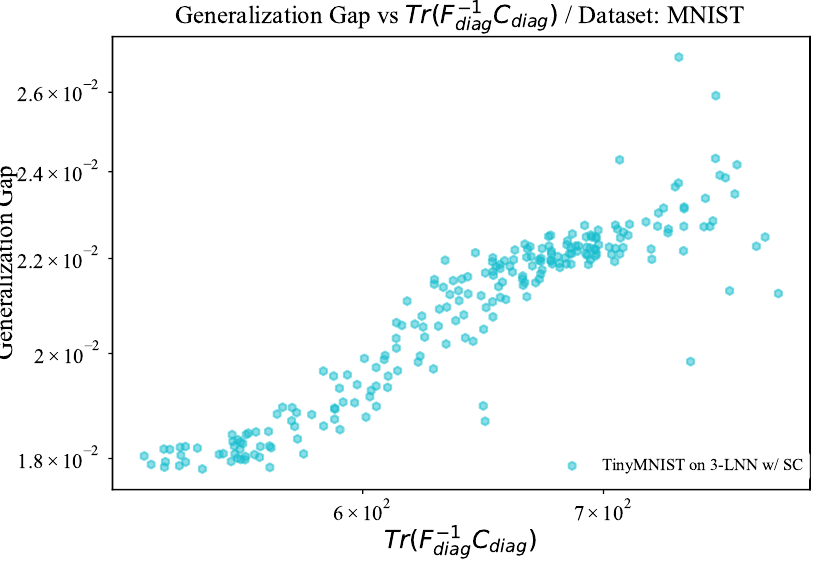}
	\includegraphics[width=0.3\linewidth]{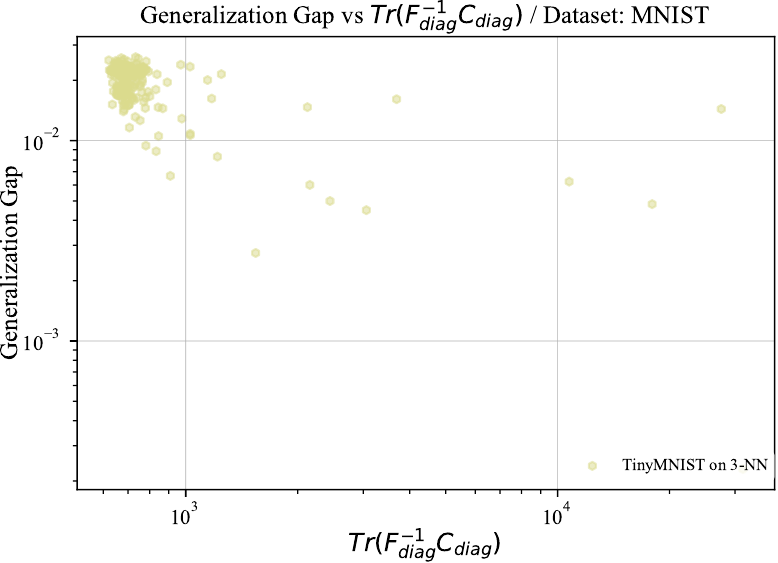}
	\includegraphics[width=0.3\linewidth]{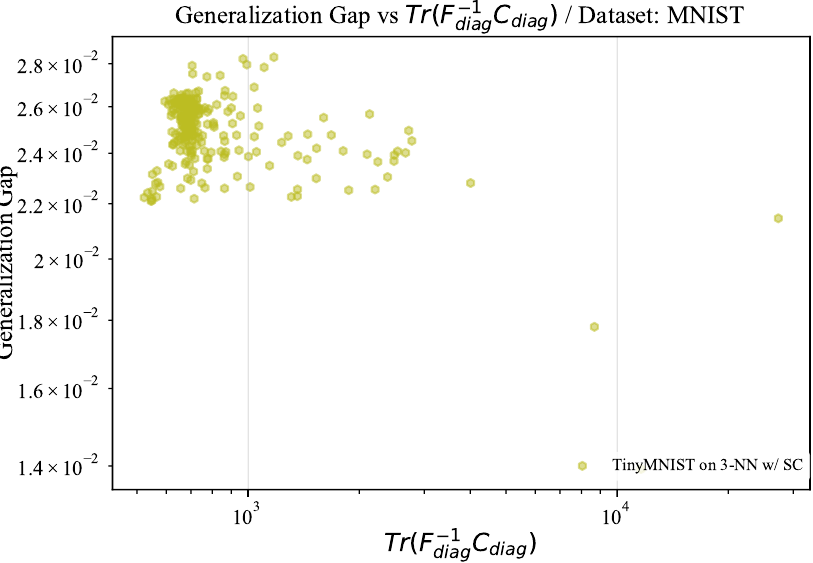}
	\subcaption{\scriptsize{: Generalization gap vs diagonal TIC.}}

\end{minipage}
\begin{minipage}{\linewidth}
    \centering
	\includegraphics[width=0.3\linewidth]{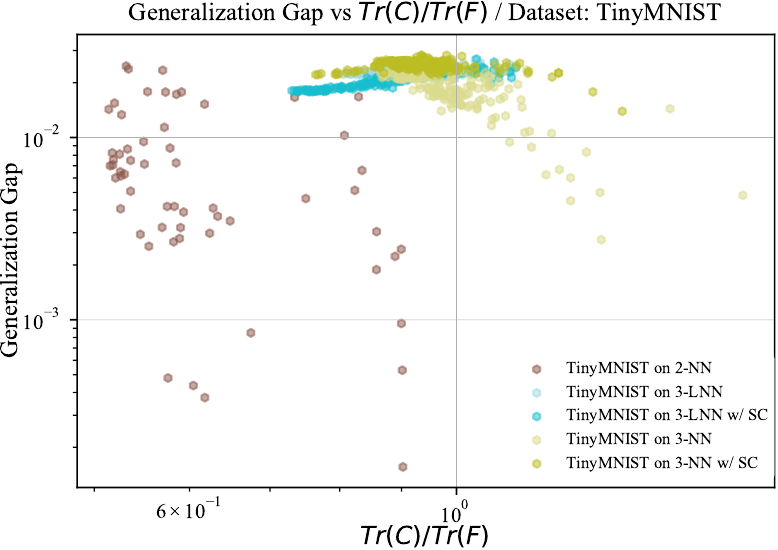}
	\includegraphics[width=0.3\linewidth]{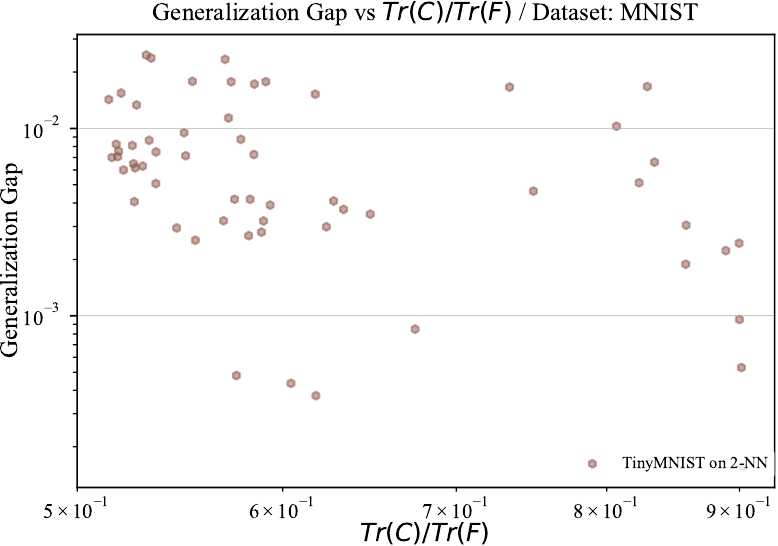}
	\includegraphics[width=0.3\linewidth]{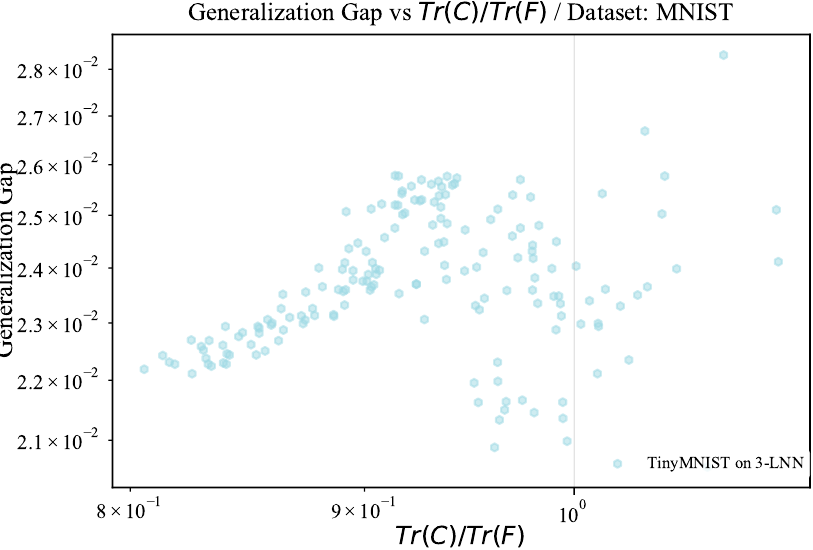}
	\includegraphics[width=0.3\linewidth]{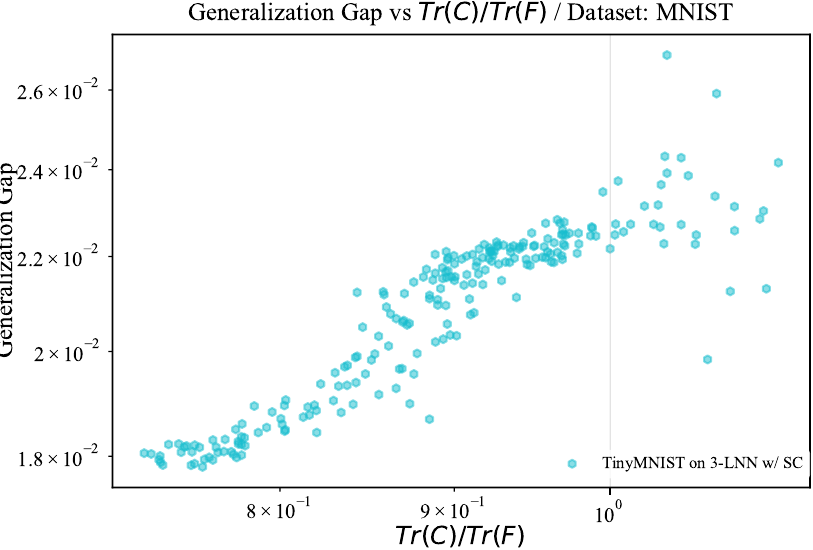}
	\includegraphics[width=0.3\linewidth]{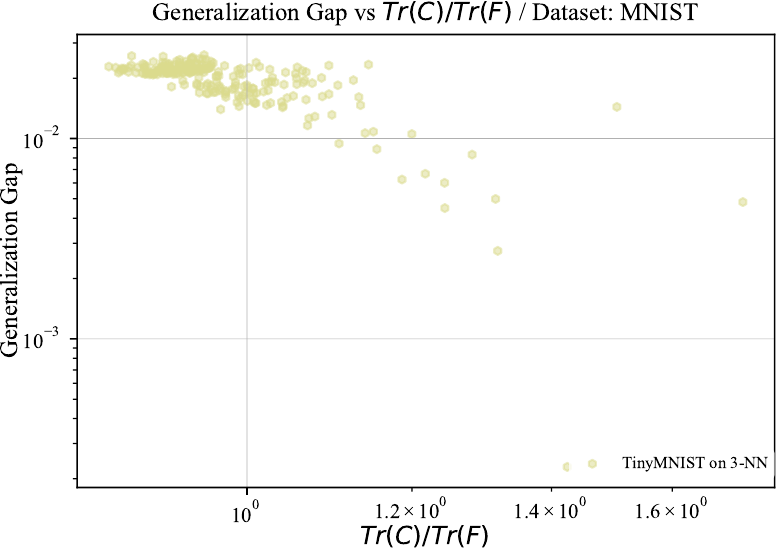}
	\includegraphics[width=0.3\linewidth]{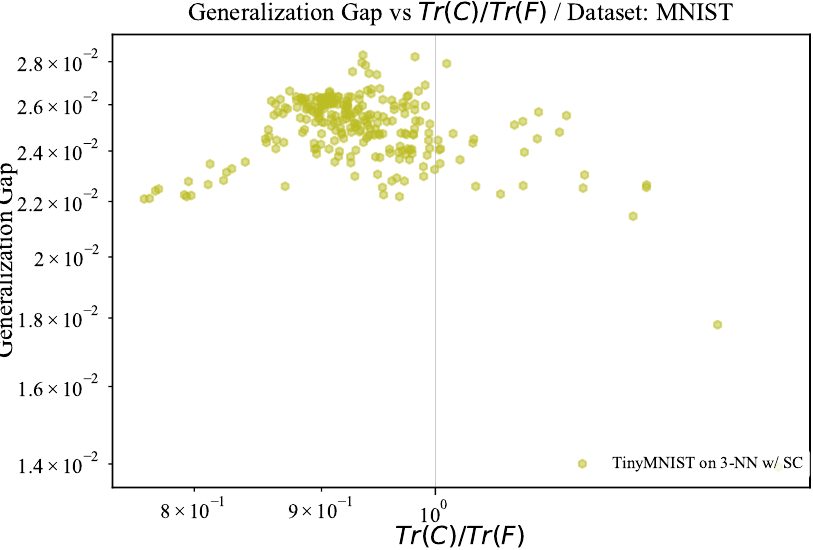}
	\subcaption{\scriptsize{: Generalization gap vs lower bound TIC.}}
\end{minipage}
	\caption{Small-scale experiments: comparison of the TIC estimate and generalization gap.}
	\label{fig:tic-gen-comparison22}
\end{figure}

\clearpage

\begin{table}[h]
\caption{Correlation: TIC Estimates $\mathrm{Tr}({\boldsymbol{F}(\bm{ {\theta}})}^{-1} \boldsymbol{C}(\bm{ {\theta})})$ and Generalization Gap}
\centering
\label{table:small_exact_coef}
\setlength{\tabcolsep}{0.5em} %
{\renewcommand{\arraystretch}{1.2}%
\begin{tabular}{llccc}
\hline
Dataset    & Model             & Spearman's Correlation  & Kendall's $\tau$  & Pearson's Correlation \\ \hline\hline
Tiny MNIST & 2-NN   & 0.532                  & 0.38                         & 0.27                  \\
Tiny MNIST & 3-NN        & -0.62                  & -0.427                       & -0.679                \\
Tiny MNIST & 3-NN w/ SC  & 0.258                  & 0.175                        & -0.332                \\
Tiny MNIST & 3-LNN       & 0.292                  & 0.271                        & 0.305                 \\
Tiny MNIST & 3-LNN w/ SC & 0.886                  & 0.75                         & 0.922                
\end{tabular}}
\end{table}

\begin{table}[h]
\caption{Correlation: TIC Estimates $\mathrm{Tr}({\boldsymbol{F}_{\text{blockdiag}}(\bm{ {\theta})}}^{-1} \boldsymbol{C}_{\text{blockdiag}}(\bm{ {\theta})})$ and Generalization Gap}
\centering
\label{table:small_block_coef}
\setlength{\tabcolsep}{0.5em} %
{\renewcommand{\arraystretch}{1.2}%
\begin{tabular}{llccc}
\hline
Dataset    & Model             & Spearman's Correlation & Kendall's $\tau$ & Pearson's Correlation \\ \hline \hline
Tiny MNIST & 2-NN   & 0.524      & 0.372            & 0.288                 \\
Tiny MNIST & 3-NN        & -0.549     & -0.366           & -0.622                \\
Tiny MNIST & 3-NN w/ SC  & 0.364      & 0.244            & -0.08                 \\
Tiny MNIST & 3-LNN       & 0.26       & 0.257            & 0.252                 \\
Tiny MNIST & 3-LNN w/ SC & 0.937      & 0.823            & 0.944                
\end{tabular}}
\end{table}

\begin{table}[h]
\caption{Correlation: TIC Estimates $\mathrm{Tr}({\boldsymbol{F}_{\text{diag}}(\bm{ {\theta})}}^{-1} \boldsymbol{C}_{\text{diag}}(\bm{ {\theta})})$ and Generalization Gap}
\centering
\label{table:small_diag_coef}
\setlength{\tabcolsep}{0.5em} %
{\renewcommand{\arraystretch}{1.2}%
\begin{tabular}{llccc}
\hline
Dataset    & Model             & Spearman's Correlation & Kendall's $\tau$ & Pearson's Correlation \\ \hline \hline
Tiny MNIST & 2-NN   & -0.309     & -0.23            & -0.234                \\
Tiny MNIST & 3-NN        & -0.297     & -0.203           & -0.415                \\
Tiny MNIST & 3-NN w/ SC  & -0.176     & -0.128           & -0.415                \\
Tiny MNIST & 3-LNN       & 0.275      & 0.237            & 0.288                 \\
Tiny MNIST & 3-LNN w/ SC & 0.932      & 0.796            & 0.918                
\end{tabular}}
\end{table}
\clearpage

\subsection{Additional Results of Practical-Scale Experiments}
\label{appendix_medium_scale}

In this section, we present the results of a practical-scale setting that we have not presented in the main paper.
We observed that in a small-scale setting, DNNs with a large number of layers and SC tend to have a high rank correlation in the TIC estimator and a strong correlation with the generalization gap.
Since it is not computationally feasible to compute the TIC in practical DNNs, this section details experimental results regarding the performance of two approximate estimators of the TIC, using diagonal approximation and its lower bound.

\subsubsection{Practical-Scale Experiments: MNIST Case}

\begin{figure}[h]
    \centering
	\includegraphics[width=0.45\linewidth]{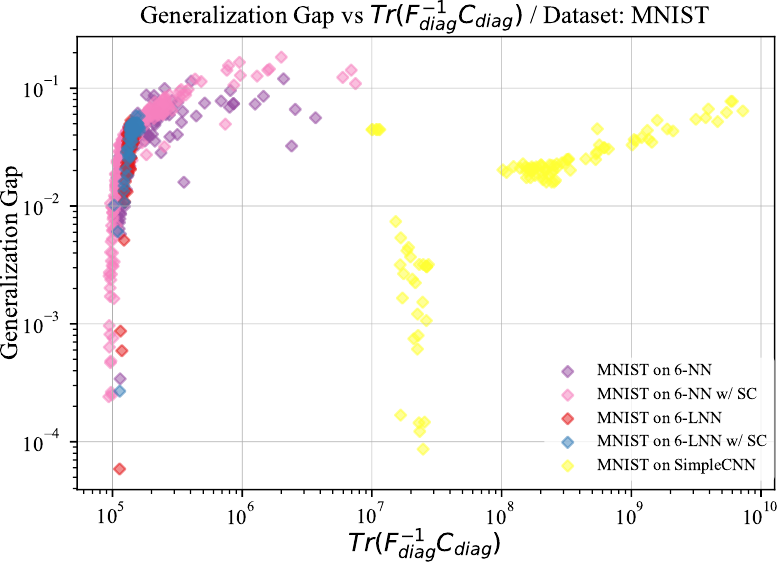}
	\includegraphics[width=0.45\linewidth]{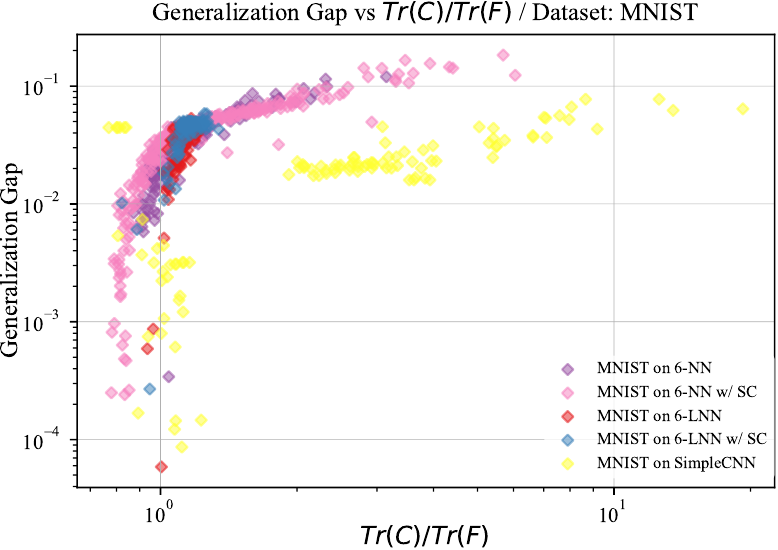}
	\\
	\includegraphics[width=0.45\linewidth]{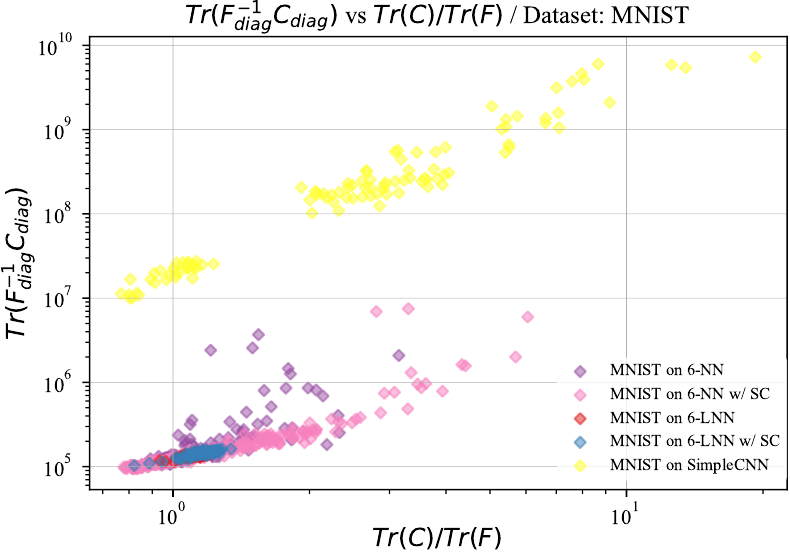}
	\caption{Practical-scale MNIST experiments: comparison of the TIC estimates. }
	\label{fig:medium1}
\end{figure}

As shown in Figure \ref{fig:medium1}, we observed the correlation between the TIC estimator and the generalization gap for the diagonal approximation and the lower bound approach. Both were found to be highly correlated and good estimators of the generalization gap.

\begin{figure}[h]
    \centering
    \includegraphics[width=0.45\linewidth]{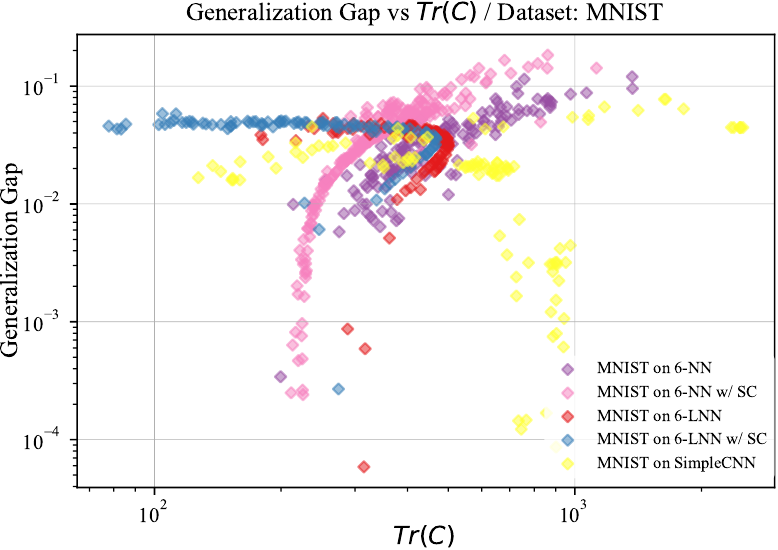}
    \includegraphics[width=0.45\linewidth]{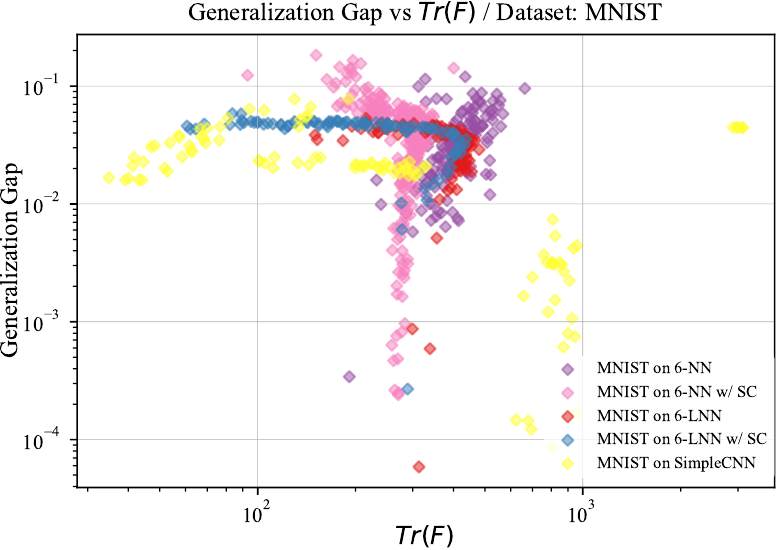}
    \\  
    \includegraphics[width=0.45\linewidth]{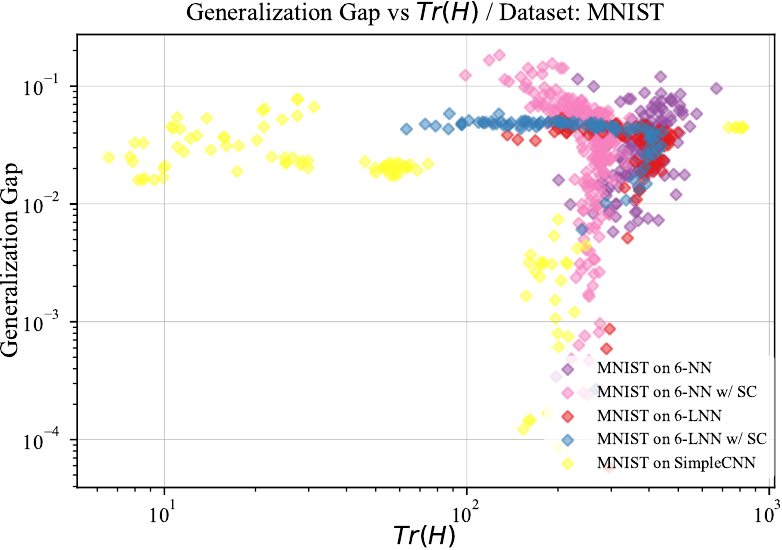}
    \includegraphics[width=0.45\linewidth]{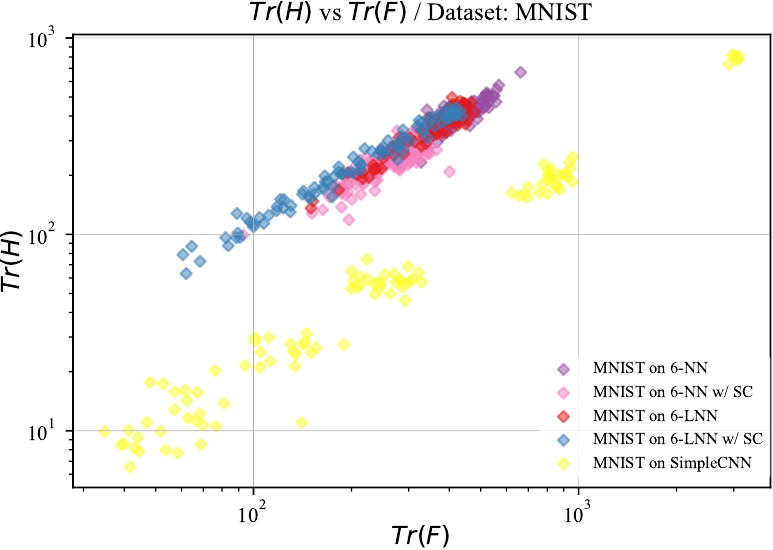}
	\caption{Practical-scale MNIST experiments: elements of the TIC estimates.}
	\label{fig:medium2}
\end{figure}

We also investigated whether the estimated TIC from lower bounds is due to only one of the components of the trace.
As shown in Figure \ref{fig:medium2}, the value of the trace itself was not correlated with the generalization gap. It was also confirmed that the different models behaved differently.
At the same time, we also confirmed that $\mathrm{Tr}\left(\boldsymbol{F}(\bm{ {\theta})} \right)$ is a good approximation of trace $\mathrm{Tr}\left(\boldsymbol{H}(\bm{ {\theta})} \right)$.

\clearpage

\subsubsection{Practical-Scale Experiments: CIFAR10 and CIFAR100 Cases}

\begin{figure}[h]
    \centering
	\includegraphics[width=0.45\linewidth]{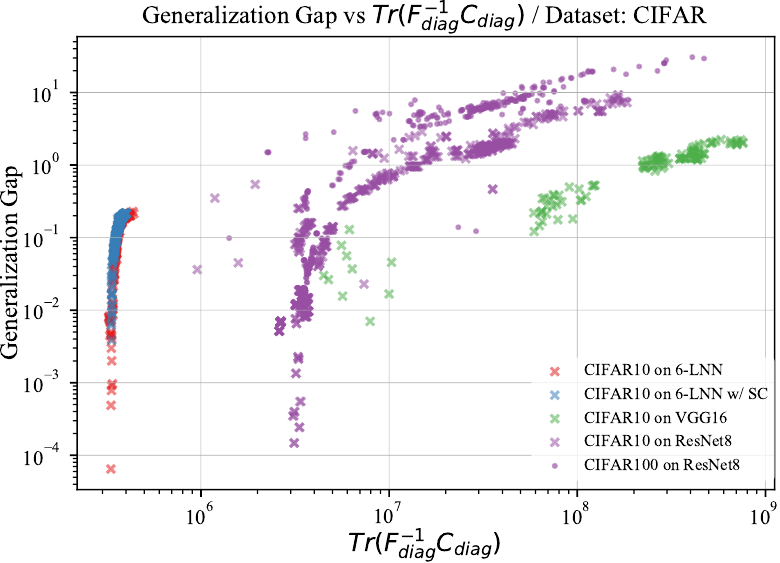}
	\includegraphics[width=0.45\linewidth]{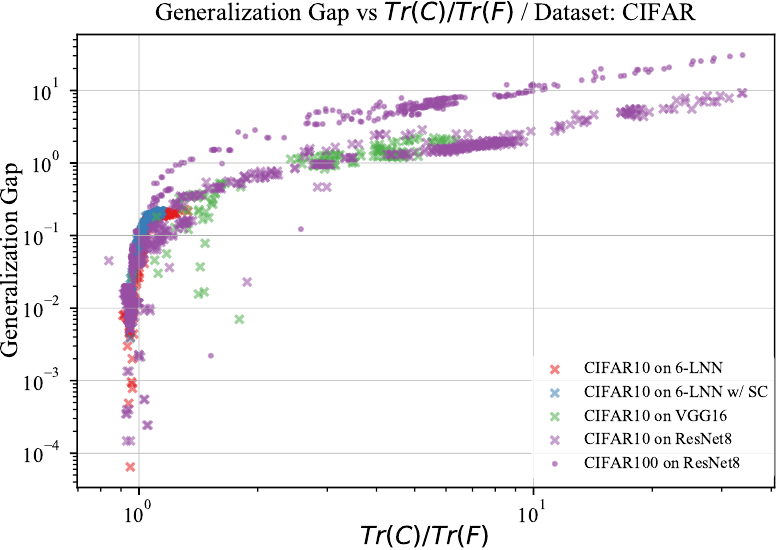}
	\\
	\includegraphics[width=0.45\linewidth]{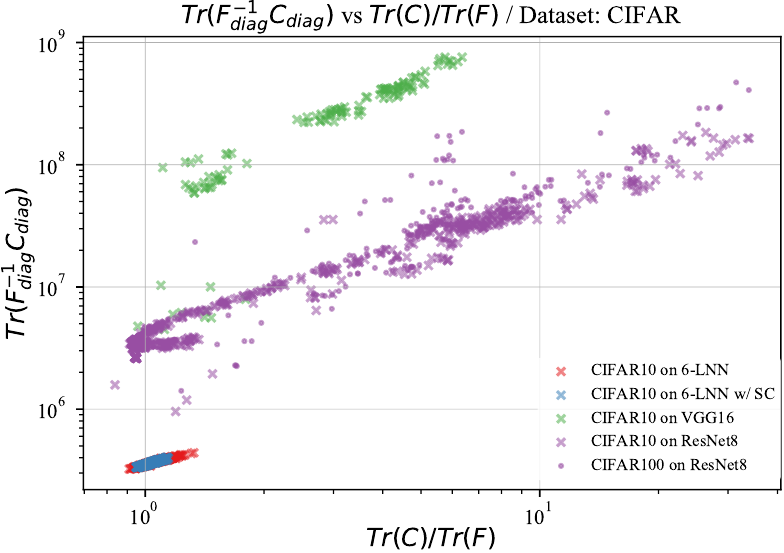}
	\caption{Practical-scale CIFAR experiments: comparison of the TIC estimates.}
	\label{fig:medium3}
\end{figure}

\begin{figure}[h]
    \centering
    \includegraphics[width=0.45\linewidth]{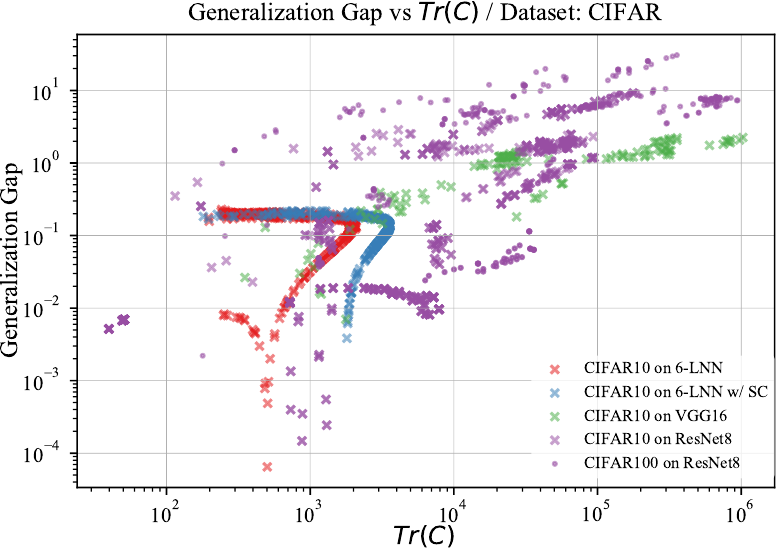}
    \includegraphics[width=0.45\linewidth]{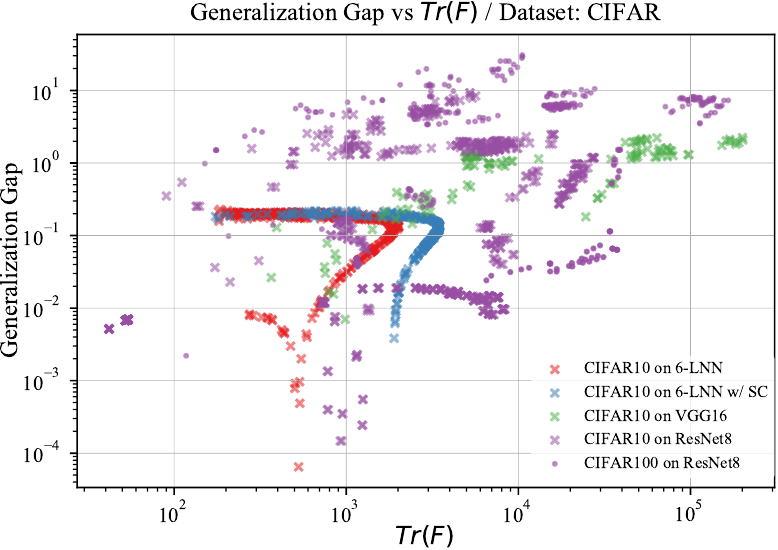}
    \\  
    \includegraphics[width=0.45\linewidth]{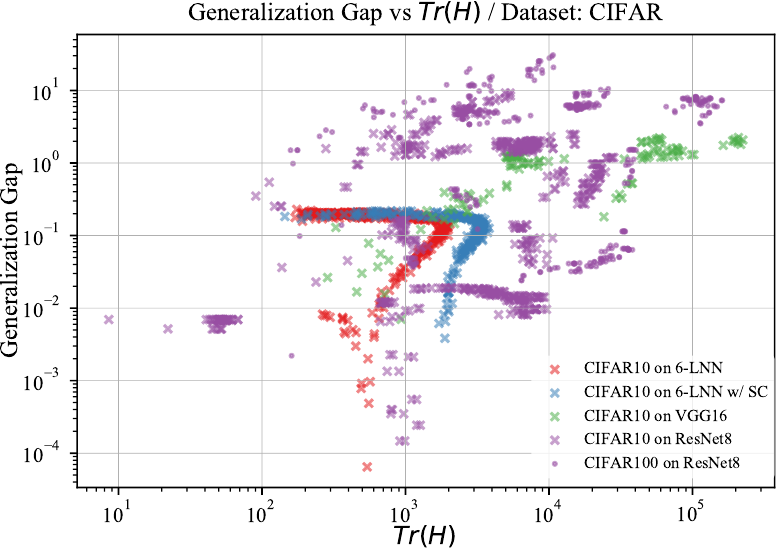}
    \includegraphics[width=0.45\linewidth]{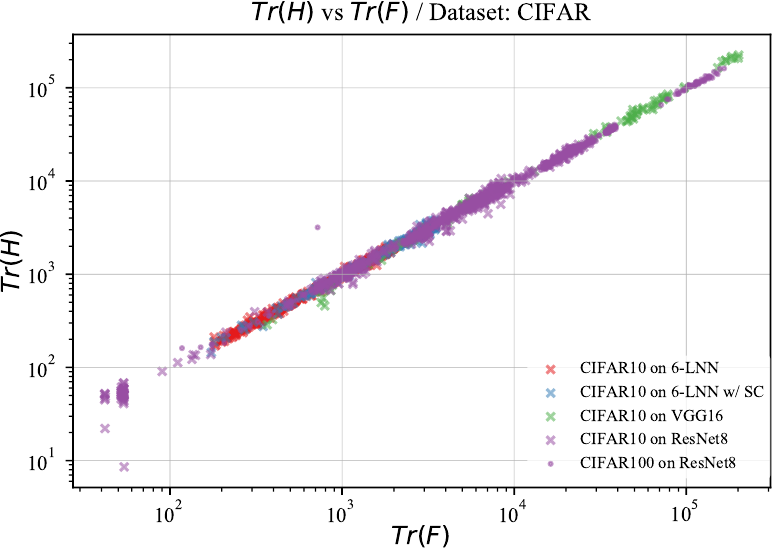}
	\caption{Practical-scale CIFAR experiments: elements of the TIC estimates.}
	\label{fig:medium4}
\end{figure}

\subsubsection{Correlation between TIC Estimates and generalization gap in training process}

Within the scope of our experiments, we found that the TIC can estimate the generalization gap even in the middle of learning for problem settings that are considered to belong to the NTK.

\begin{figure}[h]
    \centering
    \includegraphics[width=0.45\linewidth]{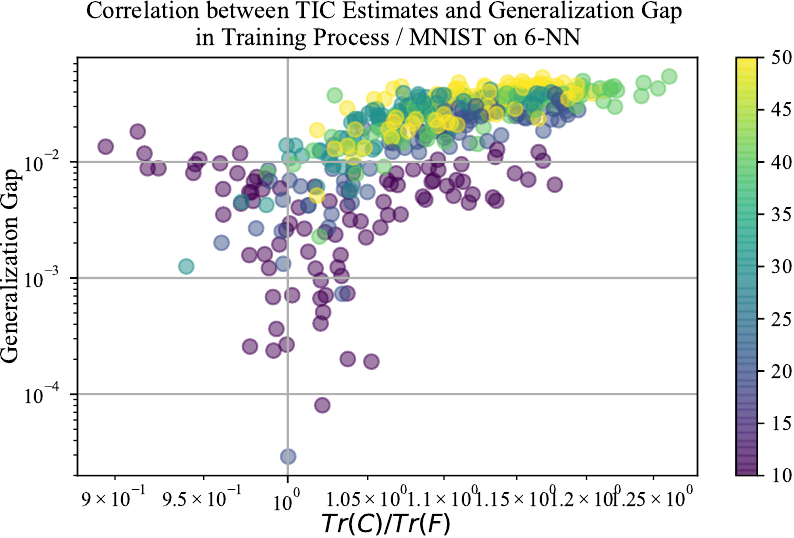}
    \includegraphics[width=0.45\linewidth]{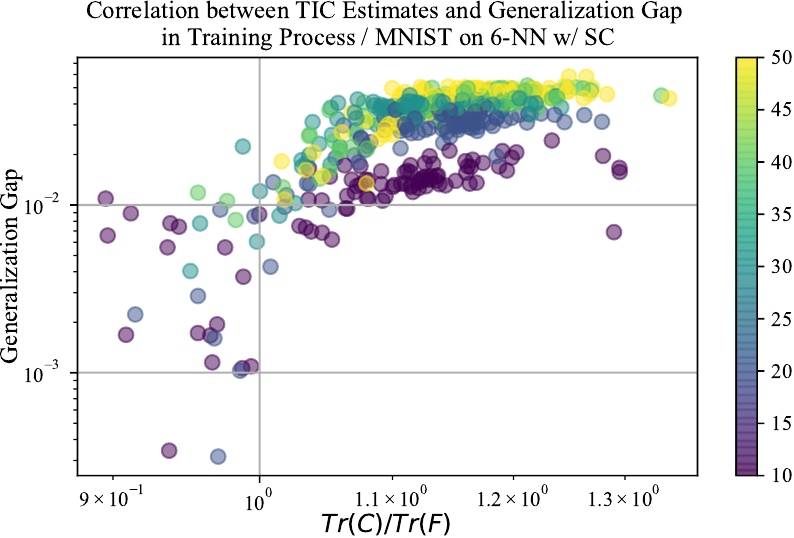}
    \\  
    \includegraphics[width=0.45\linewidth]{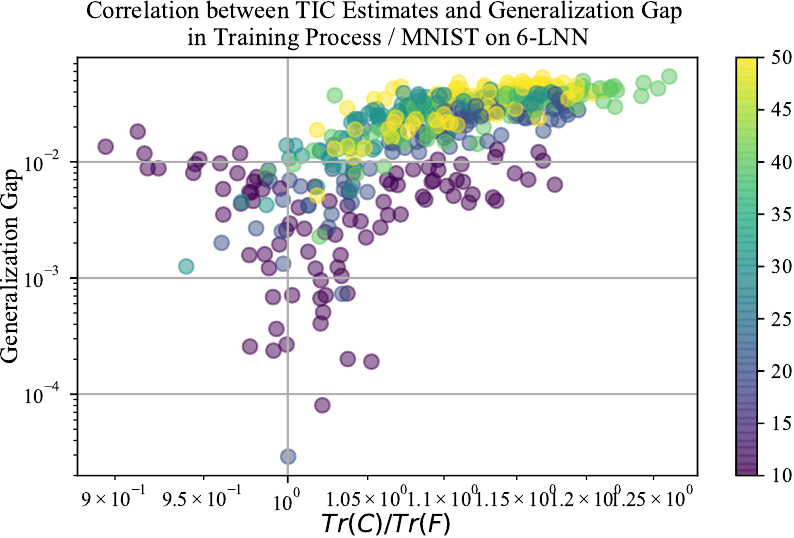}
    \includegraphics[width=0.45\linewidth]{figs/training_process/mnist_lnn_sc.pdf}
    \\
    \includegraphics[width=0.45\linewidth]{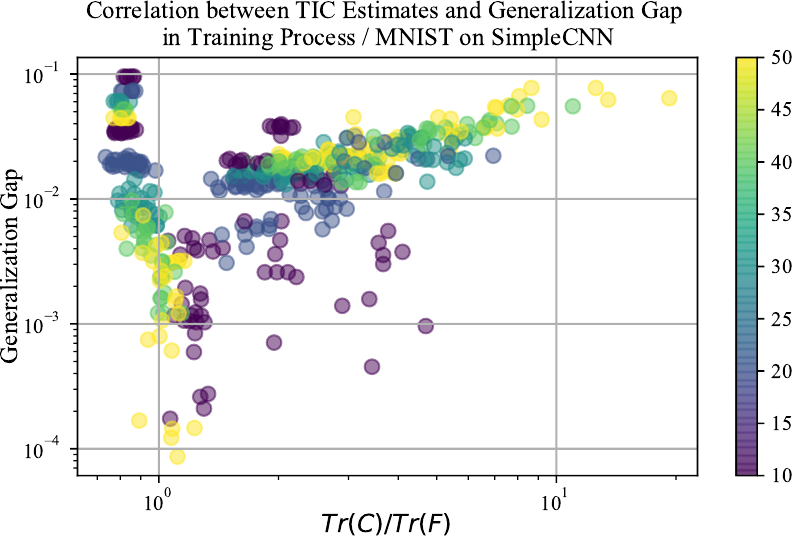}
	\caption{Correlation between generalization gap and TIC estimates in MNIST experiments, through training process. The color map shows the epoch.}
	\label{fig:medium5}
\end{figure}

\begin{figure}[h]
    \centering
    \includegraphics[width=0.45\linewidth]{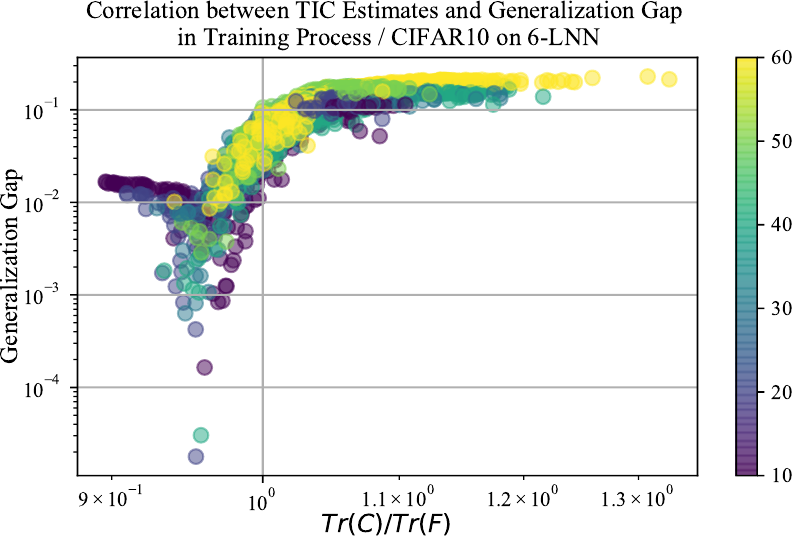}
    \includegraphics[width=0.45\linewidth]{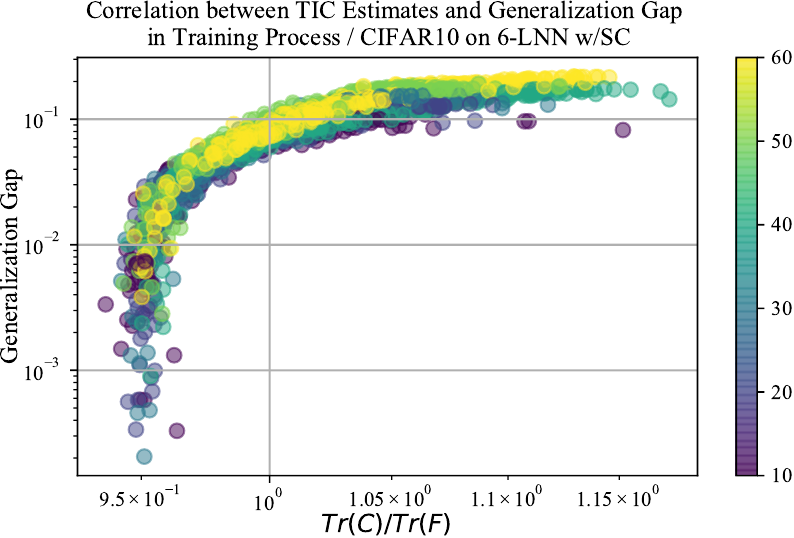}
    \\  
    \includegraphics[width=0.45\linewidth]{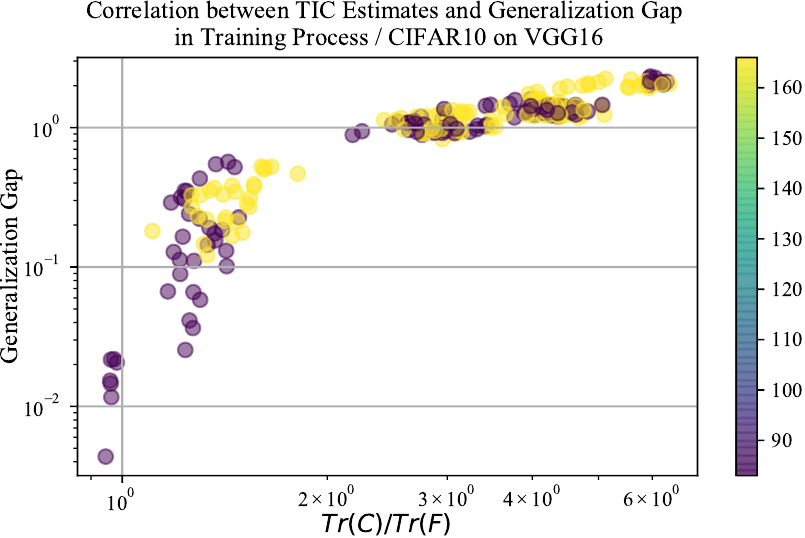}
    \includegraphics[width=0.45\linewidth]{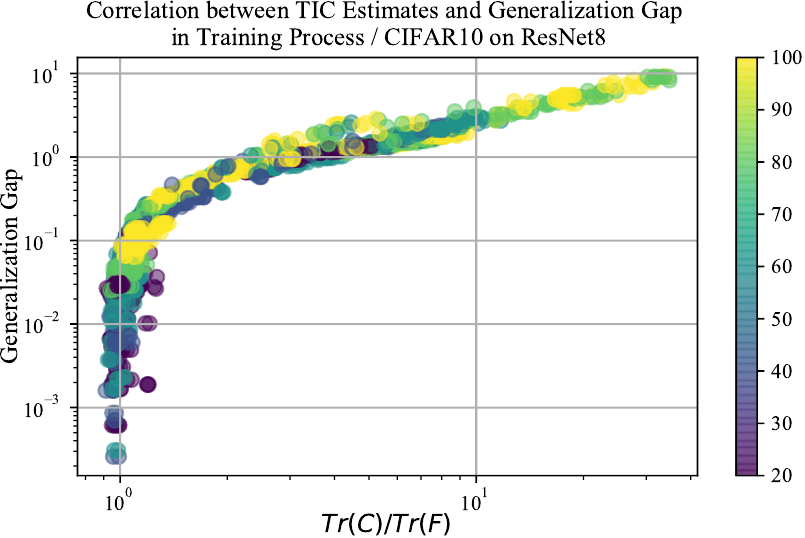}
    \\
    \includegraphics[width=0.45\linewidth]{figs/training_process/cifar100_resnet8.pdf}
	\caption{Correlation between generalization gap and TIC estimates in CIFAR experiments, through training process. The color map shows the epoch.}
	\label{fig:medium6}
\end{figure}
\clearpage

\subsection{Calculation Time Measurements}
 \label{sec:runtime_measurements}

First, we will explain the execution time for the small NN/LNN case.
In the case of using $\mathrm{Tr}\left(\boldsymbol{F}(\bm{ {\theta})} \right)$ and $\mathrm{Tr}\left(\boldsymbol{C}(\bm{ {\theta})} \right)$, the computation time for block-diagonal approximation is only 40\% faster than exact computation.
The computational complexity should have been reduced from $O(d^3)$ to $O(d_l^3)$ by the block-diagonal approximation.
However, overhead, such as memory copying, is dominant, and there was no significant difference in execution time compared to the theoretical amount of computation.
From block-diagonal to diagonal approximation, the experimental results show a reduction of up to 50\% in computation time. Computational complexity is reduced from $O(d_l^3)$ to $O(d)$.

Secondly, the use of $\mathrm{Tr}\left(\boldsymbol{H}(\bm{ {\theta})} \right)$ and $\mathrm{Tr}\left(\boldsymbol{C}(\bm{ {\theta})} \right)$ requires up to 900\% more time than the use of $\mathrm{Tr}\left(\boldsymbol{F}(\bm{ {\theta})} \right)$ and $\mathrm{Tr}\left(\boldsymbol{C}(\bm{ {\theta})} \right)$ in the exact case.
The choice to use $\mathrm{Tr}\left(\boldsymbol{F}(\bm{ {\theta})} \right)$ instead of $\mathrm{Tr}\left(\boldsymbol{H}(\bm{ {\theta})} \right)$ is, therefore, justified in terms of the reduction in computational time.
In the case of block diagonalization, the speedup was only a few percentage points.
In the case of diagonalization, no significant speedup was observed, but the speedup was more than 40 times when trace approximation was performed using the Hutchinson method.
However, the estimation using $\mathrm{Tr}\left(\boldsymbol{F}(\bm{ {\theta})} \right)$ and $\mathrm{Tr}\left(\boldsymbol{C}(\bm{ {\theta})} \right)$ was faster in the small-scale problem setting.

It was observed that in such a small-scale problem setting, there is a 50 times difference in execution time between using $\mathrm{Tr}\left(\boldsymbol{H}(\bm{ {\theta})} \right)$ and $\mathrm{Tr}\left(\boldsymbol{C}(\bm{ {\theta})} \right)$ with exact TIC and computing $\mathrm{Tr}\left(\boldsymbol{F}(\bm{ {\theta})} \right)$ and $\mathrm{Tr}\left(\boldsymbol{C}(\bm{ {\theta})} \right)$ simultaneously and using diagonal approximation.

This speedup should be more significant for larger models.
As mentioned in section \ref{sec_2_2}, it was not feasible to calculate $\mathrm{Tr}\left(\boldsymbol{H}(\bm{ {\theta})} \right)$ in ResNet-8, which requires more than 2.00 TB of memory  without approximation.
While execution time is important, the most important feature is that it is possible to calculate the TIC by approximation.

Next, we compare the execution time on the practical scale.
Among the cases where $\mathrm{Tr}\left(\boldsymbol{F}(\bm{ {\theta})} \right)$ and $\mathrm{Tr}\left(\boldsymbol{C}(\bm{ {\theta})} \right)$ is used, we investigated how much the time can be reduced when $\mathrm{Tr}\left(\boldsymbol{F}(\bm{ {\theta})} \right)$ and $\mathrm{Tr}\left(\boldsymbol{C}(\bm{ {\theta})} \right)$ are computed simultaneously compared to the case where they are computed separately.
In the case of the small-scale and practical-scale networks, the time is reduced by half; however, in the case of SimpleCNN , VGG16, ResNet8, etc., the time was not reduced significantly.
For relatively small models, which could be reduced by about 50\%, the approximation by simultaneous $\mathrm{Tr}\left(\boldsymbol{F}(\bm{ {\theta})} \right)$ and $\mathrm{Tr}\left(\boldsymbol{C}(\bm{ {\theta})} \right)$ calculations was faster than using the Hutchinson method.
In contrast, for networks with a large number of dimensions in the final layer, such as VGG16, the calculation of $\mathrm{Tr}\left(\boldsymbol{C}(\bm{ {\theta})} \right)$ and a fast approximation of $\mathrm{Tr}\left(\boldsymbol{H}(\bm{ {\theta})} \right)$ resulted in a speedup of 10\%-25\% relative to the simultaneous calculation of $\mathrm{Tr}\left(\boldsymbol{F}(\bm{ {\theta})} \right)$ and $\mathrm{Tr}\left(\boldsymbol{C}(\bm{ {\theta})} \right)$.

\begin{landscape}

\begin{table}[]
\caption{\textbf{Full Details of the Runtime Measurement Experiment.} Unit: seconds}
\label{table:runtime_measurements}

\begin{tabular}{|l|l|l|l|l|l|l|l|l|l|l|}
\hline
Dataset    & Model             & \begin{tabular}[c]{@{}l@{}}Exact \\ w/FC\end{tabular} & \begin{tabular}[c]{@{}l@{}}Block Diag \\ w/FC\end{tabular} & \begin{tabular}[c]{@{}l@{}}Diag \\ w/FC\end{tabular} & \begin{tabular}[c]{@{}l@{}}Lower Bound \\ w/FC\end{tabular} & \begin{tabular}[c]{@{}l@{}}Exact \\ w/HC\end{tabular} & \begin{tabular}[c]{@{}l@{}}Block Diag \\ w/HC\end{tabular} & \begin{tabular}[c]{@{}l@{}}Diag \\ w/HC\end{tabular} & \begin{tabular}[c]{@{}l@{}}Lower Bound \\ w/HC\end{tabular} & \begin{tabular}[c]{@{}l@{}}Lower Bound \\ w/HC \\ by Hutchinson's \\Method\end{tabular} \\ \hline
Tiny MNIST & 2-wide NN   & 10.9093                                               & 6.7516                                                     & 3.7273                                               & 1.9069                                                      & 105.4282                                              & 96.0647                                                    & 92.979                                               & 82.0081                                                     & 2.7814                                                                                \\ \hline
Tiny MNIST & 3-NN        & 3.8641                                                & 3.8435                                                     & 3.7937                                               & 1.994                                                       & 23.9083                                               & 22.4055                                                    & 21.7705                                              & 22.4203                                                     & 2.5705                                                                                \\ \hline
Tiny MNIST & 3-NN w/ SC  & 3.8999                                                & 3.8568                                                     & 3.8239                                               & 2.0225                                                      & 26.2303                                               & 24.9561                                                    & 25.0151                                              & 24.9217                                                     & 2.8932                                                                                \\ \hline
Tiny MNIST & 3-LNN       & 3.8656                                                & 3.847                                                      & 3.7991                                               & 1.9521                                                      & 23.188                                                & 22.2685                                                    & 22.1641                                              & 21.6511                                                     & 2.8645                                                                                \\ \hline
Tiny MNIST & 3-LNN w/ SC & 3.8771                                                & 3.8962                                                     & 3.8159                                               & 1.9594                                                      & 23.8571                                               & 22.8106                                                    & 23.0338                                              & 22.0721                                                     & 2.7198                                                                                \\ \hline
MNIST      & 6-NN        & N/A                                                   & N/A                                                        & 3.7235                                               & 1.9172                                                      & N/A                                                   & N/A                                                        & N/A                                                  & N/A                                                         & 2.8949                                                                                \\ \hline
MNIST      & 6-NN w/ SC  & N/A                                                   & N/A                                                        & 3.6813                                               & 1.9459                                                      & N/A                                                   & N/A                                                        & N/A                                                  & N/A                                                         & 3.3679                                                                                \\ \hline
MNIST      & 6-LNN       & N/A                                                   & N/A                                                        & 3.4822                                               & 1.9148                                                      & N/A                                                   & N/A                                                        & N/A                                                  & N/A                                                         & 2.8089                                                                                \\ \hline
MNIST      & 6-LNN w/ SC & N/A                                                   & N/A                                                        & 3.502                                                & 1.9329                                                      & N/A                                                   & N/A                                                        & N/A                                                  & N/A                                                         & 2.9382                                                                                \\ \hline
MNIST      & Simple CNN        & N/A                                                   & N/A                                                        & 4.7059                                               & 2.6337                                                      & N/A                                                   & N/A                                                        & N/A                                                  & N/A                                                         & 8.7958                                                                                \\ \hline
CIFAR10    & 6-LNN       & N/A                                                   & N/A                                                        & 3.6742                                               & 1.9295                                                      & N/A                                                   & N/A                                                        & N/A                                                  & N/A                                                         & 4.1043                                                                                \\ \hline
CIFAR10    & 6-LNN w/ SC & N/A                                                   & N/A                                                        & 3.7003                                               & 1.9453                                                      & N/A                                                   & N/A                                                        & N/A                                                  & N/A                                                         & 2.8996                                                                                \\ \hline
CIFAR10    & ResNet8           & N/A                                                   & N/A                                                        & 12.4677                                              & 10.0031                                                     & N/A                                                   & N/A                                                        & N/A                                                  & N/A                                                         & 90.9633                                                                               \\ \hline
CIFAR10    & VGG16             & N/A                                                   & N/A                                                        & 13.3077                                              & 11.3721                                                     & N/A                                                   & N/A                                                        & N/A                                                  & N/A                                                         & 60.716                                                                                \\ \hline
CIFAR100   & ResNet8           & N/A                                                   & N/A                                                        & 12.4292                                              & 9.9655                                                      & N/A                                                   & N/A                                                        & N/A                                                  & N/A                                                         & 88.7423                                                                               \\ \hline
\end{tabular}
\end{table}

\end{landscape}
\clearpage

\subsection{Experiments with $d/n$ Changes within TinyMNIST}
 \label{sec:tinymnist_additional}
 
 We created a restricted dataset, SmallTinyMNIST, which uses only 5\% of the TinyMNIST data.
 
\begin{table}[h]
\centering
\caption{Additional problem settings are highlighted in bold text}
\begin{tabular}{lll}
\hline
Category                                                                                                                                                                                                                       & Problem Setting: Dataset \& Model                                                                                                                                                                                                                                            & Ratio: $d/n$                                                                                   \\ \hline
\begin{tabular}[c]{@{}l@{}}Small Scale\\ Data Size \textless 1 MB\\ Model \textless 50 KB\end{tabular} & \begin{tabular}[c]{@{}l@{}}TinyMNIST on 2-NN w/o SC\\ TinyMNIST on 3-NN w/o SC, 3-NN w/ SC\\ TinyMNIST on 3-LNN w/o SC, 3-LNN w/ SC\\ {\textbf{ SmallTinyMNIST on 3-LNN w/o SC}}\\ {\textbf{ TinyMNIST on Wide 3-NN w/ SC}}\end{tabular} & \begin{tabular}[c]{@{}l@{}}0.09\\ 0.02\\ 0.02\\ {\textbf{0.36}}\\ {\textbf{ 13.90}}\end{tabular} \\ \hline
\end{tabular}
\end{table}

We conducted training evaluations using SmallTinyMNIST to investigate the relationship between the TIC estimator and the generalization gap for patterns with relatively large $d/n$ ratios.

We also prepared a network with a large number of $d$ as a 3-layer Wide-NN, and confirmed the effectiveness of the TIC in settings where the $d/n$ ratio exceeds 10.

\begin{figure}[h]
    \centering
	\includegraphics[width=0.49\linewidth]{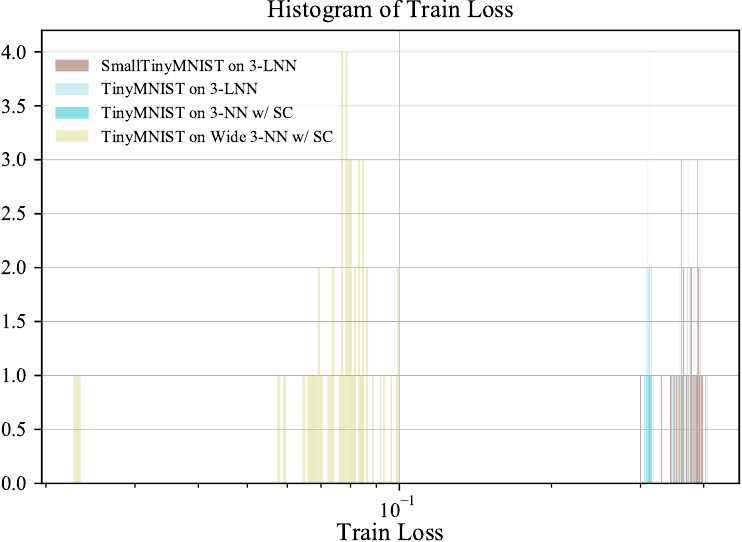}
	\includegraphics[width=0.49\linewidth]{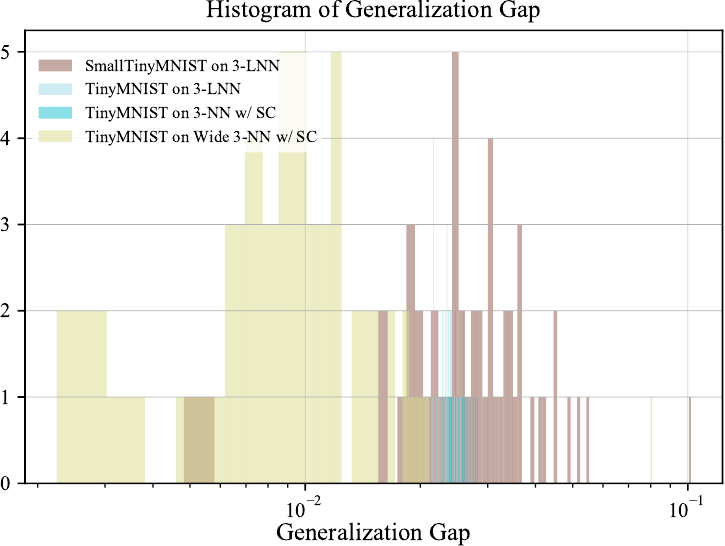}

	\caption{Distribution of training loss and generalization gap on the trained models.}
	\label{fig:tinymnist_dsitribution_loss}
\end{figure}

\begin{figure}[h]
    \centering
	\includegraphics[width=0.90\linewidth]{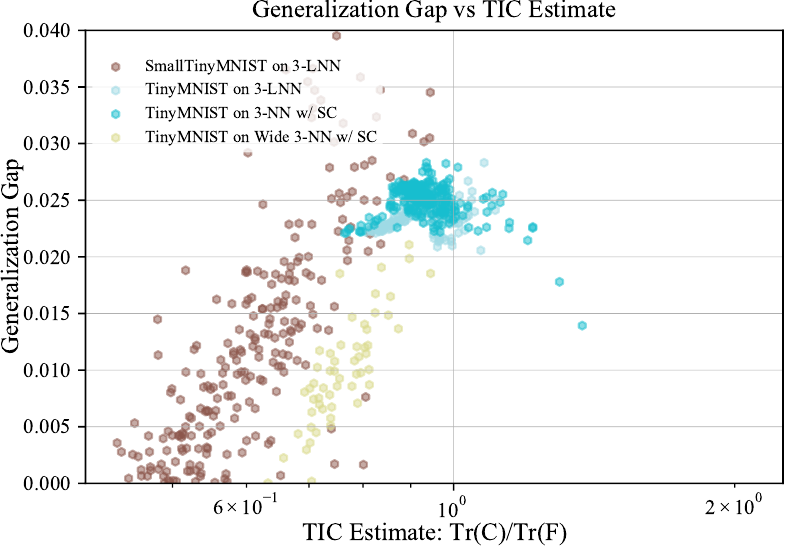}
	\caption{Relationship between TIC estimates and generalization gap.}
	\label{fig:tinymnist_corr_plot}
\end{figure}

The distributions of the loss and the generalization gap are shown in Figure \ref{fig:tinymnist_dsitribution_loss}. For comparative purposes, problem settings with the same or related  networks are included in the plots and tables.
The correlation between the TIC and the generalization gap is shown in Figure \ref{fig:tinymnist_corr_plot}.

\begin{table}[h]
\caption{Correlation: TIC estimates $\mathrm{Tr}(\boldsymbol{C}(\bm{ {\theta})}) / \mathrm{Tr}(\boldsymbol{F}(\bm{ {\theta})})$ and generalization gap}
\centering
\label{table:tinymnist_correlation}
\setlength{\tabcolsep}{0.5em} %
{\renewcommand{\arraystretch}{1.2}%
\begin{tabular}{llccc}
\hline
Model             & Dataset     & Spearman's Correlation & Kendall's $\tau$ & Pearson's Correlation \\ \hline\hline
3-LNN       & Tiny MNIST  & 0.277                  & 0.238                        & 0.256                 \\
3-LNN       & Small Tiny MNIST  & 0.806                  & 0.622                        & 0.811                \\
\hline
3-NN w/ SC  & Tiny MNIST  & -0.19                  & -0.137                       & -0.347                \\
3-Wide NN w/ SC  & Tiny MNIST  & 0.834                 & 0.656                  & 0.967                \\
\hline
\end{tabular}}
\end{table}

\end{document}